\documentclass[journal]{IEEEtran}

\ifCLASSINFOpdf

\else

\fi

\usepackage{graphicx,wrapfig,fullpage,amsmath,hhline,epsfig,verbatim,url,amssymb,multicol,multirow,cite}
\usepackage{times,color,soul} 
\usepackage[left=0.75in,top=0.70in,right=0.75in,bottom=0.80in]{geometry}
\usepackage{amsbsy,latexsym,mathrsfs,mathtools}

\usepackage{mathptmx} 
\DeclareMathAlphabet{\mathcal}{OMS}{cmsy}{m}{n}
\usepackage{bm}
\usepackage{float}
\usepackage{color,soul}
\usepackage{dsfont}
\usepackage{comment}
\usepackage{algorithm}
\usepackage{algorithmic}
\usepackage{psfrag}   
\usepackage{common_drs_vars}

\usepackage{url}
\usepackage{setspace}
\usepackage[table]{xcolor}
\usepackage{paralist}
\usepackage{booktabs}
\usepackage{gensymb}
\usepackage[switch]{lineno}

\usepackage{flushend}
\usepackage{algorithm} 
\usepackage{algorithmic}  
\usepackage[algo2e]{algorithm2e} 
\usepackage{diagbox}

\definecolor{darkgreen}{rgb}{0,0.6,0}
\definecolor{darkblue}{rgb}{0,0,0.6}
\usepackage[bookmarks=true,colorlinks=true,pdfpagemode=UseNone,citecolor=darkgreen,linkcolor=black,urlcolor=darkblue]{hyperref}

\newtheorem{thm}{Theorem}
\newtheorem{rmk}{Remark}
\newtheorem{prop}{Proposition}

\begin{document}

\title{Time-Varying Foot-Placement Control for Underactuated Humanoid Walking on Swaying Rigid Surfaces}

\author{Yuan Gao$^{*,1}$, Victor Paredes$^{*,2}$, Yukai Gong$^{3}$, Zijian He$^4$, Ayonga Hereid$^2$, Yan Gu$^4$
\thanks{$*$ Equal contribution.}
\thanks{$^{1}$Y. Gao is with the College of Engineering, University of Massachusetts Lowell, Lowell, MA 01854, USA.
E-mail: {yuan\_gao@student.uml.edu.}}%
\thanks{$^{2}$V. Paredes and A. Hereid are with the Department of Mechanical and Aerospace Engineering, The Ohio State University, Columbus, OH 43210, USA.
E-mails: {paredescauna.1@buckeyemail.osu.edu, hereid.1@osu.edu.}}%
\thanks{$^{3}$Y. Gong is with the Robotics Department, University of Michigan, Ann Arbor, MI 48105, USA. E-mail: ykgong@umich.edu.}
\thanks{$^{4}$Z. He and Y. Gu are with the School of Mechanical Engineering,
Purdue University, West Lafayette, IN 47907, USA.
E-mails: \{he348, yangu\}@purdue.edu.}%
}

\maketitle

\begin{abstract}
Locomotion on dynamic rigid surface (i.e., rigid surface accelerating in an inertial frame) presents complex challenges for controller design, which are essential for deploying humanoid robots in dynamic real-world environments such as moving trains, ships, and airplanes. 
This paper introduces a real-time, provably stabilizing control approach for underactuated humanoid walking on periodically swaying rigid surface.
The first key contribution is the analytical extension of the classical angular momentum-based linear inverted pendulum model from static to swaying grounds. This extension results in a time-varying, nonhomogeneous robot model, which is fundamentally different from the existing pendulum models.
We synthesize a discrete footstep control law for the model and derive a new set of sufficient stability conditions that verify the controller's stabilizing effect.
Another key contribution is the development of a hierarchical control framework that incorporates the proposed footstep control law as its higher-layer planner to ensure the stability of underactuated walking. The closed-loop stability of the complete hybrid, full-order robot dynamics under this control framework is provably analyzed based on nonlinear control theory.
Finally, experiments conducted on a Digit humanoid robot, both in simulations and with hardware, demonstrate the framework's effectiveness in addressing underactuated bipedal locomotion on swaying ground, even in the presence of uncertain surface motions and unknown external pushes.

\end{abstract}


\IEEEpeerreviewmaketitle

\section{Introduction}
\label{section: introduction}

Locomotion is a fundamental capability of legged robots, enabling them to navigate diverse environments to perform a variety of real-world tasks, such as warehouse automation, home assistance, package delivery, and search and rescue.
Although robust walking on complex yet static terrains such as slopes~\cite{kim2007walking}, stairs~\cite{caron2019stair}, and gravel~\cite{pajon2017walking} has been extensively studied, the focus has primarily been on overcoming significant challenges in controller design due to the inherently unstable, hybrid, nonlinear dynamics of robots, compounded by uncertainties such as terrain variations~\cite{bauby2000active}.

Conversely, bipedal locomotion on dynamic rigid surface (DRS)—rigid surface that move within an inertial frame, such as those on buses, vessels, and airplanes—remains relatively underexplored. This is despite the critical importance of such capabilities in scenarios involving inspection, maintenance, and emergency response on moving transportation vehicles and oil platforms.

\begin{figure}[t]
    \centering
    \includegraphics[width=1\linewidth]{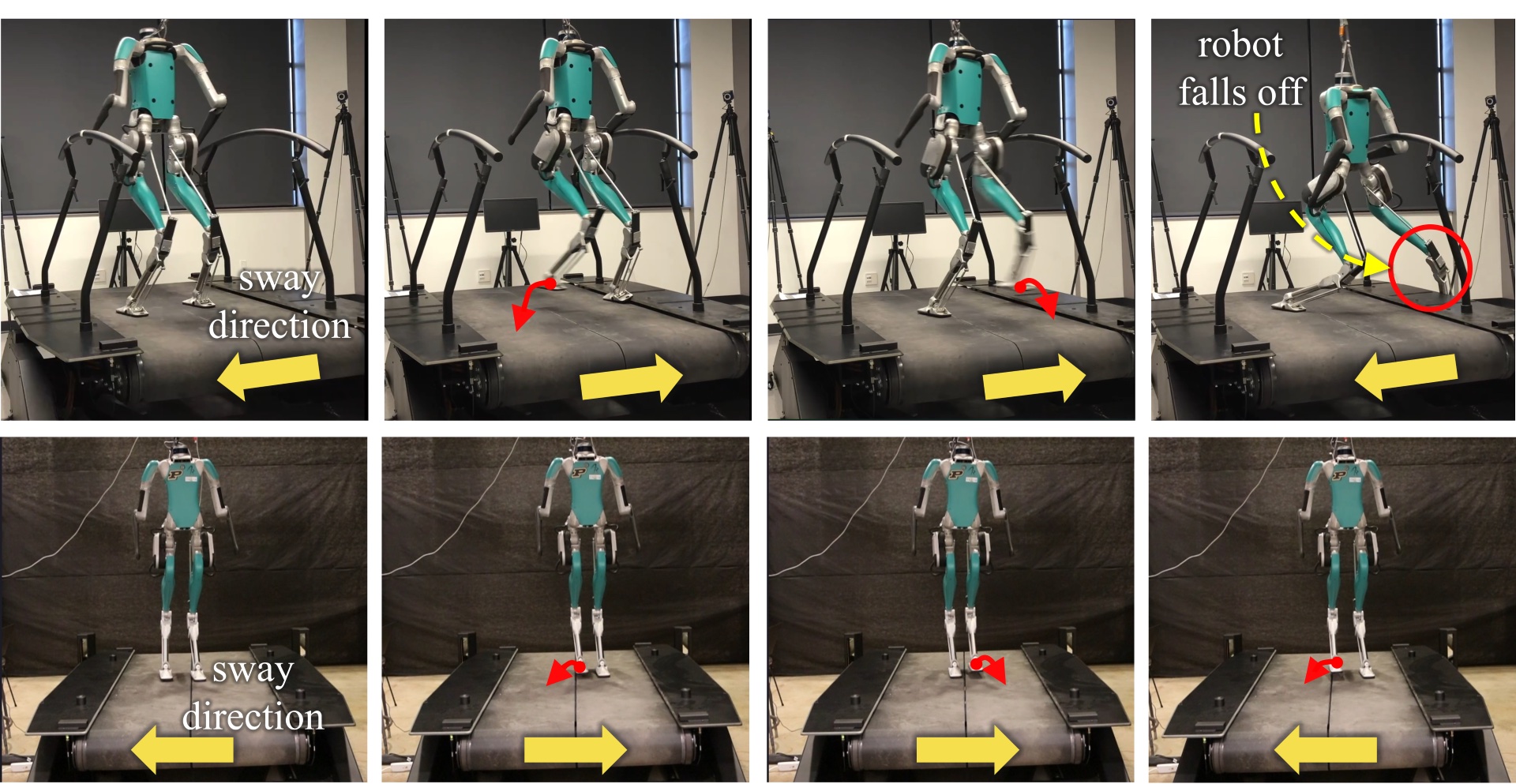}
    \vspace{-0.2 in}
    \caption{Top: The proprietary stepping-in-place controller of the humanoid robot Digit fails to maintain stability during DRS sway at a frequency of 0.25 Hz and amplitude of 5 cm. Bottom: Our proposed framework reliably achieves stable walking under identical conditions.} 
    \label{Fig:digit_fail.png}
    \vspace{-0.1 in}
\end{figure}

Current robust controllers designed for static terrains may only handle mild DRS motions and are prone to failure under more intense or prolonged DRS movements. For instance, as depicted in Fig. \ref{Fig:digit_fail.png}, the proprietary controller of the Digit humanoid robot fails to maintain stable locomotion under moderate DRS sway. The primary issue is that omitting explicit consideration of surface motion results in it acting as a temporally persistent disturbance, giving the robot insufficient recovery time. This may induce excessive control actions that exceed the robot's actuator capabilities, leading to potential instability and safety risks.

Thus, this paper introduces a control framework that explicitly incorporates DRS motion into the dynamic modeling and controller design of the robot. Our approach not only aims to enable stable locomotion on DRS for underactuated legged robots but also seeks to offer performance guarantees through provable stabilization of the walking dynamics associated with DRS locomotion.

\vspace{-0.1 in}
\subsection{Related work}

\subsubsection{Stability criteria and control for fully actuated walking}
Fully actuated walking has been long studied and can be used to facilitate complex tasks beyond walking, such as simultaneous execution of locomotion and manipulation~\cite{kim2018computationally}.
For fully actuated walking where a robot's number of degrees of freedom (DoFs) equals the number of actuators, an extensively studied walking control paradigm is based on the zero moment point (ZMP) condition~\cite{vukobratovic2006zmp, firmani2013theoretical}.
This condition constrains the robot's center of pressure to be strictly within the support polygon, thus restricting a support foot from rolling about its edge.
By enforcing the ZMP condition and the other ground-contact constraints (e.g., friction cone and unilateral constraints) on the full-order robot model (i.e., a model describing the robot dynamics associated with all DoFs), existing motion generation strategies can produce reference motions that correspond to balanced walking. 
Despite the effectiveness of these methods for static terrain, their computational complexity often precludes real-time application due to the high dimensionality and nonlinear nature of legged robot dynamics.

To address this, reduced-order models such as the linear inverted pendulum (LIP) have been developed, simplifying the center of mass (CoM) dynamics to enable feasible motion planning under the ZMP constraints~\cite{kajita20013d, tedrake2015closed, sugihara20213d}. Methods including the capture point~\cite{pratt2006capture, hof2005condition, englsberger2011bipedal,caron2019capturability} and divergent component of motion (DCM)~\cite{takenaka2009real,englsberger2013three
, englsberger2015three} also employ the LIP to enable computationally efficient gait generation.

Still, existing gait generation methods based on the ZMP, Capture Point, and DCM primarily focus on fully actuated robots, rather than underactuated walking where the number of DoFs is greater than the number of actuators. 
Further, as these approaches use robot models (e.g., the classical LIP model) that assume the ground is static, they may not satisfactorily ensure stability or robustness of walking when the ground accelerates in an inertial frame.

\subsubsection{Stability criteria and control for underactuated walking}

Underactuation in walking robots can be induced by a passive support foot (e.g., point foot).
With deactivated foot joints, a robot with finite-size feet effectively become a robot with point feet, which simplifies the analysis of foot-ground contact~\cite{gong2022zero}.
Also, exploiting underactuation can facilitate dynamic, agile, and energy-efficient gait~\cite{collins2005bipedal, reher2021dynamic}.
Yet, underactuation poses additional challenges as unactuated dynamics cannot be directly altered by joint torques~\cite{westervelt2007feedback}.

The hybrid zero dynamics (HZD) framework~\cite{westervelt2007feedback, poulakakis2009spring, sreenath2011compliant, sreenath2013embedding} is an extensively studied control approach for underactuated bipedal walking.
This approach treats the legged robot’s hybrid dynamics model, which describes both the continuous motion (e.g., foot swinging) and discrete events (e.g., foot landings), to ensure the provable stability of periodic orbits within the robot's state space.
While effective for both fully actuated and underactuated robots, the complexity and computational demands of HZD can limit its application to pre-calculated gaits~\cite{ramezani2014performance, Hereid2017FROST,gao2019global,paredes2020dynamic,wensing2023optimization,gao2019dscc}, which may be restrictive for real-world scenarios.
Recent studies have attempted to mitigate these issues via offline generation of a family of different gaits~\cite{motahar2016composing}, improving the robustness under disturbances~\cite{hartley2017stabilization} and enabling walking over discrete terrain~\cite{nguyen2017dynamic}.

To enable flexible and efficient real-time adjustment of gait characteristics such as step lengths, which are critical for guaranteeing walking stability and enhancing robustness, Xiong et al.~\cite{xiong2019orbit,xiong20223} analytically extended the classical continuous-time LIP model with a discrete model of foot landings, creating a hybrid LIP (H-LIP) model. This development led to the establishment of provable stability conditions that guide the planning of stabilizing foot-landing positions. Similarly, Gong et al.~\cite{gong2022zero} introduced a hybrid version of the angular-momentum-based LIP (ALIP) model, explicitly incorporating a robot's contact angular momentum as a state variable to improve robustness against motor torques and ensure the invariance of the contact angular momentum at foot-touchdown impacts. These reduced-order models have recently been extended to accommodate vertical CoM movements, facilitating navigation over rough terrain and stairs~\cite{xiong2021slip,dai2022bipedal}, and have been integrated with adaptive regulators to enhance the robustness of locomotion~\cite{paredes2020dynamic}.

While these approaches have achieved remarkable performance on physical robots traversing various complex real-world environments, they generally assume that the ground is stationary. Thus, significant ground motion introduces unmodeled uncertainties that can lead to walking instability under these previous control frameworks.

\subsubsection{Robust walking control for static terrain}

Beyond the existing work reviewed earlier, model predictive control schemes
are well-suited for addressing perturbations such as uncertain rough terrain since they explicitly enforce the physical limits and constraints~\cite{scianca2020mpc, gibson2022terrain, acosta2023bipedal,gu2024robust}. 
In addition to model-based approaches, reinforcement learning based control has realized highly robust and agile walking in a wide variety of complex real-world terrains~\cite{xie2018feedback, castillo2021robust, li2021reinforcement,siekmann2021blind, castillo2022reinforcement, li2023robust}. 
However, these methods often assume static ground conditions and does not explicitly account for ground motion,
which may not ensure locomotion stability on accelerating ground.

\subsubsection{Stability conditions and control for accelerating ground}

Producing stable gait on DRS presents a complex control challenge due to the inherently time-varying robot dynamics emerging from robot-DRS interactions (Fig.~\ref{Fig:digit_fail.png}). 
Recently, there has been an increasing interest in studying the modeling, control design, and stability analysis of quadrupedal walking on vertically moving DRS. 
Iqbal et al.~\cite{9108552} introduced a provably stabilizing nonlinear control approach that explicitly deals with the hybrid, nonlinear, and time-varying, full-order robot dynamics associated with quadrupedal walking on DRS.
This approach, however, relies heavily on offline motion generation to meet real-time implementation needs, limiting its practical application.

Furthermore, the classical LIP model has been modified to overcome its static terrain assumption by explicitly considering the vertical movement of DRS, allowing for real-time motion generation and control for quadrupedal walking~\cite{iqbal2023asymptotic,iqbal2023analytical, iqbal2022drs}. Although these modifications provide stability conditions for the extended LIP models during vertical DRS movements, they cannot be directly applied to address horizontally swaying surfaces. This is because the robot dynamics differ fundamentally during horizontal versus vertical motions. This research aims to fill this gap by creating a new dynamics model and control framework for locomotion on swaying DRS.


\vspace{-0.1 in}
\subsection{Contribution}

This study aims to create a real-time, provably stabilizing control approach for underactuated humanoid walking during periodic and horizontal DRS motions, even in the presence of uncertainties such as sudden pushes and uncertain ground movements.
To the best of our knowledge, this study represents one of the earliest efforts to explicitly address temporally persistently swaying ground and enable reliable underactuated bipedal locomotion on accelerating ground.

Our preliminary work~\cite{gao2023time} analytically extended the ALIP model~\cite{gong2022zero} from static to swaying DRS, producing a new reduced-order model referred to as ALIP-DRS. 
This initial study also provably analyzed the model's stability and synthesized a hierarchical control framework based on the resulting stability condition. 
Yet, the condition assumed that the robot's gait period matched the DRS motion period exactly, which can be restrictive for practical applications. Moreover, the stability of the complete full-order closed-loop system under the proposed framework was not analyzed, and experimental validation on hardware was absent.

Building upon our preliminary work~\cite{gao2023time}, the substantial, new contributions of this study are:
\begin{itemize}
\item [(a)] Theoretical generalization of the stability condition for the ALIP-DRS model by allowing the ratio between the gait and the DRS motion periods to be any positive rational number.
Previously, the preliminary condition~\cite{gao2023time} assumed a unity ratio.
Also, supporting propositions of the main stability theorem are provided, whereas previously only the theorem was presented. 
\item [(b)] Development of a three-layer control framework that achieves stable underactuated bipedal walking during DRS sway.
The framework extends beyond the preliminary control framework in~\cite{gao2023time}, which was limited to the special stability condition for the ALIP-DRS.
\item [(c)] Overall Lyapunov-based stability analysis for the complete closed-loop full-order system under the proposed control framework, which verifies the provable stability of the system and was not given in~\cite{gao2023time}.
\item [(d)] Experimental validation assessing the effectiveness of the proposed control approach in ensuring walking stability and robustness under various DRS sway motions and uncertainties such as uncertain DRS motions in multiple directions and unknown sudden pushes.
\end{itemize}

This paper is structured as follows.
Section~\ref{Section: Preliminaries} introduces the proposed ALIP-DRS model for locomotion during DRS sway.
Section~\ref{sec: ALIP control and stability} presents the proposed discrete footstep control and stability analysis for the ALIP-DRS.
Section~\ref{section: Hierarchical Planning and Control Framework} describes a hierarchical control framework that utilizes the ALIP-DRS footstep controller to indirectly stabilize the unactuated dynamics of a full-dimensional underactuated robot.
Section~\ref{Section: Stability} provides the stability analysis of the complete full-order system under the proposed framework.
Sections~\ref{Section: simulation} and~\ref{Section: experiment} report the simulation and hardware experiment results.
Section~\ref{section: discussion} discusses the capabilities and limitations of this work.
Section~\ref{section: conclusions} gives the concluding remarks.

\vspace{-0.1 in}
\section{ALIP-DRS Model}
\label{Section: Preliminaries}

This section introduces the derivation of the proposed ALIP-DRS model for bipedal walking on swaying DRS.

\begin{figure}[t]
    \centering    \includegraphics[width=1\linewidth]{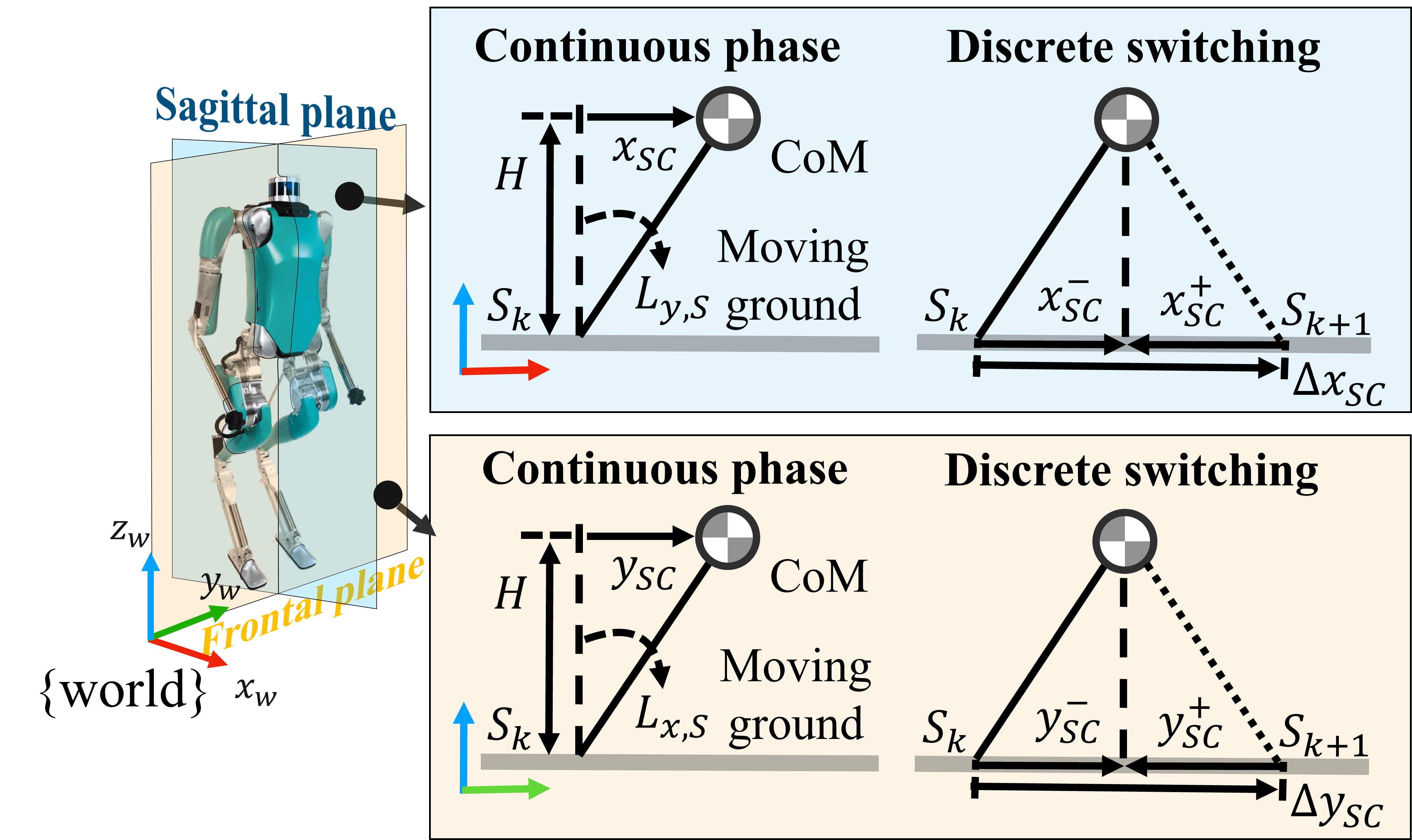}
    \vspace{-0.2 in}
    \caption{Illustration of the proposed hybrid ALIP-DRS model, showing both continuous and discrete components in the sagittal and frontal planes.}
    \label{fig:ALIP state}
    \vspace{-0.2 in}
\end{figure}

The key to the derivation is the analytical extension of the classical ALIP model~\cite{gong2022zero} from static to dynamic terrain, achieved by explicitly incorporating the DRS sway in the model.
This incorporation leads to the distinct time-varying and nonhomogeneous nature of the ALIP-DRS.
Also, we explicitly consider the discrete robot dynamics at foot-landing events, which allows us to exploit foot placement in the controller design to ensure ALIP-DRS's stability.
This consideration results in the hybrid nature of the ALIP-DRS.
Figure~\ref{fig:ALIP state} illustrates the three-dimensional (3-D) ALIP-DRS.

\vspace{-0.1 in}
\subsection{Contact Angular Momentum during DRS Sway}

To analytically extend the classical ALIP from static ground to swaying DRS, we first mathematically define the robot's contact angular momentum during DRS sway.

The contact angular momentum, denoted as $\mathbf{L}_{S}=[L_{x,S},L_{y,S},L_{z,S}]^{\intercal}$, 
refers to the robot's total angular momentum about the ground-contact point.
We choose $L_S$ as a state variable due to the benefits~\cite{gong2022zero} reviewed in Sec.~\ref{section: introduction}.

We use $S$ to denote the contact point on the DRS, which moves along with the DRS in the inertial world frame.
Also, let $m$, $\mathbf{v}_{CoM}$, $\mathbf{p}_{SC}$, and $\mathbf{L}_{CoM}$ denote the robot's total mass, absolute CoM velocity in the world frame, CoM position relative to the contact point $S$, and angular momentum about the CoM, respectively.
Then, $\mathbf{L}_{S}$ is given by:
\begin{equation}
    \mathbf{L}_S = \mathbf{L}_{CoM}
    +
    \mathbf{p}_{SC}\times (m\mathbf{v}_{CoM}).
    \label{Eq: LS and LCOM}
\end{equation}

To derive the dynamics of $\mathbf{L}_{S}$, we introduce the point $A$, which is a static point in the world frame that aligns with 
$S$ at the given time instant.
Further, let $\mathbf{p}_{AC}$ and $\mathbf{p}_{SA}$ be the position vectors pointing from $A$ to CoM and from $S$ to $A$, respectively.
Then, we introduce the angular momentum about the point $A$, denoted as $\mathbf{L}_{A}\in\mathbb{R}^3$, which is related to $\mathbf{L}_{CoM}$ through $\mathbf{L}_{A} = \mathbf{L}_{CoM} + \mathbf{p}_{AC} \times (m\mathbf{v}_{CoM})$.
Thus, we have:
\begin{equation}
    \mathbf{L}_S = \mathbf{L}_{A}
    +
    \mathbf{p}_{SA}\times (m\mathbf{v}_{CoM}).
    \label{Eq: LA and LS}
\end{equation}
Note that $\mathbf{p}_{SA}$ is a zero vector since $A$ and $S$ coincide at the given time, but its time derivative is not necessarily zero due to the nontrivial movement of $S$ in the world frame.

\vspace{-0.1 in}
\subsection{Continuous-Phase Dynamics of ALIP-DRS}

During the continuous phase of bipedal walking, one foot contacts the ground while the other swings in the air.

\subsubsection{Dynamics of contact angular momentum $\mathbf{L}_S$}
To model the dynamics of $\mathbf{L}_S$, we first derive that for $\mathbf{L}_A$.
As $\dot{\mathbf{L}}_A$ equals the sum of the external moments about point $A$, we have:
\begin{equation}
\label{eq:dLA2}
\dot{\mathbf{L}}_A = \mathbf{p}_{AC} \times (m \mathbf{g}) + \boldsymbol{\tau}_A
=  \mathbf{p}_{SC} \times (m \mathbf{g}) + \boldsymbol{\tau}_A
,
\end{equation}
where $\mathbf{g} = [0,0,-g]^{\intercal}$ is the gravitational acceleration with magnitude $g$
and $\boldsymbol{\tau}_A=[\tau_{A,x},\tau_{A,y},\tau_{A,z}]^{\intercal}\in\mathbb{R}^3$ is the external torque that is applied to the contact point.

Next, 
we take the time derivative of both sides of~\eqref{Eq: LA and LS}, yielding $\dot{\mathbf{L}}_S = \dot{\mathbf{L}}_A+ \dot{\mathbf{p}}_{SA}\times (m\mathbf{v}_{CoM})+\mathbf{p}_{SA}\times (m\dot{\mathbf{v}}_{CoM})$.
Since $\mathbf{p}_{SA} = \mathbf{0}$ with $\mathbf{0}$ a zero verctor with an appropriate dimension and $\dot{\mathbf{p}}_{SA} =- \dot{\mathbf{p}}_{S}$ with $\mathbf{p}_S=[x_S,y_S,z_S]^{\intercal}\in\mathbb{R}^3$ the absolute position of the point $S$, this equation becomes:
\begin{equation}
\label{eq:dLS2}
    \dot{\mathbf{L}}_S = \dot{\mathbf{L}}_A- \dot{\mathbf{p}}_{S}\times (m\mathbf{v}_{CoM}).
\end{equation}

Combining~\eqref{eq:dLA2} and~\eqref{eq:dLS2} gives the dynamics equation of $\mathbf{L}_S$: 
\begin{equation}
\label{Eq: dLS3}
\dot{\mathbf{L}}_S = 
\mathbf{p}_{SC} \times (m \mathbf{g})+\boldsymbol{\tau}_A
-
\dot{\mathbf{p}}_{S}\times (m\mathbf{v}_{CoM}).
\end{equation}

Here we assume that
the CoM velocity is parallel to the DRS velocity at point $S$.
This helps prevent leg overstretch during surface motion, and
simplifies the robot model by leading to $\dot{\mathbf{p}}_{S}\times (m\mathbf{v}_{CoM})=\mathbf{0}$.
Thus,~\eqref{Eq: dLS3} becomes: 
\begin{equation}
\label{Eq: dLS4}
\dot{\mathbf{L}}_S = 
\mathbf{p}_{SC} \times (m \mathbf{g})+\boldsymbol{\tau}_A.
\end{equation}

As the toe joints of Digit's support foot are disabled to simplify the treatment of foot-ground contact as reviewed in Sec.~\ref{section: introduction}, we have $\tau_{A,x}=\tau_{A,y}=0$.
Then, the scalar form of~\eqref{Eq: dLS4} in $x$- and $y$-directions is given by:
\begin{equation}
\label{equ:dLS4 scalar}
    \begin{bmatrix}
        \dot{L}_{x,S}
        \\
        \dot{L}_{y,S}
    \end{bmatrix}
    =
    \begin{bmatrix}
        -mgy_{SC}
        \\
        mgx_{SC}
    \end{bmatrix}.
\end{equation}

\subsubsection{Dynamics of CoM position ${\mathbf{p}}_{SC}$}

By definition, the relative CoM velocity $\dot{\mathbf{p}}_{SC}$ is $\dot{\mathbf{p}}_{SC}
    =
    \mathbf{v}_{CoM}
    -
    \dot{\mathbf{p}}_{S}$.
Under the assumption that the CoM velocity is parallel to the DRS velocity at $S$ and given that the DRS only moves horizontally, we have $\dot{z}_S=0$ and $\dot{z}_{SC}=0$.
By expressing $\mathbf{v}_{CoM}$ using~\eqref{Eq: LS and LCOM},
the dynamics of the horizontal CoM position is:
\begin{equation}
\label{equ:dP_SC scalar}
    \begin{bmatrix}
        \dot{x}_{SC}
        \\
        \dot{y}_{SC}
    \end{bmatrix}
    =
    \begin{bmatrix}
        \frac{1}{mH}L_{y,S}
        \\
        \frac{-1}{mH}L_{x,S}
    \end{bmatrix}
    -
    \begin{bmatrix}
        \dot{x}_S(t)
        \\
        \dot{y}_S(t)
    \end{bmatrix}
    -
    \begin{bmatrix}
        \frac{1}{mH}L_{y,CoM}
        \\
        \frac{-1}{mH}L_{x,CoM}
    \end{bmatrix},
\end{equation}
where $L_{x,CoM}$ and $L_{y,CoM}$ are $L_{CoM}$'s horizontal elements.

\subsubsection{Continuous-phase ALIP-DRS dynamics}

As robots do not typically demonstrate significant angular momentum about the CoM during walking, we consider $\mathbf{L}_{CoM}=\mathbf{0}$~\cite{gong2022zero}.
Then, based on~\eqref{equ:dLS4 scalar} and~\eqref{equ:dP_SC scalar}, the ALIP-DRS dynamics in the sagittal and frontal planes are respectively expressed as:
\begin{equation}
\label{equ:ALIPDRS}
\underbrace{
    \begin{bmatrix}
        \dot{x}_{SC}
        \\
        \dot{L}_{y,S}
    \end{bmatrix}
    }_{=:\dot{\mathbf{x}}}
    =
    \underbrace{
    \begin{bmatrix}
        0 & \frac{1}{mH}
        \\
        mg & 0
    \end{bmatrix}
    }_{=:\mathbf{A}_x}
    \underbrace{
    \begin{bmatrix}
        {x}_{SC}
        \\
        {L}_{y,S}
    \end{bmatrix}
    }_{=:\mathbf{x}}
    +
    \underbrace{
    \begin{bmatrix}
       - \dot{x}_{S}(t)
        \\
        0
    \end{bmatrix}
    }_{=:\mathbf{f}_x(t)}
    ~\text{and}~
\end{equation}
\begin{equation}
\label{equ:ALIPDRS_lateral}
\underbrace{
    \begin{bmatrix}
        \dot{y}_{SC}
        \\
        \dot{L}_{x,S}
    \end{bmatrix}
    }_{=:\dot{\mathbf{y}}}
    =
    \underbrace{
    \begin{bmatrix}
        0 & -\frac{1}{mH}
        \\
        -mg & 0
    \end{bmatrix}
    }_{=:\mathbf{A}_y}
    \underbrace{
    \begin{bmatrix}
        {y}_{SC}
        \\
        {L}_{x,S}
    \end{bmatrix}
    }_{=:\mathbf{y}}
    +
    \underbrace{
    \begin{bmatrix}
        - \dot{y}_{S}(t)
        \\
        0
    \end{bmatrix}
    }_{=:\mathbf{f}_y(t)}.
\end{equation}

\begin{figure}[t]
    \centering
    \includegraphics[width=1\linewidth]{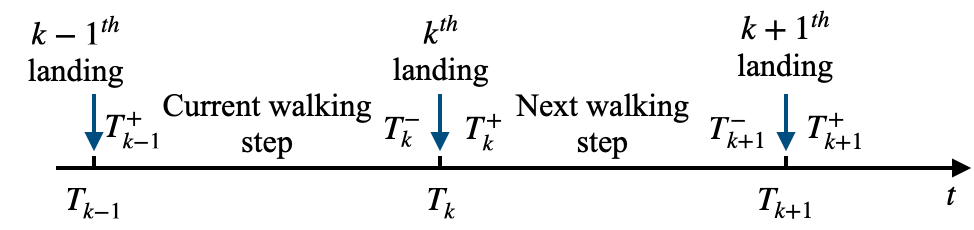}
    \vspace{-0.2 in}
    \caption{Illustration of foot-landing time instants. $T_{k-1}^+$ and $T_k^-$ are the start and end time instants of the current walking step, respectively. The next walking step begins at $t=T_{k}^+$ and ends at $T_{k+1}^-$.}
    \label{fig:time_line}
    \vspace{-0.2 in}
\end{figure}

\begin{rmk}[\textbf{Periodic DRS sway}]
This study considers DRS motions $\dot{x}_S(t)$ and $\dot{y}_S(t)$ that are continuously differentiable and periodic with $\dot{x}_S(t)=\dot{x}_S(t+T_{x,DRS})$ and $\dot{y}_S(t)=\dot{y}_S(t+T_{y,DRS})$ for any $t>0$. Here $T_{x,DRS}$ and $T_{y,DRS}$ are the least periods.
To highlight the explicit time dependence of $\dot{x}_S(t)$ and $\dot{y}_S(t)$, their expressions keep the time argument.
\label{rmk1}
\end{rmk}

Note that the continuous-phase ALIP-DRS model in~\eqref{equ:ALIPDRS} and~\eqref{equ:ALIPDRS_lateral} is explicitly time-varying and nonhomogeneous due to the presence of the time-varying forcing terms $\mathbf{f}_x(t)$ and $\mathbf{f}_y(t)$. These terms are induced by the horizontal DRS motion $\dot{x}_{S}(t)$ and $\dot{y}_{S}(t)$. This time-varying and nonhomogeneous property is fundamentally different from existing LIP models,  which are further discussed later in Remark~\ref{rmk2}.

\vspace{-0.1 in}
\subsection{Discrete Foot-Switching Model of ALIP-DRS}

To derive the time evolution of the CoM position ($x_{SC}$, $y_{SC}$) and contact angular momentum ($L_{x,S}$, $L_{y,S}$) across a foot-landing event, we consider the $k^{th}$ landing event at the time instant $T_{k}^-$ for any $k \in \mathbb{N}$, as illustrated in Fig.~\ref{fig:time_line}.
We use $(\cdot)^-$ and $(\cdot)^+$ to denote the values of $(\cdot)$ just before and after a foot-landing instant, respectively.

\subsubsection{Impact invariance of contact angular momentum ${\mathbf{L}}_S$}

Upon completing one walking step, the swing foot touches the walking surface while the old support foot begins to swing.
Thus, there is a sudden change in the contact-point position, as illustrated in Fig.~\ref{fig:ALIP state}. 

Let $\mathbf{L}_{S,k}$ denote the value of the angular momentum $\mathbf{L}_S$ about the $k^{th}$ contact point. This contact point is initiated at the $(k-1)^{th}$ landing at $T_{k-1}^-$.
Let $\mathbf{L}_{S,k+1}$ represent the angular momentum about the new $(k+1)^{th}$ contact point initiated at $T_{k}^-$.
At $T_{k}^-$, $\mathbf{L}_{S,k+1}^-$ and $\mathbf{L}_{S,k}^-$  are related through:
\begin{equation}
\label{eq:AM transfer}
    \mathbf{L}^-_{S,k+1} = \mathbf{L}^-_{S,k}+\mathbf{p}_{(k+1)\rightarrow k}\times (m\mathbf{v}^-_{CoM}),
\end{equation}
where $\mathbf{p}_{(k+1)\rightarrow k}$ is the position vector pointing from the $(k+1)^{th}$ to the $k^{th}$ contact point.

As the robot walks on a horizontal terrain with zero vertical CoM velocity (i.e., $\dot{z}_{SC}=0$)~\cite{gong2022zero}, we have $\mathbf{p}_{(k+1)\rightarrow k}\times (m\mathbf{v}^-_{CoM})=\mathbf{0}$, leading to $\mathbf{L}^-_{S,k+1} = \mathbf{L}^-_{S,k}$.
Meanwhile, since the $k^{th}$ landing impact results in zero impulse torque about the $(k+1)^{th}$ contact point, $\mathbf{L}_{S,k+1}$ remains unchanged across the $k^{th}$ landing event; that is, $\mathbf{L}^+_{S,k+1}=\mathbf{L}^-_{S,k+1}$.
Thus, the robot's contact angular momentum $\mathbf{L}_S$ is invariant to the landing impact, with $\mathbf{L}^+_{S,k+1} = \mathbf{L}^-_{S,k}$. 
Then, the changes in the contact angular momenta across a foot-landing event satisfy $\Delta L_{y,S} := L_{y,S}^+ - L_{y,S}^- = 0$ and $\Delta L_{x,S} := L_{x,S}^+ - L_{x,S}^- = 0$.

\subsubsection{Jump in relative CoM position}
At a foot-landing event, the horizontal CoM position (${x}_{SC}$,~${y}_{SC}$) jumps because the horizontal contact-point position (${x}_S$,~${y}_S$) jumps (Fig.~\ref{fig:ALIP state}). 
We use $\Delta x_{SC}$ and $\Delta y_{SC}$ to denote such changes in $x_{SC}$ and $y_{SC}$, respectively; that is,
$\Delta x_{SC} := x_{SC}^+ - x_{SC}^-$
and 
$\Delta y_{SC} := y_{SC}^+ - y_{SC}^-$.

Using $u_x$ and $u_y$ to denote the step lengths along the $x$- and $y$-directions, we obtain the discrete open-loop dynamics of $x_{SC}$ and $y_{SC}$ as:
$\Delta x_{SC} = -u_x$ and $\Delta y_{SC}=-u_y$.
Note that $\mathbf{p}_{(k+1)\rightarrow k}=[-u_x,-u_y,0]^\intercal$ for a horizontal, swaying DRS.

Here we treat the footstep position $(u_x,u_y)$ as the discrete control input to the proposed ALIP-DRS model, to be designed later.
This treatment allows us to explicitly and systematically consider foot placement in the stability analysis and control of the ALIP-DRS model.

\subsubsection{Discrete dynamics}

In summary, the discrete portion of the open-loop ALIP-DRS dynamics is expressed as:
\begin{equation}
\label{equ: delta X_}
\begin{aligned}
    \underbrace{
    \begin{bmatrix}
        \Delta x_{SC}
        \\
        \Delta L_{y,S}
    \end{bmatrix}
    }_{=:\Delta \mathbf{x}}
    &=
    \underbrace{
    \begin{bmatrix}
    -1
        \\
        0
    \end{bmatrix}
    }_{=:\mathbf{b}_x}
    u_x
\end{aligned}
~\text{and}~~~~
\begin{aligned}
    \underbrace{
    \begin{bmatrix}
        \Delta y_{SC}
        \\
        \Delta L_{x,S}
    \end{bmatrix}
    }_{=:\Delta \mathbf{y}}
    &=
    \underbrace{
    \begin{bmatrix}
    -1
        \\
        0
    \end{bmatrix}
    }_{=:\mathbf{b}_y}
    u_y.
\end{aligned}
\end{equation}

\vspace{-0.1 in}
\subsection{Open-loop Hybrid ALIP-DRS Model}

From the continuous-phase dynamics in~\eqref{equ:ALIPDRS} and~\eqref{equ:ALIPDRS_lateral} and the discrete one in~\eqref{equ: delta X_},
we know the sagittal and frontal subsystems of the ALIP-DRS share the same structure.
For brevity and without loss of generality, we primarily focus on the sagittal-plane subsystem in the controller design and stability analysis for the ALIP-DRS model.
Combining the sagittal components of those three equations gives the state-space form of the open-loop ALIP-DRS system as:
\begin{equation}
    \begin{cases}
    \begin{aligned}
&\dot{\mathbf{x}} = \mathbf{A}_x \mathbf{x} + \mathbf{f}_x(t),&\text{if}~t \neq T_k^-;
\\
&\Delta \mathbf{x}
= \mathbf{b}_x u_x, &\text{if}~t = T_k^-.
\end{aligned}
    \end{cases}
    \label{eq: open loop ALIP-DRS}
\end{equation}

\begin{rmk}[\textbf{Time-varying and nonhomogeneous ALIP-DRS}]
Equation~\eqref{eq: open loop ALIP-DRS} reveals that the open-loop ALIP-DRS dynamics is linear, hybrid, time-varying, and nonhomogeneous. 
The ALIP-DRS is fundamentally different from the existing LIP~\cite{kajita20013d}, H-LIP~\cite{xiong20223}, and ALIP~\cite{gong2022zero}, which are time-invariant and nonhomogeneous since they consider static terrains.
With a stationary ground, $\mathbf{f}_x (t)=\mathbf{0}$ holds, and ALIP-DRS's continuous part reduces to the existing ALIP.
Further, the ALIP-DRS is distinct from the existing time-varying LIP models for vertically moving surface~\cite{iqbal2023analytical,iqbal2023asymptotic} and variable CoM height~\cite{caron2019capturability}, which are all homogeneous.
    \label{rmk2}
\end{rmk}

\section{ALIP-DRS Footstep Control and Stability Analysis}
\label{sec: ALIP control and stability}

This section introduces the discrete footstep control and stability analysis for the ALIP-DRS.
The controller serves as the higher-layer planner of a hierarchical control framework, which indirectly stabilizes the unactuated dynamics of an underactuated walking robot, as detailed in Sec.~\ref{section: Hierarchical Planning and Control Framework}.


The design of this controller builds upon the existing footstep controller~\cite{gong2022zero}, adapting it from the classical ALIP model to the proposed ALIP-DRS model, as discussed in subsection A. Given that the ALIP-DRS is time-varying and nonhomogeneous in contrast to the time-invariant and homogeneous nature of the classical ALIP, the previous stability conditions are not suitable for ALIP-DRS. Thus, new conditions are introduced in subsections B and C.

\vspace{-0.1in}
\subsection{Discrete-Time Foot Placement Control}

The objective of the discrete-time footstep control is to find the expression of the foot-landing position $u_x$ such that the hybrid ALIP-DRS system in \eqref{eq: open loop ALIP-DRS} is provably stabilized.

The key idea of the controller is by adjusting the swing-foot landing location at the end of the current step (e.g., $t=T_K^-$), the contact angular momentum at the end of the next step (e.g., $t=T_{K+1}^-$) can be regulated. These time instants are illustrated in Fig.~\ref{fig:time_line}.

\subsubsection{Solution to the continuous-phase ALIP-DRS dynamics}

From the linear system theory, the solution of~\eqref{equ:ALIPDRS} is:
\begin{equation}
\begin{aligned}
\label{equ:linear system solution}
    \begin{bmatrix}
        x_{SC}(t_2)
        \\
        L_{y,S}(t_2)
    \end{bmatrix}
    &=
    e^{\mathbf{A}_x(t_2-t_1)}
    \begin{bmatrix}
        x_{SC}(t_1)
        \\
        L_{y,S}(t_1)
    \end{bmatrix}
    +
    \begin{bmatrix}
        V_{x,1} \\
        V_{x,2}
    \end{bmatrix}
\end{aligned}
\end{equation}
on $t \in [t_1, t_2]$, for any $t_1,t_2 >0$. Here, {\small $[V_{x,1}(t_1,t_2), V_{x,2}(t_1,t_2)]^\intercal:=\mathbf{V}_x(t_1,t_2):=\int_{t_1}^{t_2}e^{\mathbf{A}_x(t_2-\tau)}\mathbf{f}_x(\tau)d\tau$}.

This study considers known DRS motion, and thus $\mathbf{f}_x(t)$ is known for any $t>0$.
Then, since $\mathbf{f}_x(t)$ is known, the value of $\mathbf{V}_x(t_1,t_2)$ can be readily computed for any $t_1,t_2 >0$. 

\subsubsection{Expressing pre-impact contact angular momentum at $t=T_{k+1}^-$}

From the second row of~\eqref{equ:linear system solution}, the contact angular momentum at the end of the next step, $L_{y,S}(T_{k+1}^-)$, is related to that at the beginning of the next step, $L_{y,S}(T_k^+)$, through:
\begin{equation}
\label{equ:linear system solution row 2 T_k T_k+1}
\begin{aligned}
L_{y,S}(T_{k+1}^-)&=mHl\sinh(lT_{step})x_{SC}(T_k^+)
\\
&+\cosh(lT_{step})L_{y,S}(T_k^+)
    +V_{x,2}(T_k^+,T_{k+1}^-),
\end{aligned}
\end{equation}
where $l:=\sqrt{\tfrac{g}{H}}$ and $T_{step} := T_{k+1}-T_{k}$ is the step duration. 

The role switching of the swing and support feet leads to:
\vspace{-0.05 in}
\begin{equation}
\label{equ:foot switch}
    x_{SC}(T_k^+) = x_{SwC}(T_k^-),
\end{equation}
where $x_{SwC}$ is the forward CoM position relative to the swing foot.
Meanwhile, since contact angular momentum is impact invariant as discussed in Sec.~\ref{Section: Preliminaries}, we have:
\vspace{-0.05 in}
\begin{equation}
\label{equ: impact invariant}
    L_{y,S}(T_k^+)=L_{y,S}(T_k^-).
\end{equation}
Then, combining~\eqref{equ:linear system solution row 2 T_k T_k+1}-\eqref{equ: impact invariant} gives:
\begin{equation}
\begin{aligned}
    \label{equ:linear system solution row 2 T_k T_k+1 combined}
    L_{y,S}(T_{k+1}^-)&=mHl\sinh(lT_{step})x_{SwC}(T_k^-)
    \\
    &+\cosh(lT_{step})L_{y,S}(T_k^-)
    +V_{x,2}(T_k^+,T_{k+1}^-).
\end{aligned}
\end{equation}

\subsubsection{Matching the predicted and desired contact angular momenta at $T_{K+1}^-$}

Let $\bar{L}_{y,S}$ denote the desired contact angular momentum in the sagittal plane, which can be treated as a user input.
Based on~\eqref{equ:dP_SC scalar} and given the desired forward ALIP-DRS velocity $v_x^{des}$, $\bar{L}_{y,S}$ can be specified as: $\bar{L}_{y,S} = mH v_x^{des}$.

From~\eqref{equ:linear system solution row 2 T_k T_k+1 combined}, we know that to ensure $L_{y,S}(T_{k+1}^-) = \Bar{L}_{y,S}(T_{k+1}^-)$, the swing-foot landing position at the end of the current step $T_k^-$ should be:
\begin{equation}
\small
    \label{equ:swing foot user input}
    x_{SwC}(T_k^-)
    = \frac{\bar{L}_{y,S}(T_{k+1}^-)-V_{x,2}(T_k^+,T_{k+1}^-)-\cosh(lT_{step})L_{y,S}(T_k^-)}{mHl\sinh(lT_{step})}.
\end{equation}

\begin{figure}[t]
    \centering\includegraphics[width=1\linewidth]
{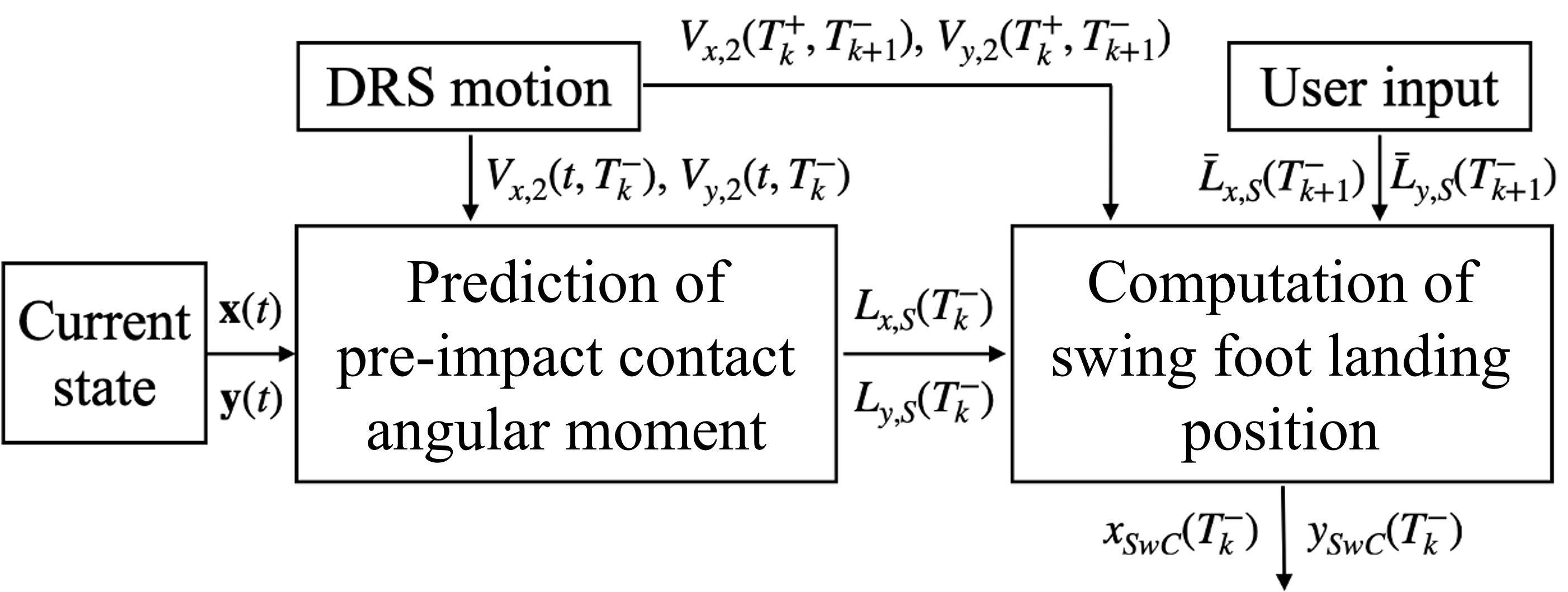}
    \vspace{-0.3 in}
    \caption{Computation of the swing-foot landing location for the end of the current walking step at $T_{k}^-$. The ALIP-DRS state at the current time step $t$ ($\mathbf{x}(t)$ and $\mathbf{y}(t)$) and the nominal DRS motion profile ($x_S(t)$, $y_S(t)$) are used to compute the estimated pre-impact contact angular momentum ($L_{x,S}(T_{k}^-)$, $L_{y,S}(T_{k}^-)$) based on~\eqref{equ:linear system solution} and~\eqref{equ:linear system solution lateral}. This prediction and the corresponding user-specified desired angular momentum ($\bar{L}_{x,S}(T_{k+1}^-)$, $\bar{L}_{y,S}(T_{k+1}^-)$) are used to compute the swing-foot placement ($x_{SwC}(T_k^-)$, $y_{SwC}(T_k^-)$). Here $V_{y,2}$, $\bar{L}_{y,S}$, and $y_{SwC}$ are counterparts of $V_{x,2}$, $\bar{L}_{x,S}$, and $x_{SwC}$ for the frontal plane.}
    \label{fig:swing_foot_compute}
    \vspace{-0.1 in}
\end{figure}

In~\eqref{equ:swing foot user input}, $V_{x,2}(T_k^+,T_{k+1}^-)$ can be directly computed as mentioned earlier.
The terms $\cosh(lT_{step})$ and $mHl\sinh(lT_{step})$ are known since $l$, $T_{step}$, $m$, and $H$ are known parameters of ALIP-DRS.
Also, $L_{y,S}(T_K^-)$ can be computed based on the second row of~\eqref{equ:linear system solution} with $t_1 = T_{k-1}$ and $t_2 = T_k$.
Thus, the forward foot-landing position at the end of the current step, $x_{SwC}(T_k^-)$, can be readily calculated.
The procedure for computing $x_{SwC}(T_k^-)$ is illustrated in Fig.~\ref{fig:swing_foot_compute}.

\subsubsection{Expressing discrete-time footstep control law $u_x$}

Recall $u_x(T_k^-) = x_{SC}(T_k^-) - x_{SwC}(T_k^-)$ (see Fig.~\ref{fig:ALIP state}). Then, the discrete footstep control law at the $k^{th}$ landing is: 
\vspace{-0.05 in}
{\small 
\begin{equation}
\begin{aligned}
    u_x(T_k^-) 
        &= x_{SC}(T_k^-) \\
        &- \frac{\bar{L}_{y,S}(T_{k+1}^-)-V_{x,2}(T_k^+,T_{k+1}^-)-\cosh(lT_{step})L_{y,S}(T_k^-)}{mHl\sinh(lT_{step})}.
\end{aligned}
\label{eq: ux}
\end{equation}}

\vspace{-0.1 in}
\subsection{Closed-loop ALIP-DRS model}

Combining~\eqref{equ:ALIPDRS},~\eqref{equ: delta X_}, and \eqref{eq: ux} yields the following closed-loop ALIP-DRS system:
\begin{equation}
\label{equ:hybrid system ALIP}
\begin{cases}
    \begin{aligned}
        \dot{\mathbf{x}} &= \mathbf{A}_x\mathbf{x} + \mathbf{f}_x(t), &t\neq T_k^- ;
        \\
        \Delta \mathbf{x} &= \mathbf{B}_x\mathbf{x}^-+\mathbf{g}_x (t), &t = T_k^- ,
    \end{aligned}
\end{cases}
\end{equation}
where $$
\footnotesize
\mathbf{B}_x = 
    \begin{bmatrix}
        -1 & -\tfrac{\cosh(lT_{step})}{mHl\sinh(lT_{step})}
        \\
        0 & 0
    \end{bmatrix}
    ~\text{\normalsize and}~
    \mathbf{g}_x(T_k^-) = 
    \begin{bmatrix}
    \tfrac{\bar{L}_{y,S}(T_{k+1}^-)-V_{x,2}(T_{k}^+,T_{k+1}^-)}{mHl\sinh(lT_{step})}
        \\
        0
    \end{bmatrix}.$$

\subsubsection{Periodic solution to the closed-loop ALIP-DRS system}

As the DRS of interest to this study sways periodically (Remark~\ref{rmk1}), we focus on periodic solutions of the closed-loop ALIP-DRS system in~\eqref{equ:hybrid system ALIP}, which corresponds to cyclic walking motions.
We use $\boldsymbol{\psi}_x(t)$ to denote the periodic solution with a least period of $T_{sys}$, i.e., $\boldsymbol{\psi}_x(t)=\boldsymbol{\psi}_x(t+T_{sys})$ for any $t>0$.
To analyze the stability of $\boldsymbol{\psi}_x(t)$ under the proposed footstep control law, we introduce the homogeneous system associated with the ALIP-DRS in~\eqref{equ:hybrid system ALIP}.

\subsubsection{Homogeneous system associated with ALIP-DRS}
To analyze the stability of linear time-varying nonhomogeneous hybrid systems, which include the closed-loop ALIP-DRS system in \eqref{equ:hybrid system ALIP}, we need to consider their corresponding homogeneous systems~\cite{bainov1993impulsive}.
The homogeneous system associated with \eqref{equ:hybrid system ALIP} is given as:
\vspace{-0.05 in}
\begin{equation}
\label{equ:hybrid system ALIP homogeneous}
\begin{cases}
    \begin{aligned}
        \mathbf{\dot{z}} &= \mathbf{A}_x\mathbf{z}^-, &t\neq T_k^-;
        \\
        \Delta \mathbf{z} &= \mathbf{B}_x\mathbf{z}^-, &t = T_k^-,
    \end{aligned}
\end{cases}
\end{equation}
where $\mathbf{z}$ is the state vector of the homogeneous system.

\subsubsection{Monodromy matrix under the special case where $T_{x,DRS} = T_{step}=T_{sys}$}

When the least periods of the DRS motion and the periodic solution are equal to the duration of one walking step (i.e., $T_{x,DRS} = T_{sys} = T_{step}$), the monodromy matrix~\cite{bainov1993impulsive} of the homogeneous system, denoted as $\mathbf{M}_x$, is expressed as: 
\begin{equation}
\begin{aligned}
    \mathbf{M}_x &= (\mathbf{I}+\mathbf{B}_x)e{^{\mathbf{A}_xT_{step}}}
    \\
    & = 
    \begin{bmatrix}
        -\cosh(lT_{step}) & -\frac{\cosh^2(lT_{step})}{mHl\sinh(lT_{step})}
        \\
        mHl\sinh(lT_{step}) & \cosh(lT_{step})
    \end{bmatrix},
\end{aligned}
\label{Mx}
\end{equation}
where $\mathbf{I}$ is an identity matrix with an appropriate dimension.

The eigenvalues of $\mathbf{M}_x$, denoted as $\lambda_{x,1}$ and $\lambda_{x,2}$, can be calculated as: $\lambda_{x,1} =\lambda_{x,2} =0$.

\begin{figure}[t]
    \centering
    \includegraphics[width=0.8\linewidth]{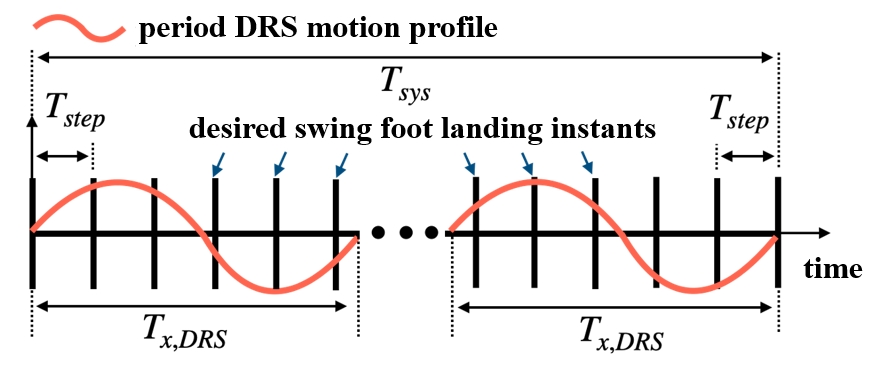}
    \vspace{-0.2 in}
    \caption{Illustration of the periods of the walking step ($T_{step}$), forward DRS motion ($T_{DRS}$), and solution of the closed-loop ALIP-DRS system ($T_{sys}$). This study considers both the special case where $T_{sys}=T_{step}=T_{x,DRS}$ and the general case where $T_{sys}=N_1 T_{step} = N_2 T_{x,DRS}$ for any $N_1,N_2 \in \mathbb{N}$.}
    \label{fig:my_label}
    \vspace{-0.2 in}
\end{figure}

\begin{rmk}[\textbf{Interpretation of the monodromy matrix}]
\label{rmk 5}
Based on the ALIP-DRS equation in~\eqref{equ:hybrid system ALIP} and the definition of the monodromy matrix $\mathbf{M}_x$ in~\eqref{Mx},
 $\mathbf{M}_x$ relates the post-impact state values $\mathbf{x}(T_{k-1}^+)$ and $\mathbf{x}(T_{k}^+)$ as:
\begin{equation}
\label{eq: S2S_dynamics_}
\textbf{x}(T_{k}^+) = \mathbf{M}_x\textbf{x}(T_{k-1}^+) +  \tilde{\mathbf{V}}_x(T_{k-1}^+, T_{k}^-),
\end{equation}
where $\tilde{\mathbf{V}}_x:=(\mathbf{I}+\mathbf{B}_x) \mathbf{V}_x(T_{k-1}^+, T_{k}^-)$.
Equation~\eqref{eq: S2S_dynamics_} is the step-to-step dynamics~\cite{xiong20223} of the ALIP-DRS.
When the ground is static, $\tilde{\mathbf{V}}_x(T_{k-1}^+, T_{k}^-)$ becomes zero, and~\eqref{eq: S2S_dynamics_} reduces to the step-to-step dynamics of the ALIP for static terrain.
\end{rmk}

\begin{figure*}[t]
    \centering
    \includegraphics[width=0.9\linewidth]{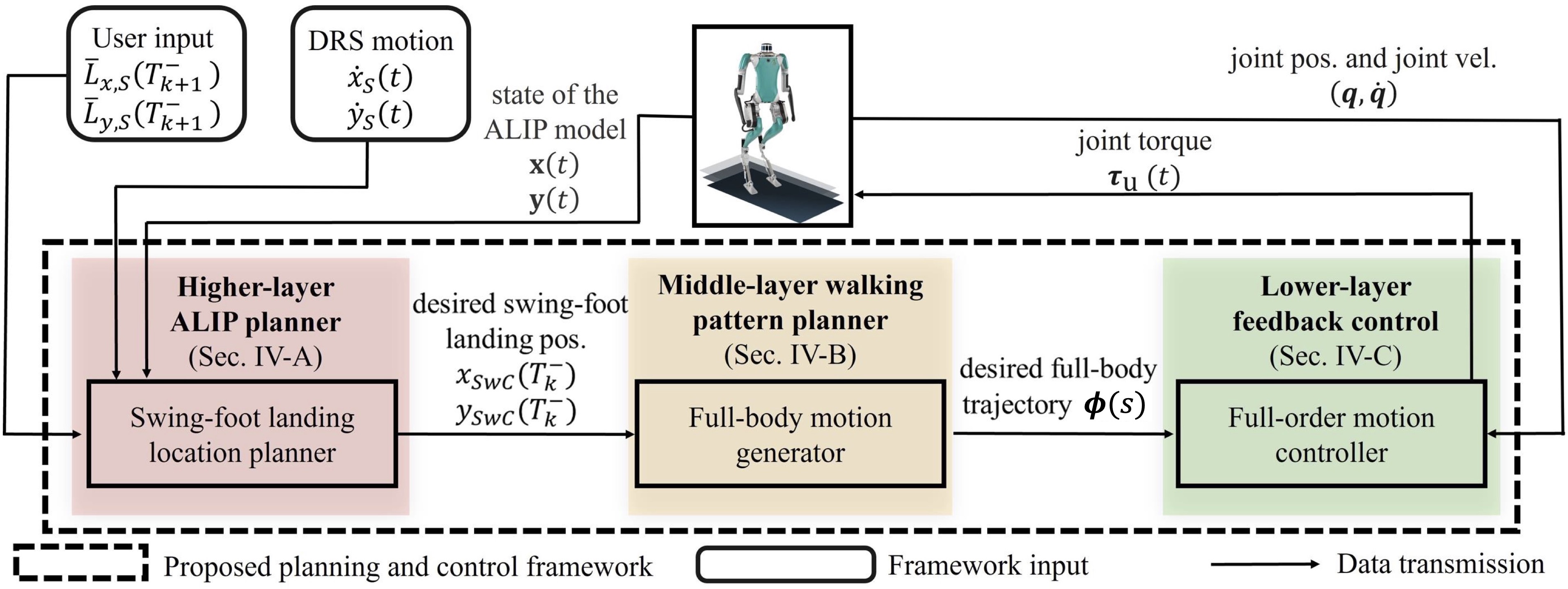}
    \vspace{-0.05 in}
    \caption{Overview of the proposed hierarchical control approach.
    The higher-layer planner generates the desired swing-foot landing locations ($x_{SwC}$, $y_{SwC}$) based on the ALIP-DRS model to help stabilize the unactuated robot dynamics.
    Subsequently, the middle-layer walking pattern generator plans the desired trajectories $\boldsymbol{\phi}(s)$ for the actively actuated joints, ensuring they are compatible with both the ALIP-DRS model and the desired swing-foot landing positions.
    Lastly, the lower-layer controller outputs the joint torques $\boldsymbol{\tau}_u$ to reliably track the full-body trajectories generated by the middle layer.}
    \label{fig:framework_overview}
    \vspace{-0.1 in}
\end{figure*}

\subsubsection{Monodromy matrix under the general case where $N_1T_{step}=N_2T_{x,DRS}=T_{sys}$}
The rest of this paper considers the general case where $N_1T_{step}=N_2T_{x,DRS}=T_{sys}$ for any $N_1,N_2 \in \mathbb{N}$.
Under this general case, the ratio between the DRS motion period and one walking-step duration can be any real positive number, instead of one as in the special case.
Figure~\ref{fig:my_label} illustrates $T_{step}$, $T_{x,DRS}$, and $T_{sys}$.

Using~\eqref{eq: S2S_dynamics_} recursively, we obtain the following step-to-step dynamics of the closed-loop ALIP-DRS system:
\begin{equation}
\label{equ: respect impact not so much}
\small
\begin{aligned}
    \textbf{x}(T_{k+N_1}^+) &=\mathbf{M}_x\mathbf{x}(T_{k+N_1-1}^+) + \tilde{\mathbf{V}}_x(T_{k+N_1-1}^+, T_{k+N_1}^-) \\
    =& \mathbf{M}_x \left[ \mathbf{M} _x\textbf{x}(T_{k+N_1-2}^+)  + \tilde{\mathbf{V}}_x(T_{k+N_1-2}^+, T_{k+N_1-1}^-) \right]
    \\
    &+ \tilde{\mathbf{V}}_x(T_{k+N_1-1}^+, T_{k+N_1}^-) \\
    =& \mathbf{M}_x^2 \textbf{x}(T_{k+N_1-2}^+) + \mathbf{M}_x \tilde{\mathbf{V}}_x(T_{k+N_1-2}^+, T_{k+N_1-1}^-)
    \\
    &+\tilde{\mathbf{V}}_x(T_{k+N_1-1}^+, T_{k+N_1}^-)
    \\
    =&...
    = 
    \underbrace{
    \mathbf{M}_x^{N_1}}_{\bar{\mathbf{M}}_x}\textbf{x}(T_{k}^+) 
    + \sum_{i=1}^{N_1} \mathbf{M}_x^{N_1-i} \tilde{\mathbf{V}}(T_{k + i-1}^+, T_{k+i}^-).
\end{aligned}
\end{equation}

Equation \eqref{equ: respect impact not so much} shows that the monodromy matrix $\Bar{\mathbf{M}}_x$ of the homogeneous system for this general scenario is related to ${\mathbf{M}}_x$ through: $\bar{\mathbf{M}}_x = \mathbf{M}_x^{N_1}$.
Thus, the two eigenvalues of $\bar{\mathbf{M}}_x$, denoted as $\bar{\lambda}_{x,1}$ and $\bar{\lambda}_{x,2}$, can be evaluated as:
\begin{equation}
    \bar{\lambda}_{x,1} = \lambda^{N_1}_{x,1} = 0 ~\text{and}~
    \bar{\lambda}_{x,2} = \lambda^{N_1}_{x,2} = 0 .
    \label{eq: lambda12-N}
\end{equation}

\subsubsection{Stability of homogeneous system}
With the properties of the monodromy matrix analyzed, we are now ready to introduce the stability condition for the homogeneous system under the general case where $N_1T_{step}=N_2T_{x,DRS}=T_{sys}$.

\begin{prop}[\textbf{Stability of homogeneous system under discrete footstep control}]
Under the general case where $N_1T_{step}=N_2T_{x,DRS}=T_{sys}$ with any $N_1,N_2 \in \mathbb{N}$,
the homogenous system in~\eqref{equ:hybrid system ALIP homogeneous} is exponentially stable under the proposed discrete footstep control law in~\eqref{eq: ux}.
\hfill $\diamondsuit$
\label{prop1}
\end{prop}

{\it Proof:}
From $T_{sys}=N_1 T_{step}$ and the definition of $T_{step}$ (i.e., $T_{step}=T_{k+1}-T_k$), we know $T_{sys}$ satisfies $T_{sys}=T_{k+N_1} - T_k$.
Also, by definition, the matrix $\mathbf{A}_x$ is constant.
Thanks to these properties, we can assess the stability of the homogeneous system based on the stability theorem of general linear time-invariant impulsive systems given in Theorem 3.5 of~\cite{bainov1993impulsive}.
By that theorem, because the eigenvalues $\bar{\lambda}_{x,1}$ and $\bar{\lambda}_{x,2}$ are strictly less than one in modulus under the proposed discrete control law in~\eqref{eq: ux}, the homogeneous system is exponentially stable under the control law.  
\hfill $\square$

With the stability of the homogeneous system confirmed in Proposition~\ref{prop1}, the exponential stability of the periodic solution $\boldsymbol{\psi}_x(t)$ for the nonhomogeneous closed-loop ALIP-DRS system in~\eqref{equ:hybrid system ALIP} is discussed next.

\vspace{-0.1 in}

\subsection{Closed-Loop Stability Analysis for ALIP-DRS}

The closed-loop stability analysis is conducted to verify the stability of the periodic solution  $\boldsymbol{\psi}_x(t)$ of the closed-loop ALIP-DRS system, as described in~\eqref{equ:hybrid system ALIP}, under the proposed footstep controller. 
The stability of $\boldsymbol{\psi}_x(t)$ is assessed by applying the stability condition for general linear time-varying nonhomogeneous hybrid systems, which include the closed-loop ALIP-DRS system in~\eqref{equ:hybrid system ALIP}.


\begin{thm}[\textbf{Exponential stability of ALIP-DRS under footstep control}]
Consider the general case where $N_1T_{step}=N_2T_{x,DRS}=T_{sys}$ with any $N_1,N_2 \in \mathbb{N}$.
Under the proposed discrete footstep control law in~\eqref{eq: ux},
the periodic solution $\boldsymbol{\psi}_x(t)$ of the nonhomogeneous hybrid ALIP-DRS system in \eqref{equ:hybrid system ALIP} is exponentially stable.
\hfill $\diamondsuit$
\label{thm1}
\end{thm}

{\it Proof:}
As confirmed in Proposition~\ref{prop1}, the homogeneous system in~\eqref{equ:hybrid system ALIP homogeneous} is exponentially stable under the proposed discrete footstep control law in~\eqref{eq: ux} for $N_1T_{step}=N_2T_{x,DRS}=T_{sys}$.
Then, by the stability theory of linear nonhomogeneous impulsive systems (Theorem 4.2 in~\cite{bainov1993impulsive}), the footstep controller ensures the exponential stability of $\boldsymbol{\psi}_x(t)$ for the nonhomogeous ALIP-DRS system in~\eqref{equ:hybrid system ALIP}.
\hfill $\square$

From~\eqref{eq: lambda12-N}, we know that the algebraic multiplicity of the eigenvalue $0$ is two while the geometric multiplicity is one.
Thus, any state $\mathbf{x}$ does not necessarily converge to the periodic solution $\boldsymbol{\psi}_x(t)$ within one period of $T_{sys}$ but is guaranteed to converge exactly to $\boldsymbol{\psi}_x(t)$ within $2T_{sys}$.


\section{Hierarchical Control Framework}

\label{section: Hierarchical Planning and Control Framework}

This section presents the proposed hierarchical control framework designed to achieve stable underactuated walking during DRS sway.
The key novelty of the framework lies in its utilization of the proposed ALIP-DRS footstep control law as a higher-layer footstep planner.

As shown in Fig.~\ref{fig:framework_overview}, the framework is structured into three layers to ensure the computational efficiency for real-time motion generation while guaranteeing stability for the complex full-order dynamics of underactuated humanoid walking on DRS.
Its higher-layer footstep planner uses the proposed ALIP-DRS footstep controller in~\eqref{eq: ux} to generate the desired footstep locations in real-time.

The middle layer produces the task-space reference trajectories for the full-order robot model that agree with the desired footstep positions supplied by the higher layer.
These trajectories conform to the simplifying assumptions of the ALIP-DRS model, such as parallel velocities for the CoM and the support point, which minimizes discrepancies between the reduced-order ALIP-DRS and the full-order robot model.
This alignment ensures that the desired footstep positions provided by the higher layer are physically feasible for the actual robot to follow, indirectly supporting the stabilization of the full-order underactuated robot model, as detailed in the next section.

The lower-layer controller ensures that the actual robot faithfully executes the planned full-body trajectories.

The overall stability of the complete full-order system under this hierarchical control framework is analyzed and verified in Sec.~\ref{Section: Stability}.

\subsection{Higher-Layer ALIP-DRS Footstep Planner}

This subsection introduces the higher-level planner, which generates the desired CoM and foot-landing positions based on the ALIP-DRS model.

The desired CoM position trajectories along the $x$- and $y$-directions are set as the solutions of the ALIP-DRS model in the sagittal and frontal planes, respectively.
The desired CoM position in the $x$-direction is given in~\eqref{equ:linear system solution}.
From the frontal ALIP-DRS model in~\eqref{equ:ALIPDRS_lateral}, its solution between [$t_1,t_2$] for any $t_1,t_2>0$ is:
\begin{equation}
\begin{aligned}
\label{equ:linear system solution lateral}
    \begin{bmatrix}
        y_{SC}(t_2)
        \\
        L_{x,S}(t_2)
    \end{bmatrix}
    =
    e^{\mathbf{A}_y(t_2-t_1)}
    \begin{bmatrix}
        y_{SC}(t_1)
        \\
        L_{x,S}(t_1)
    \end{bmatrix}
    +
    \begin{bmatrix}
        V_{y,1}(t_1,t_2) \\
        V_{y,2}(t_1,t_2)
    \end{bmatrix},
\end{aligned}
\end{equation}
where {\small $[V_{y,1}(t_1,t_2), V_{y,2}(t_1,t_2)]^\intercal:=\mathbf{V}_{y}(t_1,t_2):=\int_{t_1}^{t_2}e^{\mathbf{A}_y(t_2-\tau)}$$\mathbf{f}_y(\tau)d\tau$}.

The discrete footstep control law of the ALIP-DRS model is used to determine the desired foot-landing locations for the actual full-dimensional robot. 
This controller directly stabilizes the reduced-order ALIP-DRS model, thus contributing to the stabilization of the actual dynamics of the unactuated variables $\mathbf{x}$ and $\mathbf{y}$ as explained in Sec.~\ref{Section: Stability}.

The desired forward landing position is given by~\eqref{eq: ux}.
As the ALIP-DRS dynamics in the sagittal and frontal planes share the same structure,
the desired lateral landing location is planned in a way similar to the sagittal-plane footstep controller design introduced in Sec.~\ref{sec: ALIP control and stability}.
Analogous to~\eqref{equ:swing foot user input}, 
the lateral CoM location relative to the swing foot at the end of the current step is expressed as:
\begin{equation}
\small
    \label{equ:swing foot user input lateral}
    y_{SwC}(T_k^-) 
    = \frac{-\bar{L}_{x,S}(T_{k+1}^-)+V_{y,2}(T_k^+,T_{k+1}^-)+\cosh(lT_{step})L_{x,S}(T_k^-)}{mHl\sinh(lT_{step})},
\end{equation}
where $\bar{L}_{x,S}$ is the desired contact angular momentum.

Accordingly, the discrete footstep control law for the frontal plane, at the $k^{th}$ landing event, is given by: 
{\small \begin{equation}
\begin{aligned}
    u_y(T_k^-)
    =&y_{SC}(T_k^-) \\
    &- \frac{-\bar{L}_{x,S}(T_{k+1}^-)+V_{y,2}(T_k^+,T_{k+1}^-)+\cosh(lT_{step})L_{x,S}(T_k^-)}{mHl\sinh(lT_{step})}.
\end{aligned}
\label{eq: uy}
\end{equation}}

The value of $\bar{L}_{x,S}(T_{k+1}^-)$ can be determined based on the previous approach used for walking on static terrain~\cite{gong2021one}, which produces a zero average lateral speed.
By setting $\bar{L}_{x,S}(T_{k+1}^-)$ based on~\eqref{equ:ALIPDRS_lateral} to allow a periodic lateral motion with a desired step width $W$, $\bar{L}_{x,S}$ is given by:
\begin{equation}
    \label{equ:L_x desired}
    \bar{L}_{x,S} = 
    \begin{cases}
    \begin{aligned}
        &\tfrac{1}{2}mHW\tfrac{l\sinh(lT_{step})}{1+\cosh(lT_{step})}~\text{(right-foot-in-support)};
        \\
        -&\tfrac{1}{2}mHW\tfrac{l\sinh(lT_{step})}{1+\cosh(lT_{step})}~\text{(left-foot-in-support)}.
    \end{aligned}
\end{cases}
\end{equation}

\subsection{Middle-Layer Walking Pattern Planner}
\label{subsection: Middle-Layer Walking Pattern Planner}
This subsection explains the middle-layer planner, with the Digit robot as an illustrating example.
This layer aims to generate reference trajectories for the full-order robot model that are compatible with the ALIP-DRS model and align with the desired footstep locations determined by the higher-layer planner. These trajectories then serve as inputs to the lower-layer controller, as described in Subsection C.

\begin{figure}[t]
    \centering
    \includegraphics[width=1\linewidth]{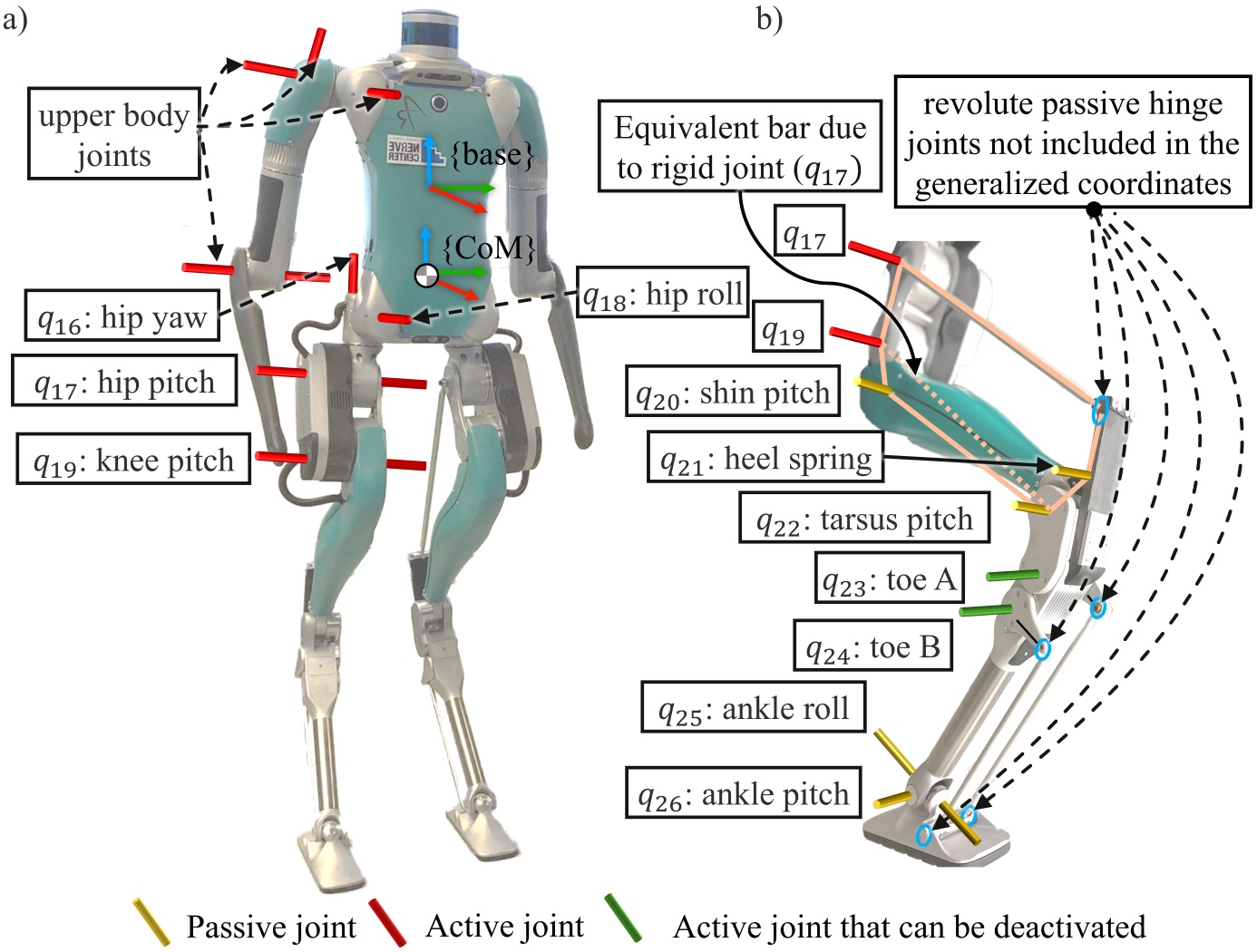}
    \vspace{-0.25 in}
    \caption{Illustration of Digit's leg joints from a) the front view and b) the side view.
    The joints highlighted in green, specifically the toe joints, are active during the swinging phase of the corresponding leg but are deactivated upon ground contact. 
    The side view additionally depicts the three closed-loop linkages, which are governed by holonomic constraints.}
    \label{fig:digit_joints}
    \vspace{-0.2 in}
\end{figure}

\subsubsection{Degree of underactuation}

To introduce the design of the desired full-dimensional reference trajectories, we first determine the number of DoFs that can be directly commanded (i.e., conversely, the degree of underactuation).

At a floating-base configuration where the robot is detached from the ground, the Digit robot has 36 DoFs, which includes the 6 DoFs from the floating-base coordinates.
An illustration of Digit's revolute joints is given in Fig.~\ref{fig:digit_joints}.

One set of six holonomic constraints are formed at the foot-ground contact region when the contact is full and static.
Another set of ten holonomic constraints are induced by the kinematic structure of the Digit robot, including: (i) two passive shin joints, with joint angles $q_{5}$ (left leg) and $q_{20}$ (right leg), which are both treated as rigid joints due to their high stiffness; (ii) two heel springs ($q_{21}$ and $q_{6}$) on both legs, which are considered as rigid links given their high stiffness; and (iii) three closed-loop linkages on each leg, which are used to actuate the passive joints and inherently possess three holonomic constraints per leg.

Under these holonomic constraints, Digit's number of DoFs during walking is:
$
\text{DoF} = 36 - 6 - 10 = 20$.

The robot has 20 independent joint actuators. To enable point contact for aligning with the ALIP-DRS model, the Toe-A and Toe-B joints (see Fig.~\ref{fig:digit_joints}) of the support foot are deactivated. 
Thus, the robot has 18 active actuators.

Given that the number of active actuators, denoted as $n_a$, is 18, we know $n_a =18 < \text{DoF}$. Thus, the Digit walking is underactuated, with two degrees of underactuation.
The resulting four-dimensional unactuated subsystem corresponds to the ALIP-DRS model in the saggital and frontal planes.

\subsubsection{Full-dimensional control variable selection}

With 18 active actuators, 18 control variables can be directly commanded.
Leaving the ALIP-DRS state $\mathbf{x}$ and $\mathbf{y}$ uncontrolled, 
the control variables $\mathbf{h}_c\in\mathbb{R}^{18}$ are chosen as:
\begin{equation}
\label{equ:hc}
\mathbf{h}_c(\mathbf{q})=
\begin{bmatrix}
    z_{SC}(\mathbf{q})\\
    \boldsymbol{\gamma}_{Tr}\\
    x_{Sw}(\mathbf{q})\\
    y_{Sw}(\mathbf{q})\\
    z_{Sw}(\mathbf{q})\\
    \boldsymbol{\gamma}_{Sw}(\mathbf{q})\\
    \mathbf{q}_{upper}
\end{bmatrix}.
\end{equation}
Here, $\mathbf{q} \in Q$ is the generalized coordinates, with $Q \subset \mathbb{R}^{36}$ the configuration space.
$\boldsymbol{\gamma}_{Tr} \in \mathbb{R}^{3}$ is the trunk orientation.
$x_{Sw}$ and $y_{Sw}$ are the swing-foot positions relative to the CoM in the $x$- and $y$-directions, respectively, while
$z_{Sw}$ is the absolute swing-foot height.
$\boldsymbol{\gamma}_{sw}(\mathbf{q}) \in \mathbb{R}^{3}$ is the swing-foot orientation.
$\mathbf{q}_{upper}\in\mathbb{R}^8$ is the upper-body joint angles.

We choose to directly regulate the CoM height $z_{SC}$ to be constant, so as to align the actual CoM height with the ALIP-DRS model.
Also, we control the trunk orientation $\boldsymbol{\gamma}_{Tr}$ to maintain an upright posture, and keep the upper-body joint angles $\mathbf{q}_{upper}$ fixed for simplicity and for avoiding unexpected arm movements. 
Finally, controlling the swing-foot position ($x_{Sw}$, $y_{Sw}$, and $z_{Sw}$) enables the actual robot to accurately execute the desired footstep locations provided by the higher-layer planner.

\subsubsection{Full-body trajectory generation}
Let $\boldsymbol{\phi}$ denote the desired trajectories of $\mathbf{h}_c$, with 
$\boldsymbol{\phi}=:[\phi_1,\phi_2,~...,~\phi_{18}]^\intercal\in\mathbb{R}^{18}$.
We utilize B\'ezier polynomials~\cite{westervelt2007feedback} encoded by a time-based phase variable $s$~\cite{gao2019dscc} to design $\boldsymbol{\phi}$.
The encoding variable $s$ represents how long a walking step has progress within a walking step, and is defined as $s:=\frac{t-T_{k-1}}{T_{step}}$ for $t \in [T_{k-1},T_k)$.
The B\'ezier polynomial $\phi_i$ is defined as
${\phi}_i(s):=\sum_{j=0}^{M}{\alpha}_{i,j} \frac{M!}{j!(M-j)!}s^j(1-s)^{M-j}$,
where $M$ represents the user-specified order of the polynomial and ${\alpha}_{i,j}$ is the coefficient of the $j^{th}$ term.
An example of $\boldsymbol{\phi}$ is given in Sec.~\ref{Section: experiment}-A.

Based on the desired footstep locations generated by the higher-layer planner as well as the robot's current actual states, the middle layer continuously updates the B\'ezier coefficients of $\phi_5(s)$ and $\phi_6(s)$, which are the desired swing-foot trajectories for $x_{Sw}$ and $y_{Sw}$, respectively. The update procedure at time step $t$ is outlined in Algorithm 1, where the value of a variable $(\cdot)$ at $t$ is denoted by $(\cdot)_t$.

 \begin{algorithm}
 \label{alg:ALIP update}
 \SetAlgoLined
  \caption{Pseudocode for updating the B\'ezier polynomial coefficients at time step $t$ based on ALIP-DRS planner}

  \While{True }{
   At time step $t$, obtain the current generalized coordinates $\mathbf{q}_t$ of the full-order robot.

   \If{
     s=0
   }
   {
   Assign the current swing-foot position in the $x$- and $y$-directions to $\alpha_{5,0}$ and $\alpha_{6,0}$:
   
  ~~$\alpha_{5,0}$ 
   $\leftarrow$ $x_{Sw}(\mathbf{q}_t)$,
   $\alpha_{6,0}$ 
   $\leftarrow$ $y_{Sw}(\mathbf{q}_t)$.
    }
    {

    }
    \eIf{
     $0\leq s \leq1$
   }{
     Convert $\mathbf{q}_t$ into the current ALIP-DRS state:

     ~~$\mathbf{x}_t\leftarrow\mathbf{x}(\mathbf{q}_t)$,
     ~~$\mathbf{y}_t\leftarrow\mathbf{y}(\mathbf{q}_t)$.

     Compute the pre-impact angular momentum $L_{y,S}(T_{k}^-)$ and $L_{x,S}(T_{k}^-)$ using~\eqref{equ:linear system solution} and~\eqref{equ:linear system solution lateral}, based on $\mathbf{x}_t$, $\mathbf{y}_t$, and known DRS motion.
    
     Compute ($x_{SwC}(T_k^-)$, $y_{SwC}(T_k^-)$) using~\eqref{equ:swing foot user input} and~\eqref{equ:swing foot user input lateral}.

     Assign the planned swing-foot landing locations ($x_{SwC}(T_k^-)$, $y_{SwC}(T_k^-)$) to the appropriate B\'ezier coefficients of $\phi_{5}$ and $\phi_{6}$:
    
     ~~$\alpha_{5,M} \leftarrow -x_{SwC}(T_k^-)$, $\alpha_{6,M} \leftarrow -y_{SwC}(T_k^-)$.

    Assign $\alpha_{5,1}$,...,$\alpha_{5,M-1}$ as the linear interpolation between $\alpha_{5,0}$ and $\alpha_{5,M}$.
    
    Assign $\alpha_{6,1}$,...,$\alpha_{6,M-1}$ as the linear interpolation between $\alpha_{6,0}$ and $\alpha_{6,M}$.
   }
   {
     No update on $\alpha_{5,0}$,...,$\alpha_{5,M}$ and $\alpha_{6,0}$,...,$\alpha_{6,M}$.
   }
  }
 \end{algorithm}

\vspace{-0.1 in}
\subsection{Lower-Layer Feedback Control}

This subsection describes the feedback controller designed to track the desired trajectories planned by the middle layer.

Various existing control techniques can be utilized to construct the lower-level controller to ensure reliable trajectory tracking.
In this study, we employ the input-output linearizing control approach, which is inspired by the Hybrid Zero Dynamics (HZD) framework~\cite{westervelt2007feedback}, as explained next.

\subsubsection{Input-output linearizing control}
\label{Sec: IO}

The input-output linearizing control law has been previously implemented in simulations for humanoid walking on static surface ~\cite{gao2019dscc} and quadrupedal robot walking during vertical DRS motion~\cite{iqbal2022drs}.

This controller uses the full-order continuous-phase robot model to exactly linearize the dynamics of the output function, and then shapes the output function dynamics based on the well-studied linear system theory.
To ensure accurate trajectory tracking, we choose the output function as the tracking error $\mathbf{h}$ defined as $\mathbf{h} := \mathbf{h}_c - \boldsymbol{\phi}$.

Using Lagrange's method, we can obtain the continuous-phase full-order model as:
\vspace{-0.05 in}
\begin{equation}
\label{equ:full order dynamics}
    \mathbf{M}(\mathbf{q})\ddot{\mathbf{q}}+\mathbf{c}(\mathbf{q},\mathbf{\dot{q}})=\mathbf{B}\boldsymbol{\tau}_u+\mathbf{J}_{c}^\intercal\mathbf{f}_{c},
\end{equation}
where $\mathbf{M}$ is the inertia matrix,
$\mathbf{c}$ is the sum of the gravitational, Coriolis, and centrifugal terms,
$\mathbf{B}$ is the input-selection matrix, and
$\boldsymbol{\tau}_u$ is a joint-torque vector.
$\mathbf{f}_c$ is the force vector that enforces the 16 holonomic constraints mentioned earlier, which includes the ground-reaction force.
$\mathbf{J}_c(\mathbf{q})$ is a Jacobian matrix associated with those holonomic constraints, whose expression is omitted for brevity.

The robot's 16 holonomic constraints are expressed as:
\vspace{-0.05 in}
\begin{equation}
\label{equ:holo constraint}
    \mathbf{J}_c \ddot{\mathbf{q}} + \dot{\mathbf{J}}_c \dot{\mathbf{q}} = \ddot{\mathbf{p}}_c(t).
\end{equation}
Here the time-varying vector $\mathbf{p}_c(t)$ is defined as $\mathbf{p}_c(t):=[{\mathbf{p}}^{\intercal}_S(t),{\boldsymbol{\gamma}}^{\intercal}_S(t),\mathbf{0}]^{\intercal}$,
where $\mathbf{p}_S(t)$ and $\boldsymbol{\gamma}_S(t)$ are the position and orientation of the foot-surface contact region that reside in the DRS, respectively, and the dimension of $\mathbf{0}$ is $1 \times 10$.
The zero vector corresponds to the ten holonomic constraints in Digit's kinematic structure, as explained in subsection B.

Integrating~\eqref{equ:full order dynamics} and~\eqref{equ:holo constraint}, we obtain:
\vspace{-0.05 in}
\begin{equation}
    \label{equ:full order dynamics with holo}
    \mathbf{M}(\mathbf{q})\ddot{\mathbf{q}}+\bar{\mathbf{c}}(t,\mathbf{q},\mathbf{\dot{q}})=\bar{\mathbf{B}}\boldsymbol{\tau}_u,
\end{equation}
where
$\bar{\mathbf{c}}:= \mathbf{c}-\mathbf{{J}}^\intercal_c(\mathbf{J}_c\mathbf{M}^{-1}\mathbf{{J}}^\intercal_c)^{-1}(\mathbf{J}_c\mathbf{M}^{-1}\mathbf{c}-\mathbf{\dot{J}}_c\mathbf{\dot{q}}+\ddot{\mathbf{p}}_c (t))$ and
$\bar{\mathbf{B}} := \mathbf{B} -\mathbf{J}^\intercal_c(\mathbf{J}_c\mathbf{M}^{-1}\mathbf{J}^\intercal_c)^{-1}(\mathbf{J}_c\mathbf{M}^{-1}\mathbf{B})$~\cite{gao2019global}.

Due to the nonlinearity of this full-order dynamics model, the output-function dynamics is also nonlinear.
To linearize the output-function dynamics, the input-output linearizing control law is expressed as:
\vspace{-0.05 in}
\begin{equation}
\small
    \label{equ:feedback control law}
	\boldsymbol{\tau}_u=
 \big(\tfrac{\partial \mathbf{h}_c}{\partial \mathbf{q}} \mathbf{M}^{-1}\bar{\mathbf{B}} \big)^{-1}
 \Big[ \big(\tfrac{\partial \mathbf{h}_c}{\partial \mathbf{q}} \big)
 \mathbf{M}^{-1}\bar{\mathbf{c}}
 +
 \mathbf{v}
 -
 \tfrac{\partial}{\partial \mathbf{q}} \big(\tfrac{\partial \mathbf{h}_c}{\partial \mathbf{q}}\mathbf{\dot{q} \big)\dot{q}}
 +
 \tfrac{1}{T_{step}^2}\tfrac{\partial \boldsymbol{\phi}}{\partial s^2} \Big],
\end{equation}
where the function $\mathbf{v}$ is designed later.
Note that $\mathbf{h}_c$ can be chosen such that there exists an open subset of the robot's configuration space $Q$ on which $\tfrac{\partial \mathbf{h}_c}{\partial \mathbf{q}} \mathbf{M}^{-1}\bar{\mathbf{B}}_j$ is invertible.

By applying the control law in~\eqref{equ:feedback control law}, we obtain the linear closed-loop dynamics of the output function $\mathbf{h}$ during a continuous phase: $\ddot{\mathbf{h}} = \mathbf{v}$.
Without of loss of generality, $\mathbf{v}$ can be chosen as a stabilizing proportional derivative (PD) controller.
That is, $\mathbf{v}= -\mathbf{K}_p\mathbf{h} - \mathbf{K}_d\dot{\mathbf{h}}$, with the PD gains $\mathbf{K}_p$ and $\mathbf{K}_d$ tuned to ensure the closed-loop stability of the output-function dynamics given by $\ddot{\mathbf{h}}+\mathbf{K}_d\dot{\mathbf{h}}+\mathbf{K}_p\mathbf{h}=\mathbf{0}$.

Although the input-output linearizing control approach explicitly addresses the nonlinear full-order model in~\eqref{equ:full order dynamics with holo}, its effectiveness relies on the model's accuracy.
Due to the presence of springs in the Digit robot (Fig. \ref{fig:digit_joints}), there is a notable mismatch between the model and the actual robot dynamics.
Thus, this study also considers an inverse kinematics approach ~\cite{nakanishi2008operational} as a candidate low-level controller.

\subsubsection{Inverse kinematics approach}
\label{Sec: IK}
The closed-form inverse kinematics approach is independent of a robot's dynamics model, and thus can be more robust than the input-output linearizing controller in~\eqref{equ:feedback control law}, even in the presence of Digit's springs.
As the original inverse kinematics approach~\cite{nakanishi2008operational} is designed for fully actuated robots, we adapt it to underactuated walking robots as follows.

Let $\mathbf{q}_a \in \mathbb{R}^{18} $ denote the angles of all actively actuated joints.
Then $\mathbf{q}_a = \mathbf{B}_a \mathbf{q}$ with $\mathbf{B}_a$ a constant selection matrix.

The inverse kinematics controller treats the velocities of the actuated joints, $\dot{\mathbf{q}}_a $, as control inputs, which can be directly set to any arbitrary values within the joint limits to command $\mathbf{q}_a$.
The objective of the controller is to make the closed-loop output-function system stable and first-order, with a linear form of
$
	\dot{\mathbf{h}}(\mathbf{q},\dot{\mathbf{q}}) + \boldsymbol{\kmat} \hvec = \mathbf{0}
$, where $\boldsymbol{\kmat} \inRm{18}{18}$ is a positive-definite control gain matrix.


By the definition of the output function $\mathbf{h}$, we know $\dot{\mathbf{h}}=\dot{\mathbf{h}}_c - \dot{\boldsymbol{\phi}}$, and this first-order system can be re-written as:
\vspace{-0.05 in}
\begin{align}
    \mathbf{J}_{h_c} (\mathbf{q}) \dot{\mathbf{q}} = \dot{\boldsymbol{\phi}} 
    - \boldsymbol{\kmat} \mathbf{h}(\mathbf{q}), 
    \label{eq: linear system ikin}
\end{align}
where $\mathbf{J}_{h_c}(\mathbf{q}) := \frac{\partial \mathbf{h}_c (\mathbf{q})}{\partial \qvec}\inRm{18}{36}$. 
Since $\mathbf{J}_{h_c}$ is not invertible, we cannot directly use this equation to solve for the reference velocities $\dot{\mathbf{q}}$ (which contains $\dot{\mathbf{q}}_a$). 

The non-invertibility of $\mathbf{J}_{h_c}$ is essentially due to the underactuation induced by Digit's passive support-toe joints as well as the presence of the 16 holonomic constraints in Digit's kinematic structure.
Thus, to solve for $\dot{\mathbf{q}}$, we map $\dot{\mathbf{q}}$ into the passive support-toe velocities $\dot{\qvec}_{Toe} \in \mathbb{R}^2$ while explicitly expressing the holonomic constraints using $\dot{\mathbf{q}}$.

Mapping $\dot{\mathbf{q}}$ into $\dot{\qvec}_{Toe}$, we obtain $\mathbf{S}_{Toe} \dot{\qvec} = \dot{\qvec}_{Toe}$,
where $\mathbf{S}_{Toe} \in \mathbb{R}^{2 \times 36}$ is a known constant selection matrix. 
Meanwhile, to respect the 16 holonomic constraints, $\dot{\mathbf{q}}$ needs to satisfy $\mathbf{J}_c (\mathbf{q}) \dot{\mathbf{q}} = \dot{\mathbf{p}}_c(t)$, which corresponds to \eqref{equ:holo constraint}.
Combining these two equations with~\eqref{eq: linear system ikin} gives:
\vspace{-0.05 in}
\begin{equation}
\label{Jk}
\mathbf{J}_{IK} (\mathbf{q})   \dot{\qvec}
    =
    \mathbf{v}_{IK}(t,\mathbf{q},\dot{\mathbf{q}})
    :=
    \begin{bmatrix}
        \dot{\boldsymbol{\phi} } - \boldsymbol{\kmat} \mathbf{h}(\mathbf{q}) \\
        \mathbf{\dot{p}}_c(t) \\
        \dot{\qvec}_{Toe}
    \end{bmatrix},
\end{equation}
where 
$\mathbf{J}_{IK}(\mathbf{q}) := [\mathbf{J}_{h_c}^\intercal(\mathbf{q}), \mathbf{J}^\intercal_c(\mathbf{q}), \mathbf{S}_{Toe}^\intercal ]^\intercal \in \mathbb{R}^{36 \times 36}$.
Note that we can select the control variables $\mathbf{h}_c$ such that $\mathbf{J}_{IK} (\mathbf{q})$ is invertible on $Q$.
Thus, the velocity command $\dot{\qvec}_a$ is given by:
\vspace{-0.05 in}
\begin{equation}
    \dot{\qvec}_a = \mathbf{B}_a \mathbf{J}_{IK}^{-1}  (\mathbf{q}) 
    \mathbf{v}_{IK}(t,\mathbf{q},\dot{\mathbf{q}}).
    \label{eq: qa}
\end{equation}

Since the actual robot is directly actuated by joint torques instead of joint velocities, we construct a PD joint-torque controller as inspired by~\cite{gao2019global}, which treats the joint velocity command in~\eqref{eq: qa} as the desired velocities denoted as $\mathbf{q}_{a,d}$.
Then, from~\eqref{eq: qa}, we have $\mathbf{q}_{a,d} = \mathbf{B}_a \mathbf{J}_{IK}^{-1}  (\mathbf{q}) 
    \mathbf{v}_{IK}(t,\mathbf{q},\dot{\mathbf{q}})$.

Thus, the PD joint-torque controller is designed as:
\vspace{-0.05 in}
\begin{equation}
    \utauvec = -\bar{\mathbf{K}}_p (\qvec_a - \mathbf{q}_{a,d}) - \bar{\mathbf{K}}_d (\dot{\qvec}_a - \dot{\mathbf{q}}_{a,d}),
    \label{eq:ik-pd-gains}
\end{equation}
where $\bar{\mathbf{K}}_p, \bar{\mathbf{K}}_d \in \mathbb{R}^{18 \times 18}$ are PD matrices to be appropriately tuned. 
This controller is experimentally implemented on the physical Digit robot, as described in Sec.~\ref{Section: experiment}.

\section{Stability Analysis for Full-Order Model}
\label{Section: Stability}
This section presents the stability analysis for the hybrid closed-loop full-order system under the proposed framework.



The stability of the closed-loop full-order system needs to be analyzed, as it is not automatically guaranteed by the proposed control framework. This necessity arises because the framework does not explicitly treat the inevitable inaccuracy in the ALIP-DRS model. While the higher-layer planner of the framework stabilizes the ALIP-DRS model, which approximates the unactuated dynamics, it does not directly stabilize the complete nonlinear model of the unactuated subsystem. Meanwhile, the lower-layer controller regulates the fully actuated output-function dynamics but does not directly influence the unactuated subsystem.

Given the nonlinear and hybrid nature of the full-order robot model, the proposed stability analysis employs Lyapunov functions~\cite{khalil1996noninear}. Without loss of generality, this analysis assumes that the framework's lower layer is the input-output linearizing controller detailed in~\eqref{equ:feedback control law}. The analysis considers the general case where $N_1T_{step}=N_2T_{x,DRS}=T_{sys}$.

\vspace{-0.1 in}
\subsection{Closed-loop Error Dynamics}

\subsubsection{State definition}

Let $\mathbf{X}$ denote the state of the full-order control system, comprising state variables of the unactuated dynamics, which is approximated by the ALIP-DRS model, and the directly controlled output-function dynamics. 

We compactly denote the output function state as $\mathbf{x}_{h} \in \mathbb{R}^{2 n_a}$, and define it as
$\mathbf{x}_{h} := \begin{bmatrix} \mathbf{h}^\intercal(t,\mathbf{q}),~  \dot{\mathbf{h}}^\intercal (t,\mathbf{q},\dot{\mathbf{q}})  \end{bmatrix}^\intercal$.
Recall $n_a=18$ for the underactuated Digit, as explained earlier.

Correspondingly, we introduce a new state $\mathbf{x}_{\eta} \in \mathbb{R}^{4}$ to denote the tracking error state associated with the unactuated ALIP-DRS state:
$
    \mathbf{x}_{\eta} := 
    \begin{bmatrix} 
    \mathbf{x}^\intercal(\mathbf{q}),~\mathbf{y}^\intercal(\mathbf{q})
    \end{bmatrix}^\intercal
    - \boldsymbol{\psi}(t)
$, where $\boldsymbol{\psi}(t)$ is the periodic solution to the closed-loop ALIP-DRS model.

Then, the complete state is defined as:
$\mathbf{X} := 
    \begin{bmatrix}
        \mathbf{x}_ {h}^\intercal,~
        \mathbf{x}_ {\eta}^\intercal
    \end{bmatrix}^\intercal.$

\subsubsection{Hybrid, nonlinear full-order closed-loop dynamics}

Based on the proposed low-level control law, the hybrid closed-loop ALIP-DRS model, and the output-function dynamics model, the closed-loop full-order dynamics is:
\vspace{-0.05 in}
\begin{equation}
\label{eq: xu dynamics}
    \begin{cases}
    \begin{cases}
            &\dot{\mathbf{x}}_{h}
            = \mathbf{A}_{h} \mathbf{x}_{h}
            \\
            &\dot{\mathbf{x}}_{\eta}
            = 
            \mathbf{A}_{\eta} \mathbf{x}_{\eta}
            +
            \mathbf{d}_{A}
    \end{cases}
        &\text{if}~ 
                t \neq T_k^-  ;
                \\
    \begin{cases}
        &
        \mathbf{x}_h^+ =  \boldsymbol{\Delta}_h(t,\mathbf{x}_{h}^-,\mathbf{x}_{\eta}^-)
        \\
        &
        \mathbf{x}_\eta^+ = 
        \boldsymbol{\Delta}_{\eta} \mathbf{x}_{\eta}^-
        +
        \mathbf{d}_{\Delta}
    \end{cases}
    
         &\text{if}~ 
                t = T_k^- ,
    \end{cases}
\end{equation}
where $k \in \mathbb{N}$,
$\mathbf{A}_{h}:= {\small\begin{bmatrix} 
	\mathbf{0} & \mathbf{I} \\
	-\mathbf{K}_{p} & -\mathbf{K}_{d} 
	\end{bmatrix}}$,
and $\mathbf{A}_{\eta} := {\small \begin{bmatrix}
     \mathbf{A}_x & \mathbf{0}\\
     \mathbf{0} & \mathbf{A}_y
 \end{bmatrix}}$.
The vector-valued nonlinear function $\boldsymbol{\Delta}_h$ is the reset map of $\mathbf{x}_h$,
which can be readily obtained based on the definition of $\mathbf{x}_h$ and the reset map of the robot's generalized coordinates and velocities given in~\cite{westervelt2007feedback}.
The constant matrix $\boldsymbol{\Delta}_\eta$ is defined as $\boldsymbol{\Delta}_\eta :={\small \begin{bmatrix}
    \mathbf{B}_x & \mathbf{0}\\
     \mathbf{0} & \mathbf{B}_y
\end{bmatrix}}$ with $\mathbf{B}_y$ the frontal-plane counterpart of $\mathbf{B}_x$.
$\mathbf{d}_{A}$ and $\mathbf{d}_{\Delta}$ represent the discrepancies between the reduced-order ALIP-DRS and the full-order model.

Here, $\| \mathbf{d}_{A} \|$ and $\|\mathbf{d}_{\Delta} \|$ are assumed to be bounded above by positive, real numbers $L_A$ and $L_\Delta$: $\| \mathbf{d}_{A} \| \leq L_A$ and $\| \mathbf{d}_{\Delta} \| \leq L_\Delta$ for any bounded
initial state satisfying $\mathbf{X}({0}) \in B_{r_d}(\mathbf{0}) := 
\{\mathbf{X}:  \| \mathbf{X} \| \leq r_d  \}
$, where $r_d$ is a real positive constant.

Without loss of generality, the stability analysis considers the $k^{th}$ gait cycle on $t\in (T_{k},T_{k+1})$ with $k \in \mathbb{N}$, which comprises a continuous phase on $t \in (T_{k},T_{k+1})$ and a switching event at $t = T_{k+1}^-$.
For brevity, the values of a variable $\star$ at $t = T_{k}^+$ and $t = T_{k}^-$ are denoted as $\star|^+_{k}$ and $\star|^-_{k}$, respectively.

\vspace{-0.1 in}
\subsection{Lyapunov Function Candidate}

As the full-order robot model comprises unactuated and fully actuated subsystems, we design the Lyapunov function candidate $V(\mathbf{X})$ as $ V (\mathbf{X}) = \beta_h V_{h} (\mathbf{x}_{h}) 
    + \beta_\eta V_\eta(\mathbf{x}_{\eta} )$,
where $V_h (\mathbf{x}_{h})$ and $V_\eta(\mathbf{x}_{\eta})$ are positive-definite functions
and $\beta_h$ and $\beta_\eta$ are positive constants to be designed.

\subsubsection{Closed-form construction of $V_h(\mathbf{x}_h)$}
We construct the component $V_h(\mathbf{x}_h)$ as the Lyapunov function for the closed-loop, continuous-phase output-function dynamics (i.e., $\ddot{\mathbf{h}}+\mathbf{K}_d \dot{\mathbf{h}} + \mathbf{K}_p \mathbf{h}=\mathbf{0}$).
Since the lower-layer controller can stabilize this output-function dynamics with appropriately chosen PD gains $\mathbf{K}_p$ and $\mathbf{K}_d$, the Converse Lyapunov Theorem~\cite{khalil1996noninear} implies that the Lyapunov function $V_h(\mathbf{x}_h)$ exists for this dynamics.
Further, since this dynamics is linear and time-invariant, 
the closed-form expression of $V_h(\mathbf{x}_h)$ can be constructed via a Lyapunov equation.

\subsubsection{Closed-form construction of $V_{\eta}(\mathbf{x}_{\eta})$}
We define $V_{\eta}(\mathbf{x}_{\eta})$ as the Lyapunov function for the closed-loop, step-to-step ALIP-DRS dynamics. This dynamics is analytically attractable and is exponentially stabilized by the proposed footstep controller, as established in Theorem~\ref{thm1}.
The Lyapunov equation for general linear discrete-time systems, which include this dynamics, is employed to derive the closed-form expression of $V_{\eta}(\mathbf{x}_{\eta})$.
Note that the sagittal-plane component of this dynamics is given in~\eqref{equ: respect impact not so much}, and its frontal-plane counterpart can be derived similarly.

\subsubsection{Convergence and Boundedness of $V_h$ and $V_{\eta}$}
To establish the stability condition for the closed-loop system in~\eqref{eq: xu dynamics}, we first analyze the convergence and boundedness of $V_h$ and $V_{\eta}$, as summarized in the following propositions.

\begin{prop}[\textbf{Continuous-phase convergence of} $V_h$]
\label{prop2}
Consider the proposed lower-layer control law in~\eqref{equ:feedback control law} and the closed-loop error dynamics in~\eqref{eq: xu dynamics}.
Also, consider the condition that the PD gains of the lower-layer control law are selected such that $\mathbf{A}_{h}$ is Hurwitz.
Then, there exist positive constants $r_{h}$, $c_{h1}$, $c_{h2}$, and $c_{h3}$ such that the function $V_{h}$ satisfies
$c_{h1}  \|   \mathbf{x}_{h}    \|^2   
    \leq 
    V_{h} (  \mathbf{x}_{h}  ) 
    \leq 	
    c_{h2}  \|   \mathbf{x}_{h}    \|^2$
and
$\dot{V}_{h}  \leq 	- c_{h3} {V}_{h}$
on $t \in (T_{k},T_{k+1})$.
Accordingly, $V_{h}$ exponentially converges as:
$V_{h}|^-_{k+1} \leq e^{-c_{3} ( T_{k+1} - T_{k} )} V_{h}|^+_{k}$
during continuous phases.
\hfill $\diamondsuit$
\end{prop}

The proof of Proposition \ref{prop2} is a direct adaptation of the Lyapunov stability theory from~\cite{khalil1996noninear} and is thus omitted.

\begin{prop}[\textbf{Step-to-step convergence of} $V_\eta$]
\label{prop3}
Consider the proposed control framework and the closed-loop system in~\eqref{eq: xu dynamics}.
There exist positive real constants $r_{\eta}$, $c_{\eta 1}$, $c_{\eta 2}$, and $c_{\eta 4}$ and non-negative real number $c_{\eta 3}<1$ such that $V_{\eta}$ satisfies
$c_{\eta 1} \| \mathbf{x}_\eta \|^2 \leq V_\eta \leq c_{\eta 2} \| \mathbf{x}_\eta \|^2$
and
${V}_{\eta} |_{k+1}^+  \leq  c_{\eta 3} {V}_{\eta}|_{k}^+ +  c_{\eta 4}$
for any bounded initial state satifying
$\mathbf{X}(0) \in B_{r_\eta}(\mathbf{0})$.
\hfill $\diamondsuit$
\end{prop}

Proposition \ref{prop3} holds because by Theorem~\ref{thm2}, the proposed discrete footstep control law is exponentially stabilizing for the ALIP-DRS model and the inaccuracy of the ALIP-DRS model (i.e., $\mathbf{d}_A$ and $\mathbf{d}_\Delta$) are bounded as mentioned earlier.

\vspace{-0.1 in}
\subsection{Main Stability Theorem}

Based on Propositions~\ref{prop1}-\ref{prop3}, the stability conditions for the hybrid error dynamics in~\eqref{eq: xu dynamics} are introduced next.

\begin{thm}[\textbf{Closed-loop stability conditions for the full-order model}]
Let all conditions in Propositions~\ref{prop1}-\ref{prop3} hold.
If the continuous-phase convergence rates $\frac{1}{c_{\eta 3}}$ and $c_{h3}$ are sufficiently high, then the origin of the hybrid closed-loop error system in~\eqref{eq: xu dynamics} is locally stable in the sense of Lyapunov under the proposed control framework.
\hfill $\diamondsuit$
\label{thm2}
\end{thm}

{\it Rationale of proof}:
By the theory of multiple Lyapunov functions~\cite{branicky1998multiple}, the origin of the hybrid closed-loop error system in~\eqref{eq: xu dynamics} is locally stable if:
(S1) the ``switch-out'' value of the Lyapunov function candidate $V(\mathbf{X})$ is bounded above by a positive-definite function of the ``switch-in'' value of $V(\mathbf{X})$ within each continuous phase
and
(S2) the values of $V(\mathbf{X})$ immediately after each switching event satisfy
$V|_{k+1}^+  \leq  V|_{k}^+$.
To prove Theorem~\ref{thm2}, we show that $V(\mathbf{X})$ satisfies these two conditions if all conditions in Theorem~\ref{thm2} are met.

{\it Proof}:
We begin the proof by first analyzing the evolution of the Lyapunov function candidate $V(\mathbf{X})$ for the $k^{th}$ complete gait cycle on $t\in (T_{k},T_{k+1}]$, which comprises one continuous phase and one switching event.

The continuous-phase boundedness of $V_h$ is given in Proposition~\ref{prop2}.
Meanwhile, since the continuous-phase uncertainty $\mathbf{d}_{A}$ in the unactuated dynamics is locally bounded and the matrix $\mathbf{A}_{\eta}$ is constant and finite,
we know that there exist positive real constants $L_1$, $L_2$, and $r_1$ such that
\begin{equation}
V_\eta |^{-}_{k+1} \leq L_1 V_\eta |^{+}_{k} + L_2
\label{V_eta-continuous}
\end{equation}
holds for any initial state satisfying $\mathbf{X}(0) \in B_{r_1}$.

From Proposition~\ref{prop2} and \eqref{V_eta-continuous}, we know the following inequality holds for any $\mathbf{X}(0) \in B_{r_1}(\mathbf{0})$:
\begin{equation}
\begin{aligned}
V |^{-}_{k+1} &=\beta_h V_h|^{-}_{k+1} + \beta_{\eta} V_\eta|^{-}_{k+1} \\
&\leq
\beta_h e^{-c_{3} ( T_{k+1} - T_{k} )} V_{h}|^+_{k} 
+ \beta_{\eta} L_1 V_\eta |^{+}_{k} + \beta_{\eta} L_2 \\
& \leq L_v V |^{+}_{k} + \beta_{\eta} L_2,
\end{aligned}
\end{equation}
where $L_v:=\max( e^{-c_{3} ( T_{k+1} - T_{k} )} ,  L_1  )$.
This inequality verifies that the stability condition (S1) is met.

Since the nonlinear reset map $\boldsymbol{\Delta}_h$ is continuously differentiable in $t$ and $\mathbf{X}$~\cite{9108552},
we know $\boldsymbol{\Delta}_h$ is locally Liptchiz in $t$ and $\mathbf{X}$~\cite{khalil1996noninear}. 
Then, by applying the triangle inequality to the reset map $\boldsymbol{\Delta}_h$~\cite{gao2023provably}, we know 
there exists positive real scalars $L_\Delta$, $\epsilon_\Delta$, and $r_\Delta$ such that
\begin{equation}
\label{equ: xplus to xplus}
    \Big \|   \mathbf{x}_h |^+_{k+1} \Big \| ^2
    \leq 
    L_h \Big \| \mathbf{x}_h |^-_{k+1}\Big \|^2 + L_\eta \Big \| \mathbf{x}_\eta |^-_{k+1}\Big \|^2
    + \epsilon_\Delta
\end{equation}
holds for a bounded initial state $\mathbf{X}(0) \in B_{r_\Delta}(\mathbf{0})$.

From Propositions~\ref{prop2} and \ref{prop3} and~\eqref{V_eta-continuous}-\eqref{equ: xplus to xplus}, we have:
\begin{equation}
    V_h |^+_{k+1} 
    \leq 
    \tfrac{L_h c_{h2}}{c_{h1}} e^{-c_{3} ( T_{k+1} - T_{k} )} V_{h}|^+_{k} 
    +
    \tfrac{L_\eta L_1 c_{h2}}{c_{\eta1}} V_{\eta}|^+_{k}
    +
    \epsilon_h,
    \label{Vh_s2s}
\end{equation}
where $\epsilon_h := \epsilon_\Delta  c_{h2} + \frac{L_\eta L_2 c_{h2}}{c_{\eta1}}$.

Then, from the definition of $V$, Proposition~\ref{prop3}, and~\eqref{Vh_s2s}, the step-to-step convergence of $V$ is given by:
\begin{equation}
V|^+_{k+1} \leq c_{v1} V|^+_{k} + c_{v2},
\end{equation}
where $c_{v1} := \max (   \tfrac{L_h c_{h2}}{c_{h1}} e^{-c_{h3} ( T_{k+1} - T_{k} )} ,  c_{\eta 3} +  \tfrac{L_1 L_\eta \beta_h c_{h2}}{\beta_\eta c_{\eta 1}} )$
and
$c_{v2} := \beta_h \epsilon_h  + \beta_\eta c_{\eta 4}$.

Since $\beta_h$ and $\beta_\eta$ can be chosen arbitrarily, we know $c_{v2}$ can be set arbitrarily small, along with $\tfrac{L_1 L_\eta \beta_h c_{h2}}{\beta_\eta c_{\eta 1}}$.
Meanwhile, we know $c_{\eta 3} < 1$.
Thus, we can choose $\beta_h$ and $\beta_\eta$ such that $c_{\eta 3} + \tfrac{L_1 L_\eta \beta_h c_{h2}}{\beta_\eta c_{\eta 1}}$ is less than one.
Meanwhile,
by carefully choosing the PD gains in~\eqref{equ:feedback control law}, the continuous-phase convergence rate $c_{h3}$ can be sufficiently large such that $ \tfrac{L_h c_{h2}}{c_{h1}} e^{-c_{h3} ( T_{k+1} - T_{k} )}$ can be any positive number less than one. 
Thus, $c_{v1} <1$ is ensured. 

Since both $c_{v1}$ and $c_{v2}$ can be any arbitrarily small positive number with $c_{v1}<1$, $V|_{k+1}^+ \leq V|_{k}^+$ is guaranteed for any $k \in \mathbb{N}$ and $\mathbf{X}(0) \in B_{r_v}(\mathbf{0})$ with $r_v=\min(r_d, r_{\eta}, r_\Delta,r_1)$.
Thus, condition (S2) is met, which completes the proof.
$\hfill
\square$

\section{Simulation  Validation}
\label{Section: simulation}
This section presents simulations performed on a Digit robot in MATLAB, validating the proposed approach under various surface motions and uncertainties.

Simulation validation is necessary as it enables us to evaluate the proposed framework beyond the physical limitations of hardware experiments, such as the frequency and magnitude constraints of the DRS movement.
The simulation video is available at: \href{https://youtu.be/NtAT0DFtMCY}{https://youtu.be/NtAT0DFtMCY}.

\begin{figure*}[h]
    \centering\includegraphics[width=0.9\linewidth]{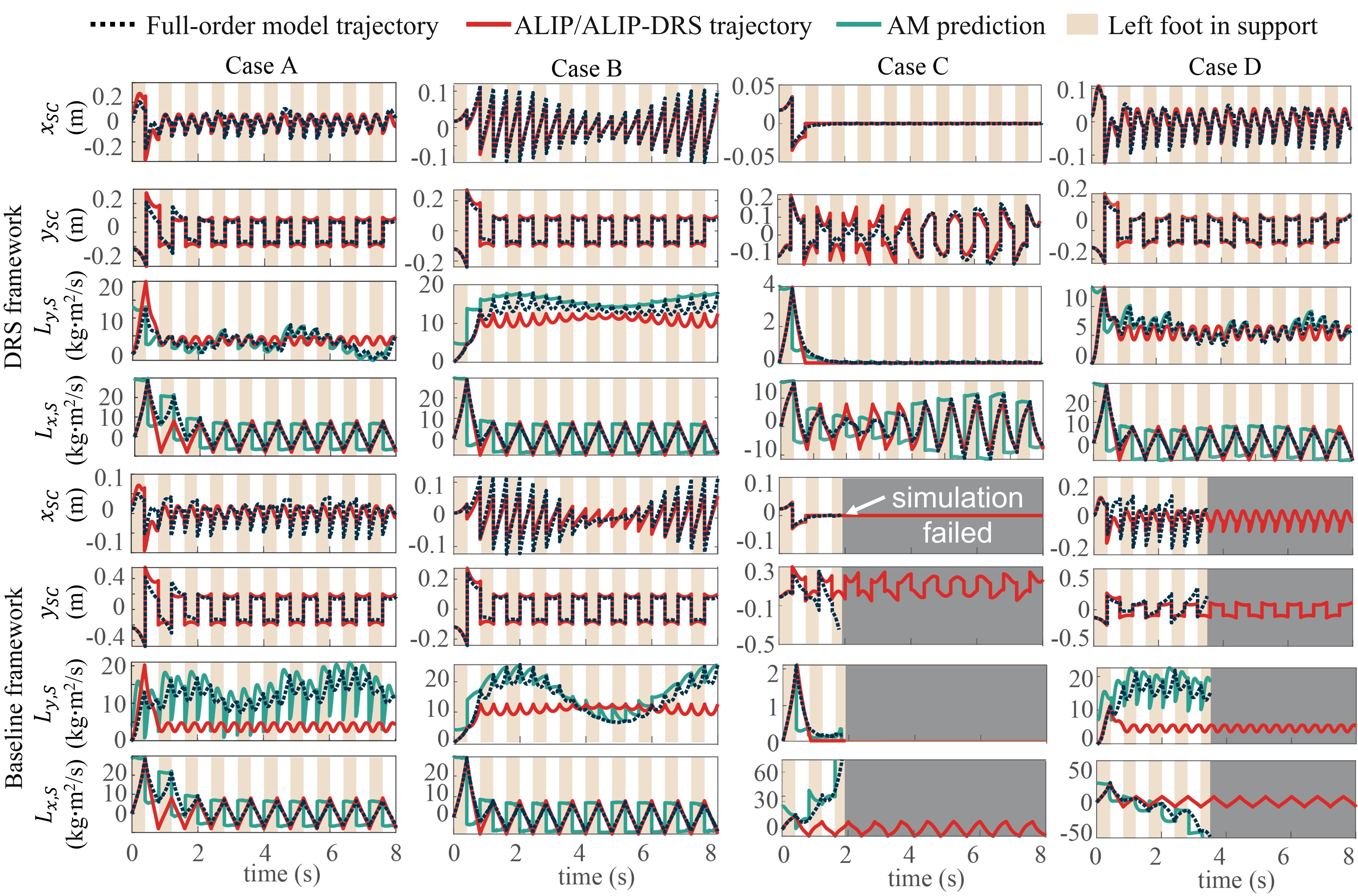}
    \vspace{-0.15 in}
    \caption{Performance comparison between the proposed DRS framework and the baseline framework under simulation Cases A-D. The beige background highlights the left-foot-in-support phase, while the white background denotes the right-foot-in-support phase. Under the proposed framework, the robot's actual trajectories (labeled ``Full-order model trajectory'') align closely with the ALIP-DRS trajectories, and the prediction of angular momentum (AM) at the end of the current step (green curve) is relatively accurate. 
    In contrast, the robot operating under the baseline framework exhibits instability in Cases C and D. Additionally, significant discrepancies are observed between the actual trajectory of the contact angular momentum $L_{y,S}$ and the ALIP trajectory, and the prediction of $L_{y,S}$ at the end of the current step lacks the accuracy seen with the proposed framework.}
    \label{fig: matlab_traj}
    \vspace{-0.2 in}
\end{figure*}

\vspace{-0.1 in}
\subsection{Simulation Setup}

\subsubsection{Setup of higher-layer planner}

Based on Digit's inertial and geometric properties, the parameters of the ALIP-DRS are set as: $H=0.9$ m, $m=46.1$ kg, $T_{step}=0.4$ s, and $W=0.2$ m.
As the ALIP-DRS model in~\eqref{equ:ALIPDRS} indicates,
setting the desired angular momentum $\bar{L}_{y,S}$ is equivalent to setting the desired forward CoM velocity.
For comprehensive assessment across various desired forward CoM velocities, different values of $\bar{L}_{y,S}$ are used, as specified in Table~\ref{Table: DRS list}. 


\subsubsection{Setup of middle and lower layers}

In MATLAB simulations, the B\'ezier polynomials defining the desired full-dimensional trajectories $\boldsymbol{\phi}$ are chosen as follows:
\begin{itemize}
    \item $\phi_1$: Set to $H$ so as to enforce the CoM height of the ALIP-DRS model on the full-order Digit model.
    \item $\phi_2$, $\phi_3$, and $\phi_4$: Set to 0 for a constant trunk orientation.
    \item $\phi_5$ and $\phi_6$: Set according to Algorithm 1.
    \item $\phi_7$: Set with B\'ezier coefficients $[0, 0.02, 0.07, 0.15,$ $ 0.07, 0.02, 0]^\intercal$ to minimize the desired swing-foot height and reduce swing-foot motion.
    \item $\phi_8$, $\phi_9$, and $\phi_{10}$: Set to $0$ to ensure a proper swing-foot orientation at touchdowns.
    \item $\phi_{11}$ to $\phi_{18}$: Set to 0 to keep the arms static.
\end{itemize}

In MATLAB, the model-based input-output linearizing controller is implemented using~\eqref{equ:feedback control law}.
The full-order dynamic matrices $\mathbf{M}$ and $\mathbf{c}$ are obtained using FROST\cite{Hereid2017FROST}.
For all cases, the control gains are set to $\mathbf{K}_p=2500 \cdot \mathbf{I}$ and $\mathbf{K}_d=100 \cdot \mathbf{I}$.
To meet the stability condition in Theorem~\ref{thm2}, these gains are tuned to ensure a sufficiently high continuous-phase convergence rate for the output function state $\mathbf{x}_h$.

\begin{table}[t]
\centering
\caption{{Simulation Cases}} 
\vspace{-0.1 in}
\small
\begin{tabular}{ p{1.0cm}|p{3.6cm}|p{2.2cm}}
\hline
\hline
\centering
Cases  & \centering DRS motion (m)& $\bar{L}_{y,S}$ ($\text{kg}\cdot\text{m}^2/\text{s}$)\\
\hline
\hline
\centering A &  
 \centering $x_S(t)=0.04 \cos(\frac{2\pi}{0.4}t)$ &~~~~~~4.1\\
\hline
\centering B &  \centering $x_S(t)=0.14 \cos(\frac{2\pi}{6}t)$ &~~~~~12.5\\
\hline
\centering C &  \centering $y_S(t)=0.06 \cos(\frac{2\pi}{0.72}t)$ &~~~~~~~0\\
\hline
\centering \multirow{2}{*}{D} &  \centering $x_S(t)=0.04 \cos(\frac{2\pi}{0.4}t)$ &\multirow{2}{*}{~~~~~6.27}
\\
& \centering$y_S(t)=0.1 \cos(\frac{2\pi}{6}t)$ &~
\\
\hline
\end{tabular}
\label{Table: DRS list}
\end{table}

\subsubsection{Simulation cases}
To assess the control framework under both a unity ratio and a real-number ratio between the DRS motion period and one walking-step duration, as well as under different DRS motion directions,
four sets of DRS motions are simulated, as summarized in Table~\ref{Table: DRS list}.
The detailed descriptions of the surface motions are:
\begin{itemize}
    \item[(Case A):] 
    DRS sways in the sagittal plane with the same period as walking (i.e., $T_{step} = T_{x,DRS}=0.4$ s).
    \item[(Case B):] DRS sways in the sagittal plane with its least period $15$ times that of walking (i.e., $15 T_{step} = T_{x,DRS}$).
    \item[(Case C):] DRS sways in the frontal plane with its least period related to the walking period as $9T_{step} = 5T_{y,DRS}$. 
    \item[(Case D):] DRS sways with different amplitudes and periods in the forward and lateral directions. The DRS-walking period ratios are $T_{x,DRS}/T_{step} = 1$ and $T_{y,DRS}/T_{step} = 15$.
\end{itemize}

In all cases, the robot faces towards the $x$-axis of the world frame.
The initial movement status of Digit is quite standing.

\subsection{Comparative Simulations}
To highlight the effectiveness of the proposed control framework (denoted as the ``DRS framework''), we compare it with a previous ALIP-based framework~\cite{gong2022zero} designed for static terrain (denoted as the ``baseline framework''). 
Figure~\ref{fig: matlab_traj} displays the simulation results under Cases A-D.

\subsubsection{Baseline framework setup} The two frameworks have the same setup of middle and lower layers, including control variable selection, full-body trajectory parameterization, and the use of an input-output linearizing controller as the lower layer.
The controller's PD gains are tuned to guarantee accurate full-body trajectory tracking.
The primary difference in the setup lies in the higher-layer planner: the DRS framework explicitly accounts for the DRS motion whereas the baseline framework assumes a static terrain.

\subsubsection{Comparison on walking stability}

As Fig.~\ref{fig: matlab_traj} indicates, the proposed DRS framework drives the full-dimensional trajectories (marked by black, dashed lines) significantly more closely to the desired trajectories (marked by red, solid lines), as compared with the baseline framework.
The DRS framework also ensures stable walking across all four cases.
Yet, the baseline framework fails to sustain stable walking in Cases C and D, causing the robot to fall over laterally. 

\subsubsection{Comparison on angular momentum prediction}
The green curves in Fig.~\ref{fig: matlab_traj} show the predicted angular momentum at the end of the current walking step.
As explained in Sec.~\ref{sec: ALIP control and stability}, at each time step within the current walking step, the higher-layer planner uses~\eqref{equ:linear system solution} and~\eqref{equ:linear system solution lateral} and the robot's current state to compute the predicted angular momentum at the end of the current walking step.
The baseline framework uses the same equations but with zero ground velocities.

Under the DRS framework, the predicted value of the contact angular momentum $L_{y,S}$ at the end of the current walking step (highlighted in green) is relatively consistent within each walking step and across different steps.
In contrast, the baseline framework's prediction varies significantly. 
This improved angular momentum prediction of the proposed framework leads to less variation in the planned foot placement location during a walking step. Thus, the desired swing-foot trajectories tracked by the lower-layer controller are smoother, causing the better tracking performance of the proposed framework compared to the baseline framework.

\begin{table}[t]
\centering
\caption{{Forward velocity regulation performance comparison}} 
\vspace{-0.1 in}
\small
\begin{tabular}{ p{0.8cm}|p{2.0cm}|p{2.0cm}|p{2.0cm}}
\hline
\hline
\centering
\multirow{3}{*}{Cases}{}  & \centering Desired forward velocity (m/s) & \centering DRS framework error (m/s)  &  
\textcolor{white}{ssss}Baseline
\textcolor{white}{sss}framework \textcolor{white}{ss} \textcolor{white}{}error (m/s)\\
\hline
\hline
\centering A &  
 \centering 0.1  &  \centering 0.005 & ~~~~~~0.32\\
\hline
\centering B &  \centering 0.3 &\centering 0.13 & ~~~~~~0.32\\
\hline
\centering C &  \centering 0 &\centering 0.002 & \cellcolor{blue!10} ~~~not stable\\
\hline
\centering D &  \centering 0.15 &\centering 0.01 & \cellcolor{blue!10} ~~~not stable
\\
\hline
\end{tabular}
\vspace{-0.2 in}
\label{Table: velocity regulation}
\end{table}

\subsubsection{Comparison on velocity regulation accuracy}

Table~\ref{Table: velocity regulation} shows the tracking errors for the forward CoM velocity under the proposed and baseline frameworks across Cases A-D. It also provides the desired forward CoM velocity for each case, corresponding to the desired contact angular momentum $L_{y,S}$ outlined in Table~\ref{Table: DRS list}. As indicated in the table, the proposed framework achieves velocity tracking errors ranging from 0.005 m/s to 0.13 m/s across the four cases. In contrast, the baseline framework shows larger errors, reaching 0.32 m/s in Cases A and B. For Cases C and D, the baseline framework’s errors are not reported due to its inability to maintain stable walking in these scenarios.

\begin{figure}[t]
    \centering
    \includegraphics[width=1\linewidth]{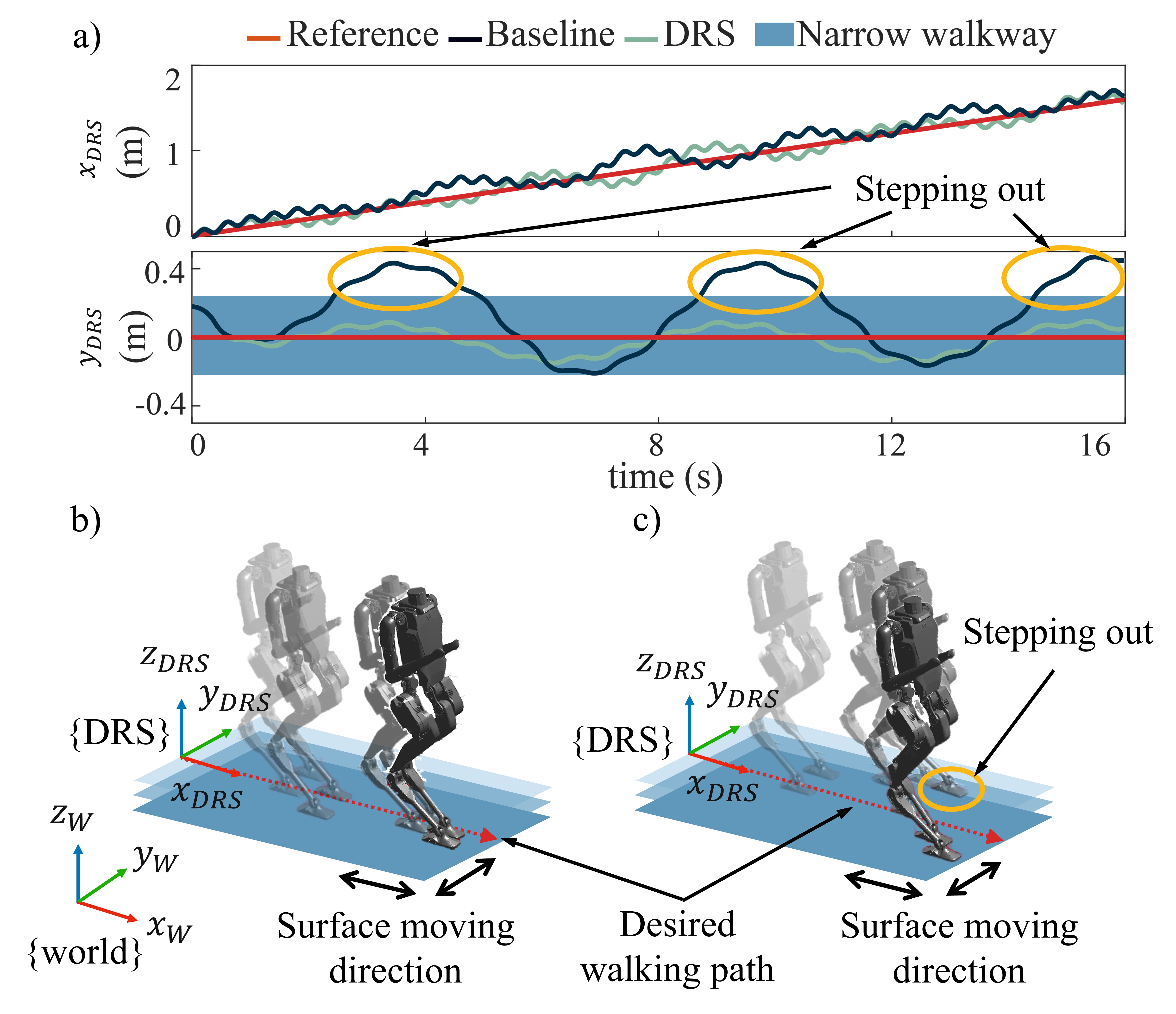}
    \vspace{-0.35 in}
    \caption{Comparison of base-position tracking accuracy between the baseline and DRS frameworks under Case D. a) The DRS framework demonstrates higher tracking accuracy in the $x_{DRS}$-direction compared with the baseline framework, allowing the robot to remain within the confines of the narrow walkway. b) Snapshots depict stable walking by Digit using the proposed DRS framework. c) Under the baseline framework, Digit steps outside the designated walkway, illustrating reduced tracking accuracy.}
    \label{fig: GPT_comparison.png}
    \vspace{-0.1 in}
\end{figure}

\subsubsection{Comparison on path tracking}

By adjusting the desired angular momenta $\bar{L}_{x,S}$ and $\bar{L}_{y,S}$, our framework enables the robot to track the desired position trajectory of its base (i.e., trunk). 
Accurate base-position tracking is critical for collision avoidance in real-world DRS environments, such as ships, which narrow walkways are common.

Using the mapping from the horizontal CoM velocities to the contact angular momenta given in~\eqref{equ:ALIPDRS} and \eqref{equ:ALIPDRS_lateral}, and by including a position-based feedback term, the desired contact angular momenta $\bar{L}_{x,S}$ and $\bar{L}_{y,S}$ are adjusted as follows:
\begin{equation*}
    \begin{aligned}
        \bar{L}_{x,S} &= K_y ( y_b - y_{b,d} )
        + m H \dot{y}_{b,d}~\text{and}~
        \\
        \bar{L}_{y,S} &= K_x ( x_b - x_{b,d} )
        + m H \dot{x}_{b,d},
    \end{aligned}
\end{equation*}
where $x_{b}$ and $y_{b}$ represent the actual base positions along the $x$- and $y$-axes of the DRS frame, respectively,
$x_{b,d}$ and $y_{b,d}$ denotes the corresponding desired positions,
and $K_x$ and $K_y$ are control gains assigned by the user.

Figure~\ref{fig: GPT_comparison.png} depicts the base-tracking performance of the Digit robot navigating through a narrow pathway under Case D. The DRS framework allows Digit to closely track both the desired walking path and the desired forward velocity within the DRS frame. However, the robot fails to stay close to the desired path under the baseline framework.

 \vspace{-0.1 in}
\subsection{Robustness Assessment}
The robustness of the proposed framework is assessed under various uncertainties.
\subsubsection{Sudden push}
\begin{figure}[t]
    \centering
\includegraphics[width=1\linewidth]{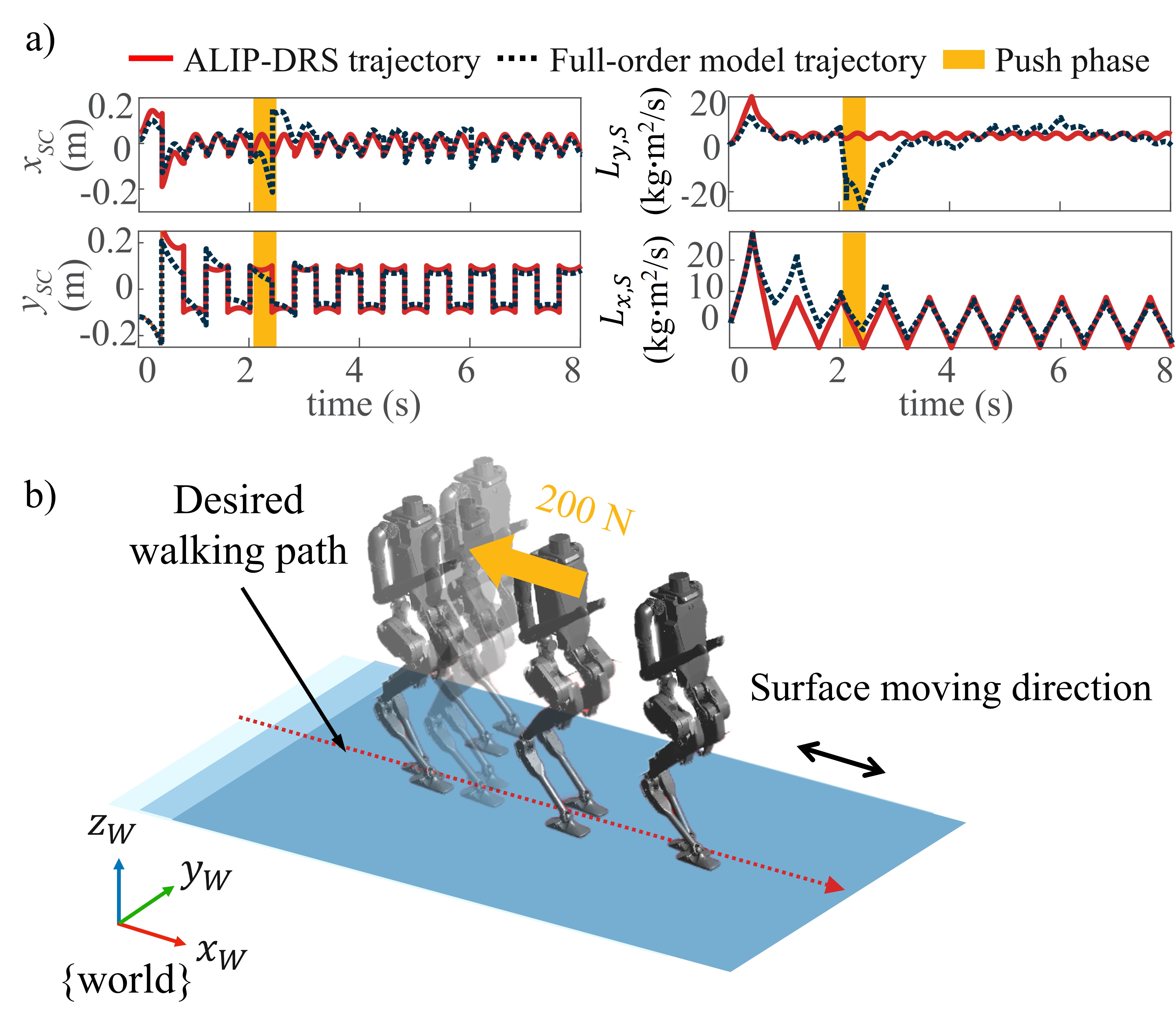}
 \vspace{-0.35 in}
    \caption{Robustness evaluation results of the proposed framework under an unknown sudden push lasting 0.1 seconds,
    with DRS motion and desired robot velocity consistent with Case A, as detailed in Tables~\ref{Table: DRS list} and \ref{table:speed regulation due to DRS motion uncertainty}.
    a) Time evolution of the CoM position and contact angular momentum trajectories, demonstrating fast convergence following the push. b) Illustration of stable robot walking before, during, and after push.}
    \label{fig: caseA_DRS_int_regulated_sudden_push}
     \vspace{-0.1 in}
\end{figure}

An unknown sudden force of 200 N is horizontally applied to Digit's chest under Case A. As shown in Fig.~\ref{fig: caseA_DRS_int_regulated_sudden_push}, the Digit robot is initially heavily affected by this unexpected push. However, after taking four steps, Digit is able to converge back to a close neighborhood of the desired trajectories, demonstrating the robustness of the proposed framework against unknown pushes.

\begin{figure}[t]
    \centering
    \includegraphics[width=1\linewidth]{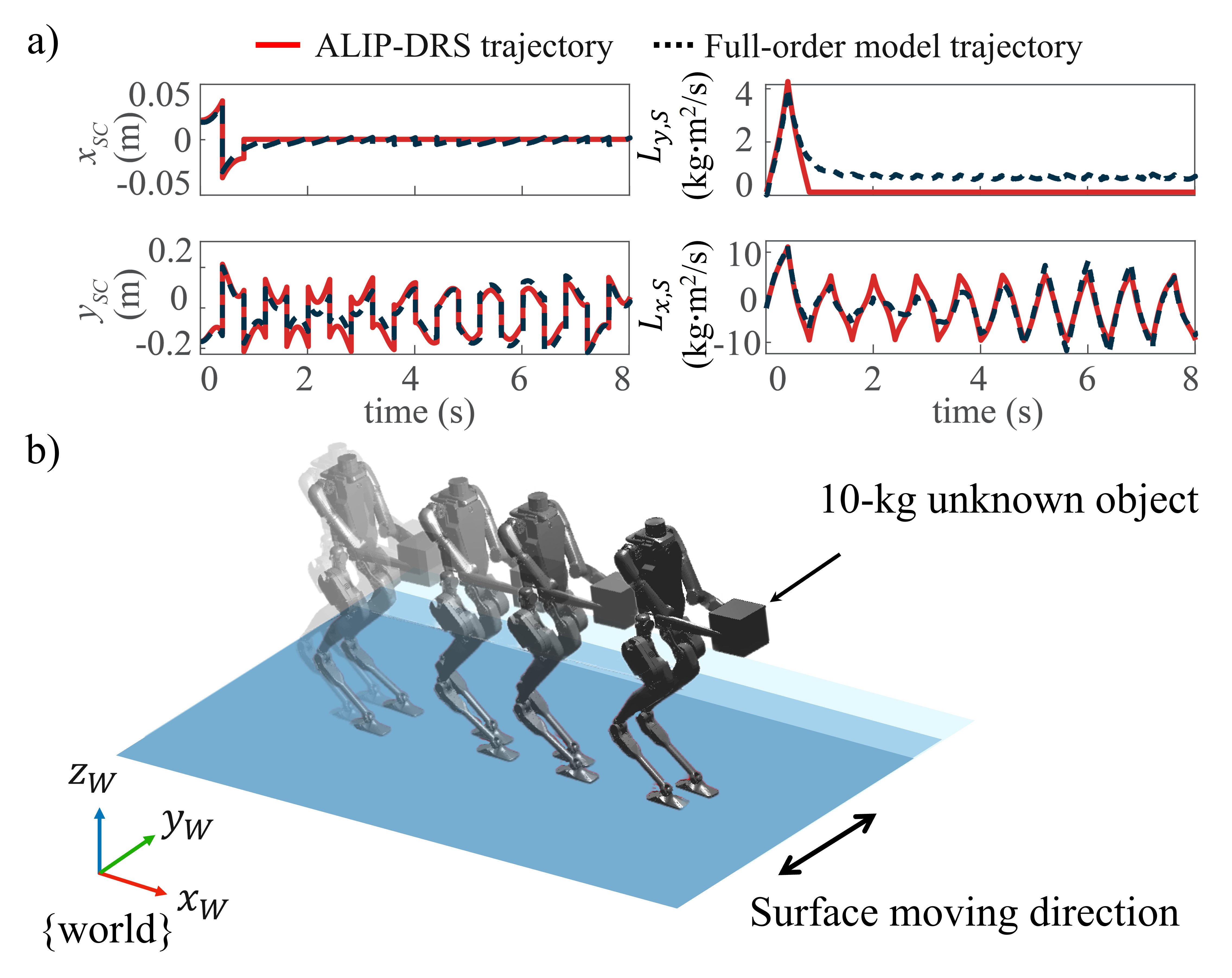}
    \vspace{-0.35 in}
    \caption{Robustness evaluation results under an unknown $10$-kg load,
    with DRS motion and desired robot velocity matching Case C, as outlined in Tables~\ref{Table: DRS list} and \ref{table:speed regulation due to DRS motion uncertainty}. Subplot a) displays the time evolution of the CoM position ($x_{SC}$, $y_{SC}$) and contact angular momenta ($L_{x,S}$, $L_{y,S}$) for both the full-order robot model and the ALIP-DRS model, highlighting their close alignment and ALIP-DRS's accuracy even with an unknown external load.
    Subplot b) illustrates that the robot maintains stable walking despite forward position drift induced by the unknown load.}
    \label{fig: caseC_DRS_int_regulated_10kg}
    \vspace{-0.1 in}
\end{figure}

\subsubsection{Unknown load}

Figure~\ref{fig: caseC_DRS_int_regulated_10kg} presents the results for Case C, where the robot is carrying a box with an unknown weight of 10 kg. The desired behavior for the robot is to walk in place on the DRS. Due to the substantial unknown load, the actual contact angular momentum $L_{y,S}$ shows a constant error of $0.7$ kg$\cdot$m$^2$/s, as depicted in subplot a. This results in a forward position drift, as illustrated in subplot b. Still, the robot maintains stable walking throughout the simulation, despite the significant unknown load.

\subsubsection{DRS motion uncertainty}
To evaluate the robustness of the proposed framework under inaccurately known surface motions, the framework is fed with incorrect DRS motion under Case A. The tested inaccurate DRS motion $ \bar{x}_s(t)$ is:
\begin{equation}
    \bar{x}_s(t) = (0.04+\delta_A)\cos(\tfrac{2\pi}{0.4}(t+\delta_t)),
\end{equation}
where $\delta_A$ and $\delta_t$ are unknown offsets.

Table~\ref{table:speed regulation due to DRS motion uncertainty} summarizes the speed regulation performance under different levels of DRS motion uncertainty in amplitude $\delta_A$ and phase $\delta_t$.
Note that the maximum values of the uncertainties are relatively substantial, with $\delta_A = 0.04$ m, which is the same as the nominal DRS motion amplitude, and $\delta_t = 0.4$ s, which equals one walking-step duration.
The shaded cells contain the actual walking speed corresponding to different uncertainty levels. In Case A, the desired forward speed is $0.1$ m/s.
Without any uncertainties ($\delta_A = \delta_t = 0$), the actual forward speed is $0.1005$ m/s, demonstrating the close alignment with the desired value. Although the speed regulation performance degrades as the DRS motion uncertainty increases, Digit maintains stable walking in all cases.

\begin{table}[t]
\caption{Actual walking speed (m/s) under various DRS motion uncertainties $\delta_A$ and $\delta_t$}
\vspace{-0.1 in}
\small
  \centering
  \begin{tabular}{|c|c|c|c|c|}
    \hline
    \diagbox[width=5em]{$\delta_A$(m)}{$\delta_t$(s)} & 0 & 0.13 & 0.26 &  0.4\\
    \hline
    0 &  \cellcolor{blue!10}0.1005  &  \cellcolor{blue!10} 0.5560 &  \cellcolor{blue!10}0.6338 & \cellcolor{blue!10}0.2837\\
    \hline
    0.013 & \cellcolor{blue!10} -0.0007 & \cellcolor{blue!10} 0.5856 &  \cellcolor{blue!10}0.7521 & \cellcolor{blue!10} -0.0150 \\
    \hline
    0.026 & \cellcolor{blue!10}-0.1024 & \cellcolor{blue!10} 0.5259 &  \cellcolor{blue!10}0.7013 &  \cellcolor{blue!10}-0.1703 \\
    \hline
    0.04 &  \cellcolor{blue!10}-0.1145 &  \cellcolor{blue!10}0.6302 &  \cellcolor{blue!10}0.7408 & \cellcolor{blue!10} -0.2001 \\
    \hline
  \end{tabular}
  \label{table:speed regulation due to DRS motion uncertainty}
  \vspace{-0.1 in}
\end{table}

\section{Hardware Experiment Validation}
\label{Section: experiment}
This section presents the hardware experiment results that demonstrate the stability, tracking performance, and robustness of underactuated walking under the proposed DRS framework. A video of the experiment is available at: \href{https://youtu.be/NtAT0DFtMCY}{https://youtu.be/NtAT0DFtMCY}.

Due to hardware constraints, such as the limited surface area and motion capabilities of the physical DRS, the locomotion tasks and DRS motions implemented in the experiments differ from those in the simulation cases. Still, the framework is tested under different DRS motion frequencies and directions, and is also evaluated in the presence of various uncertainties, including sudden pushes and uncertain surface motions.

  \vspace{-0.1 in}
\subsection{Experiment Setup}
\label{Sec: setup}

\begin{figure}[t]
    \centering
    \includegraphics[width=0.8\linewidth]{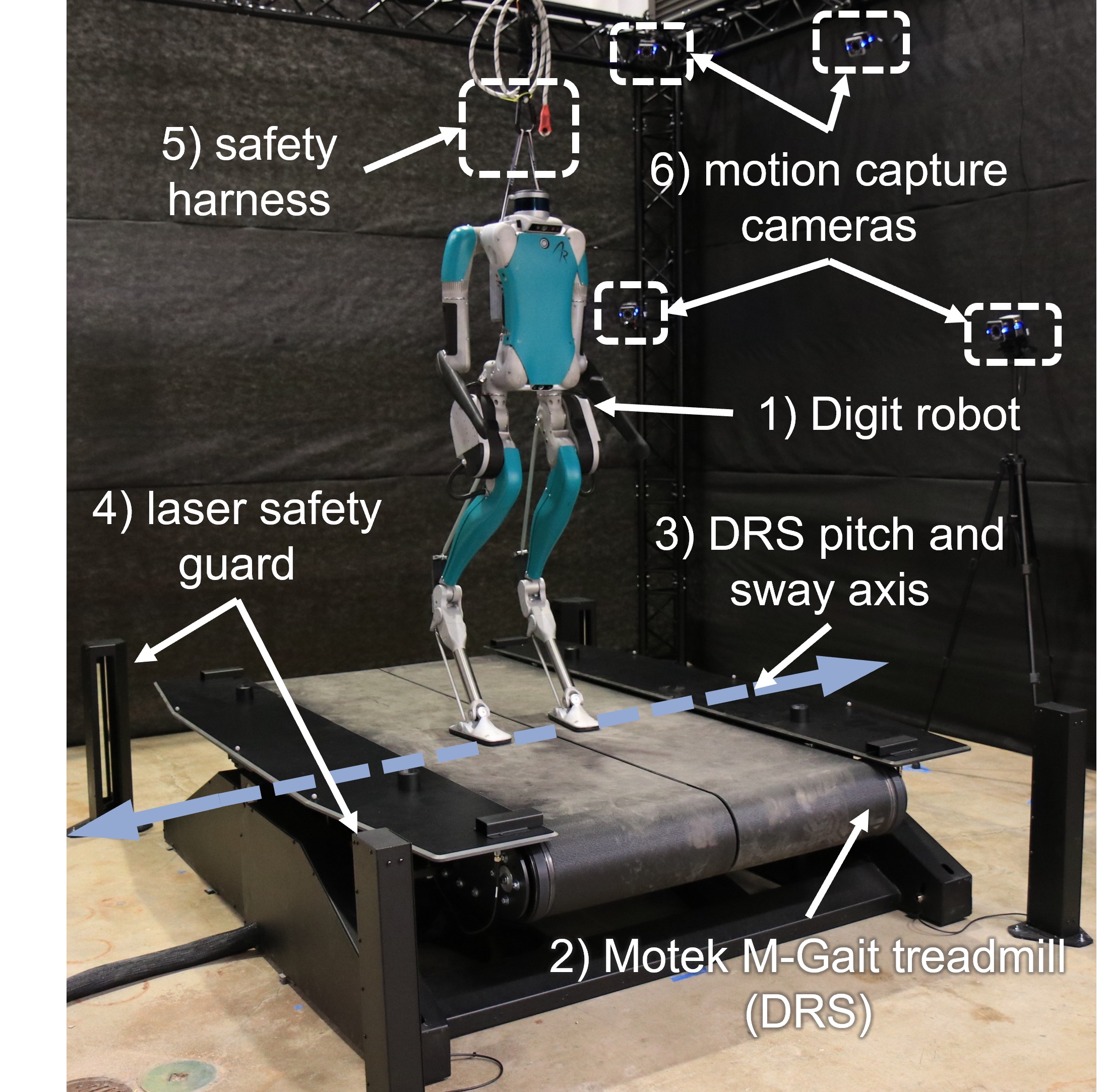}
    \caption{The experiment setup includes: 1) Digit robot, 2) Motek M-Gait treadmill (serving as DRS), equipped with 3) a pitch and sway axis, 4) a laser safety guard, 5) safety harness, and 6) motion capture cameras.}
    \label{fig: experiment setup}
     \vspace{-0.1 in}
\end{figure}

\begin{table}[t]
\centering
\caption{{Experiment Cases}} 
\vspace{-0.1 in}
\small
\begin{tabular}{ p{0.8cm}|p{3.2cm}|p{2.0cm}}
\hline
\hline
\centering
Cases  & \centering DRS expressions (m)&$\bar{L}_{y,S}$ ($\text{kgm}^2/\text{s}$)\\
\hline
\hline
\centering A &  
 \centering $y_S(t)=0.04\cos(\frac{2\pi}{6.8}t)$ &~~~~~~~0\\
 \hline
 \centering B &  
 \centering $y_S(t)=0.04\cos(\frac{2\pi}{5.6}t)$ &~~~~~~~0
 \\
\hline
\centering C &  
 \centering $x_S(t)=0.04\cos(\frac{2\pi}{6.8}t)$ &~~~~~~~0
 \\
 \hline
\centering D &  
 \centering $x_S(t)=0.04\cos(\frac{2\pi}{5.6}t)$ &~~~~~~~0
 \\
  \hline
\end{tabular}
\vspace{-0.15 in}
\label{Table: experiment cases}
\end{table}

\subsubsection{DRS motions}
A Motek M-Gait treadmill is used as the DRS (Fig.~\ref{fig: experiment setup}). 
Due to the movement limits of this treadmill, the experimentally implemented DRS motions are less aggressive than those in the simulations.
This treadmill's walking area is 1 m$\times$2 m (width$\times$length), and has a maximum sway amplitude of $5$ cm.
As the treadmill is not capable of moving with a period as short as the robot gait cycle, which is approximately $0.4$ s, we focus on testing DRS motions with a period multiple times that of the gait.
Also, the treadmill can execute linear motions with a changing direction only along a single axis, which is the ``sway axis'' in Fig.~\ref{fig: experiment setup}.
Thus, to generate DRS motions in the frontal (or sagittal) planes of the Digit robot, the robot is placed to face perpendicular (or parallel) to the treadmill's sway axis.

\subsubsection{Experiment cases}
To evaluate the robot performance under different surface motions, four experiment cases are considered, as listed in Table~\ref{Table: experiment cases}.
In all cases, the desired contact angular momentum in the sagittal plane, $\bar{L}_{y,S}$, is set to zero to enable walking-in-place for ensuring that the robot moves within the limited walking region of the treadmill.

\subsubsection{Implementation of the proposed DRS control framework}
The proposed DRS framework is implemented on the Digit robot using C++ within the ROS platform.
To inform the control framework with the robot's trunk pose and linear velocity, which are not directly measured by Digit's onboard sensors, we utilize Digit's proprietary state estimator considering its reasonable accuracy under the tested DRS motions.

For the lower-layer controller, instead of employing input-output linearizing control as in the simulations, the proposed inverse kinematics control is implemented.
To ensure a sufficient accuracy in tracking the desired full-body trajectories $\boldsymbol{\phi}$, the control gain $\boldsymbol{\kmat}$ is set as a diagonal matrix with $[10,30,30,5,18,45,$ $20,0.01,7,10,\textbf{1}_{1 \times 8}]$ as its diagonal.
Here, $\boldsymbol{\star}_{m \times n}$ denotes an ${m \times n}$ matrix whose elements all take the value of $\star$.
Furthermore, the proportional gain $\bar{\mathbf{K}}_p$ in~\eqref{eq:ik-pd-gains} is chosen as a diagonal matrix whose diagonal values depend on the support foot. This diagonal is $100 \cdot [\textbf{25}_{1\times 2}, 35, 25, \textbf{5}_{1 \times 4},\textbf{25}_{1 \times 2}, 35, \textbf{25}_{1 \times 3}, \textbf{5}_{1 \times 4}]$ and $100 \cdot [\textbf{25}_{1\times 2}, 35, \textbf{25}_{1 \times 3}, \textbf{5}_{1 \times 4},\textbf{25}_{1 \times 2}, 35, 25, \textbf{5}_{1 \times 4}]$ for the left- and right-leg-in-support phases, respectively.
Similar to the tuning of $\boldsymbol{\kmat}$, $\bar{\mathbf{K}}_p$ is set to achieve accurate tracking.
The derivative gain $\bar{\mathbf{K}}_d$ is set to zero because Digit's internal joint-space impedance controller, which is provided by the manufacturer, acts as a derivative control term.

The same set of control gains are used across all four cases to
demonstrate the effectiveness of our proposed framework on hardware without extensive parameter tuning.

\vspace{-0.1 in}
\subsection{Experiment Results}
This subsection presents the experimental results about walking stability, trajectory tracking, and robustness.

\begin{figure}[t]
    \centering
    \includegraphics[width=1\linewidth]{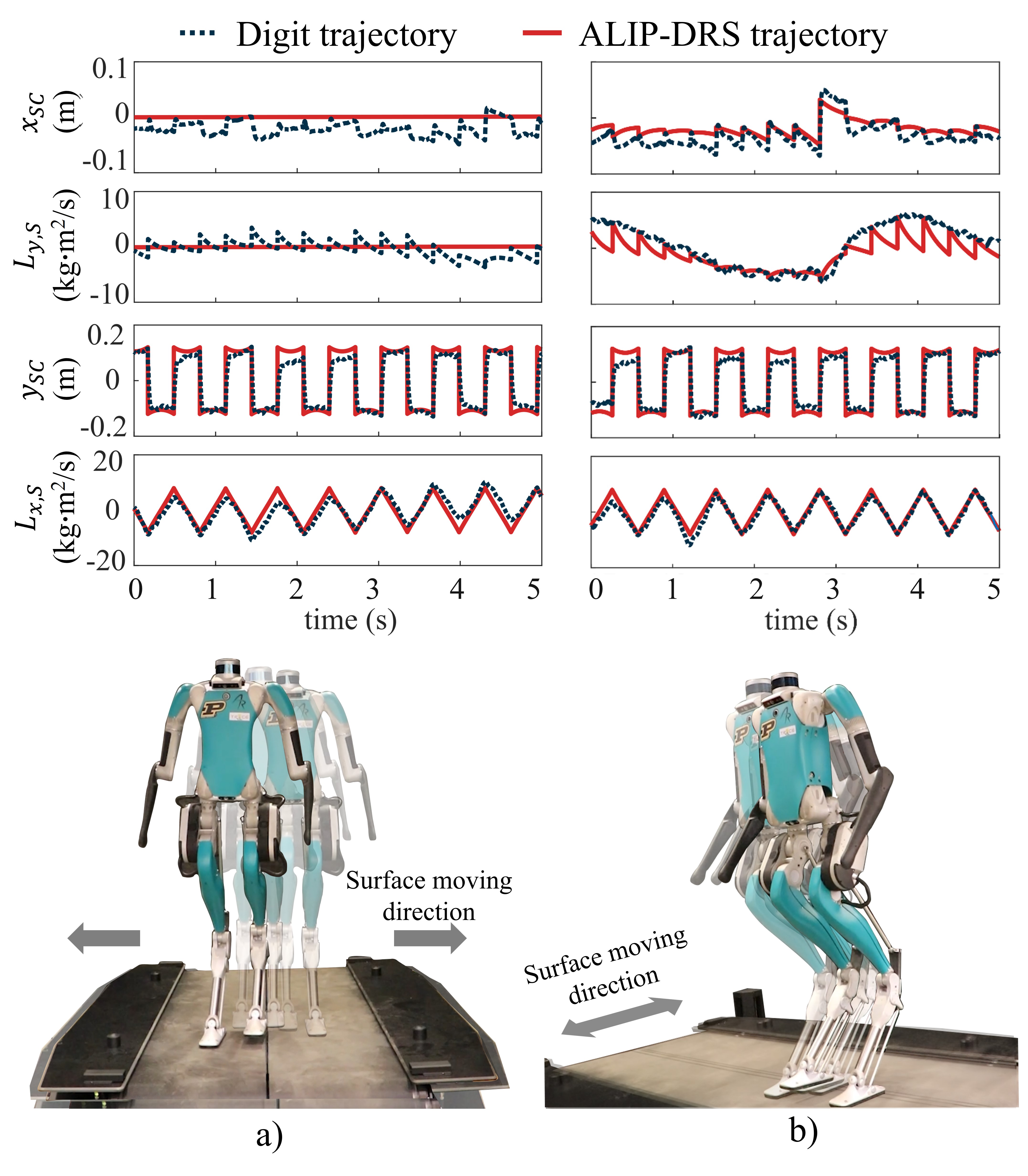}
    \vspace{-0.4 in}
    \caption{Experimental trajectory tracking results under two different DRS motions: a) Case B and b) Case D. In Case B, where the desired robot speed is zero (stepping in place), the desired ALIP-DRS trajectories for $x_{SC}$ and $L_{y,S}$ are set to zero. Digit's actual trajectories closely match the ALIP-DRS trajectories, demonstrating stable walking and ALIP-DRS's accuracy.}
    \label{fig:exp_case_B_D.png}
    \vspace{-0.1 in}
\end{figure}

\subsubsection{Walking stability and trajectory tracking of unactuated variables}

Figure~\ref{fig:exp_case_B_D.png} illustrates the trajectory tracking results for the CoM position ($x_{SC}$, $y_{SC}$) and contact angular momenta ($L_{x,S}$, $L_{y,S}$) in Cases B and D, where the DRS sways in the robot's lateral and sagittal planes, respectively.
As described in Secs.~\ref{section: Hierarchical Planning and Control Framework} and~\ref{Section: Stability},
these variables are not directly commanded by the lower-layer controller and correspond to the unactuated state $\mathbf{x}_{\eta}$. 
The figure shows that these unactuated variables remain close to the ALIP-DRS's solution, which serves as their desired trajectory as explained in Sec.~\ref{Section: Stability}.
This confirms that the proposed framework ensures stable underactuated walking during the various DRS motions in Cases A-D, and validates the accurate tracking of the actual unactuated variables.

\begin{figure}[t]
    \centering
    \includegraphics[width=1\linewidth]{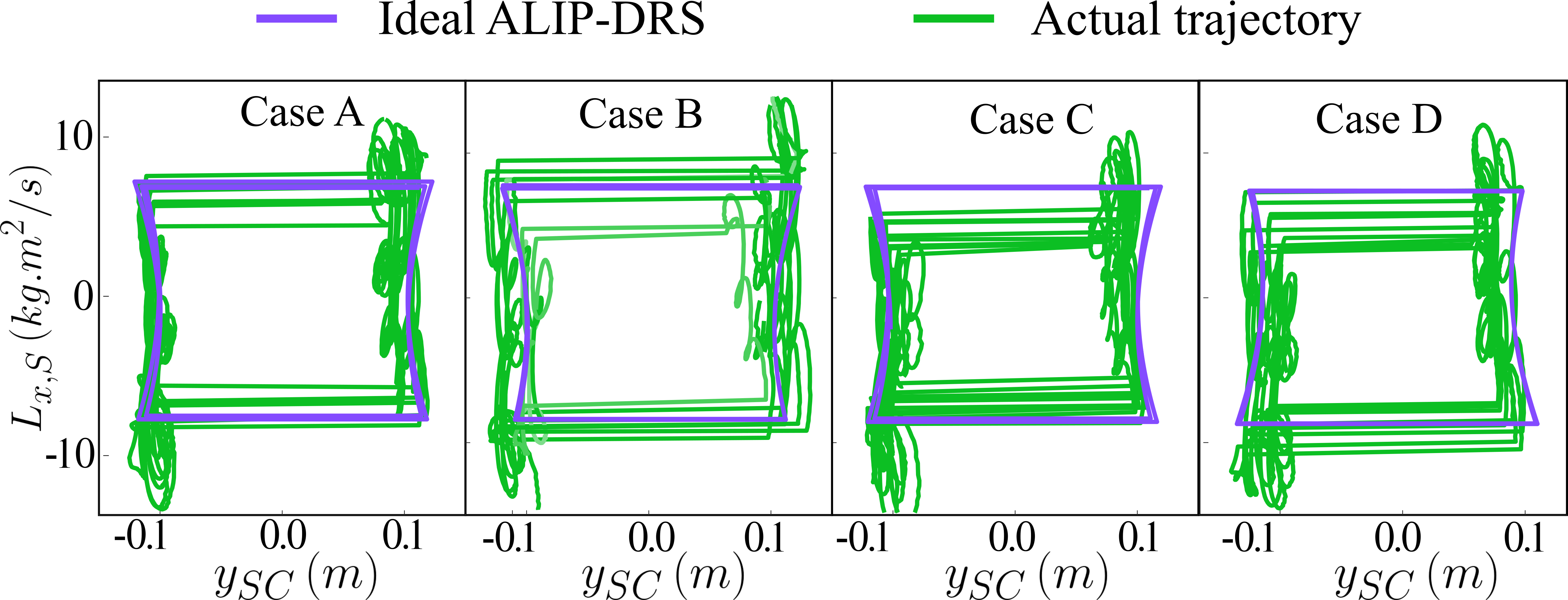}
    \vspace{-0.25 in}
    \caption{Experimental results of phase portraits under Cases A-D, illustrating the close alignment between the actual Digit motion and the ideal ALIP-DRS trajectories. This alignment further confirms the reasonable accuracy of ALIP-DRS and validates that the proposed framework enables reliable trajectory tracking for the unactuated state $\mathbf{x}_{\eta}$ across various DRS motions.}
    \label{fig:CaseABC.png}
\end{figure}

Additionally, Fig.~\ref{fig:CaseABC.png} shows the sagittal-plane phase portrait of both the ideal ALIP-DRS model and the actual robot under Cases A-D.
Note that the ideal ALIP-DRS phase portrait exhibits multiple loops, as the solution period $T_{sys}$ of the ALIP-DRS is several times the walking-step duration $T_{step}$.
The close alignment between the two sets of phase portraits further demonstrates the walking stability and indicates the relative accuracy of the ALIP-DRS model.

\begin{figure}[t]
    \centering
    \includegraphics[width=0.68\linewidth]{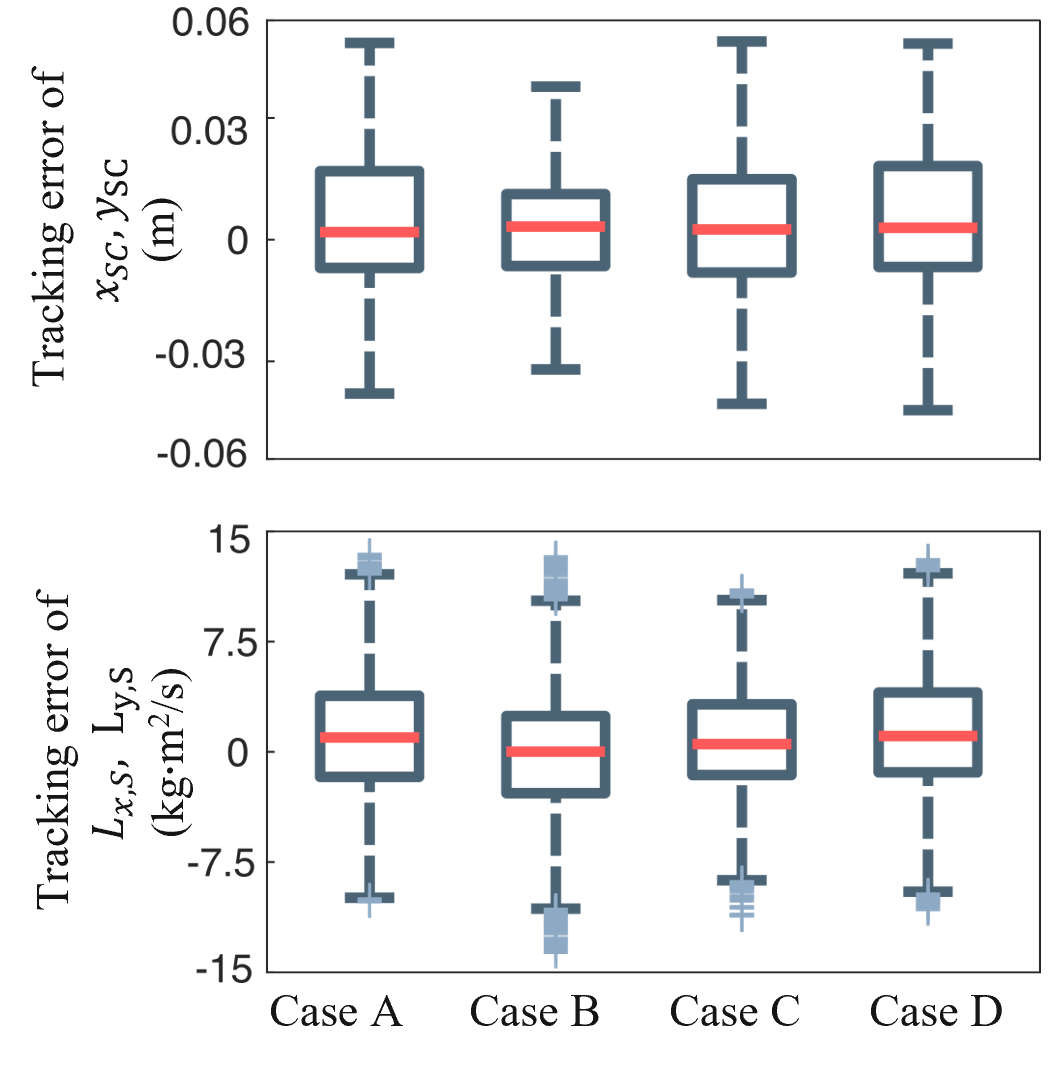}
    \vspace{-0.15 in}
    \caption{Experimental results showing trajectory discrepancies between the actual robot and the ALIP-DRS model from Cases A to D. The top and bottom graphs display the combined errors in the CoM positions ($x_{SC}$, $y_{SC}$) and angular momenta ($L_{y,S}$, $L_{x,S}$), respectively.}
    \label{fig:box_plot}
    \vspace{-0.2 in}
\end{figure}

The trajectory tracking errors of the unactuated variables under Cases A-D are presented as box-and-whisker plots in Fig.~\ref{fig:box_plot}. In each plot, the red central mark represents the mean value, while the bottom and top edges of the box correspond to the 25$^\text{th}$ and 75$^\text{th}$ percentiles, respectively. Outliers are indicated by light blue ``$+$'' symbols. The mean values of the CoM position and contact angular momentum errors are close to zero across all four cases, with relatively small upper and lower quartiles that show consistent values across the cases. These relatively small tracking errors align with the results shown in Figs.~\ref{fig:exp_case_B_D.png} and~\ref{fig:CaseABC.png}, demonstrating the high tracking accuracy achieved by the proposed framework. 

\begin{figure}
    \centering
    \includegraphics[width=0.9\linewidth]{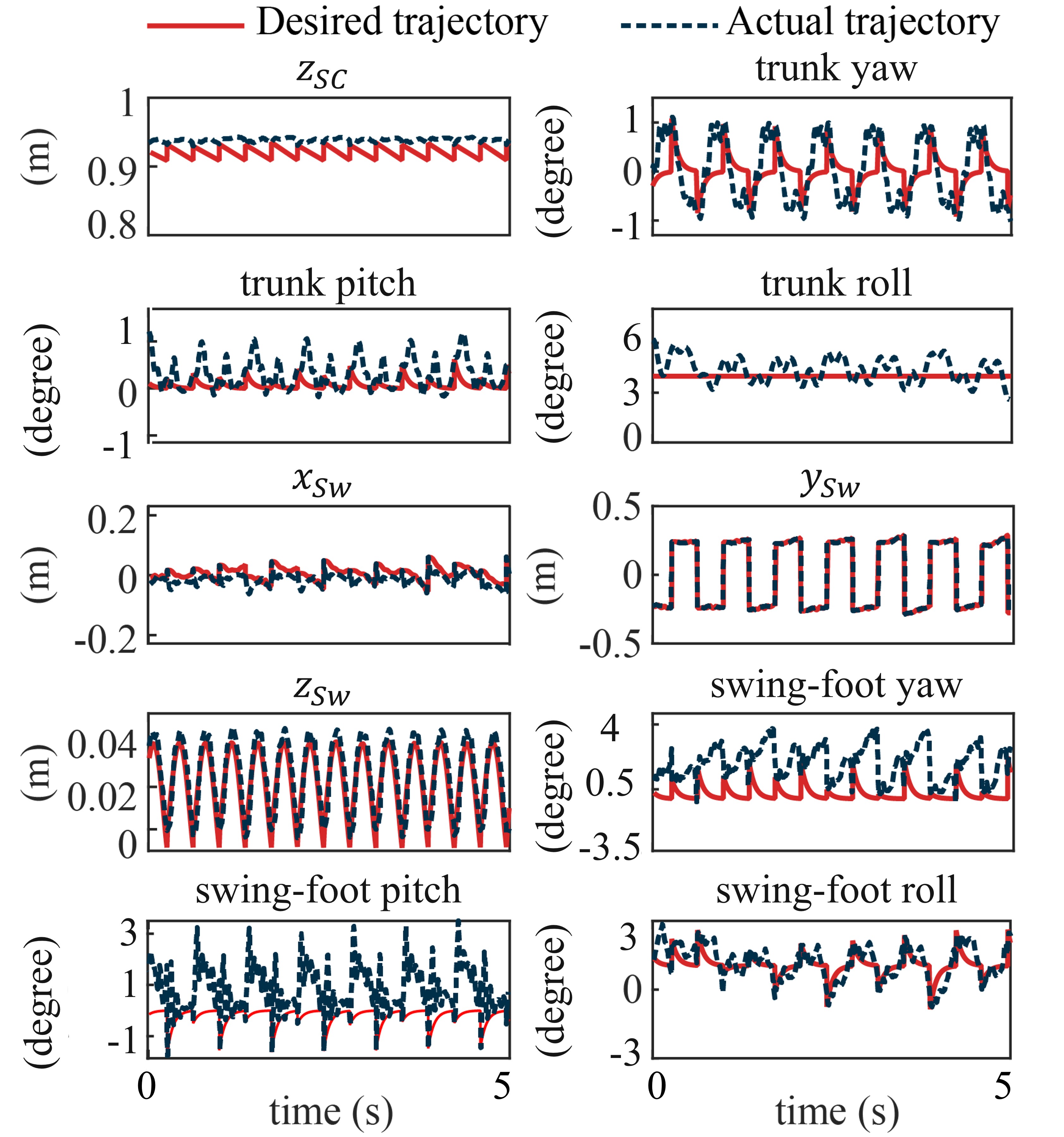}
    \vspace{-0.1 in}
    \caption{Experiment results of task-space tracking for the directly commanded state $\mathbf{x}_{h}$ under Case B. The CoM height $z_{SC}$ closely matches the desired height, indicating that the actual robot effectively adheres to the constant CoM height assumption of the ALIP-DRS model. Also, the relatively small tracking error in the desired swing-foot pose confirms accurate foot placement, demonstrating faithful execution of the discrete footstep control designed based on the ALIP-DRS model.}
    \label{fig:task_space_mujoco}
    \vspace{-0.25 in}
\end{figure}

\subsubsection{Trajectory tracking for directly commanded variables}

The full-body trajectory tracking results for the directed commanded control variables $\mathbf{h}_c$ are provided in Fig.~\ref{fig:task_space_mujoco}.
The CoM height $z_{SC}$ shows slight oscillations around the desired height of $92$ cm, with an average tracking error of $1$ cm. Note that the desired CoM height trajectory in red is the modified desired trajectory. 
Its value at the beginning of a walking step is matched with the actual CoM height, which helps reduce the jumps in the tracking error immediately after the foot-landing events and thus avoids excessively high joint torques. 
The swing-foot position tracking is accurate in all directions, with a peak tracking error of $5.5$ cm and an average tracking error of $1.2$ cm.
While the tracking of the swing-foot orientation is not highly accurate, the average error in each direction is within $2^{\degree}$, which is sufficient to ensure a proper foot posture before swing-foot touchdowns.

\subsubsection{Robustness to pushes and unknown surface motion}

The robustness of the proposed DRS framework is assessed against three types of external disturbances, which are: 
(i) sudden pushes applied by a long stick; (ii) uncertain forward and lateral DRS motions induced by orienting Digit away from its nominal facing direction by $45^\circ$; and (iii) unknown pitching DRS motion with an amplitude of $1^\circ$ and frequency of $10$ s. 
These uncertainties are tested during a treadmill sway motion with an amplitude of $0.05$ m and a period of $4$ s, which is the same as the treadmill motion during which Digit's proprietary controller failed to sustain stable walking as shown in the top row of Fig.~\ref{Fig:digit_fail.png}.

\begin{figure}[t]
    \centering
    \includegraphics[width=1\linewidth]
    {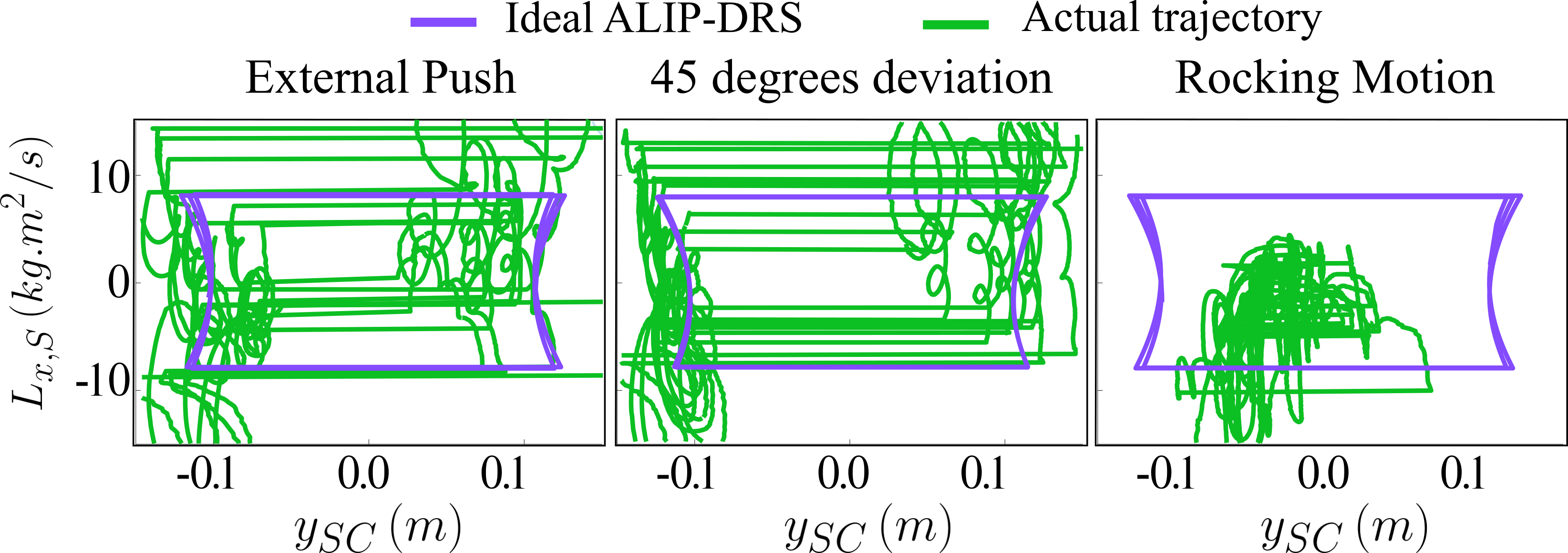}
        \vspace{-0.2 in}
    \caption{Experiment results of phase portraits
    under varying uncertainties, including external pushes (left), unknown forward motion and uncertain lateral motion of the DRS (middle),  and unknown DRS pitch motion (right).}
    \label{fig:CaseAggresive}
    \vspace{-0.1 in}
\end{figure}

\begin{figure*}[h]
    \centering
    \includegraphics[width=1\linewidth]{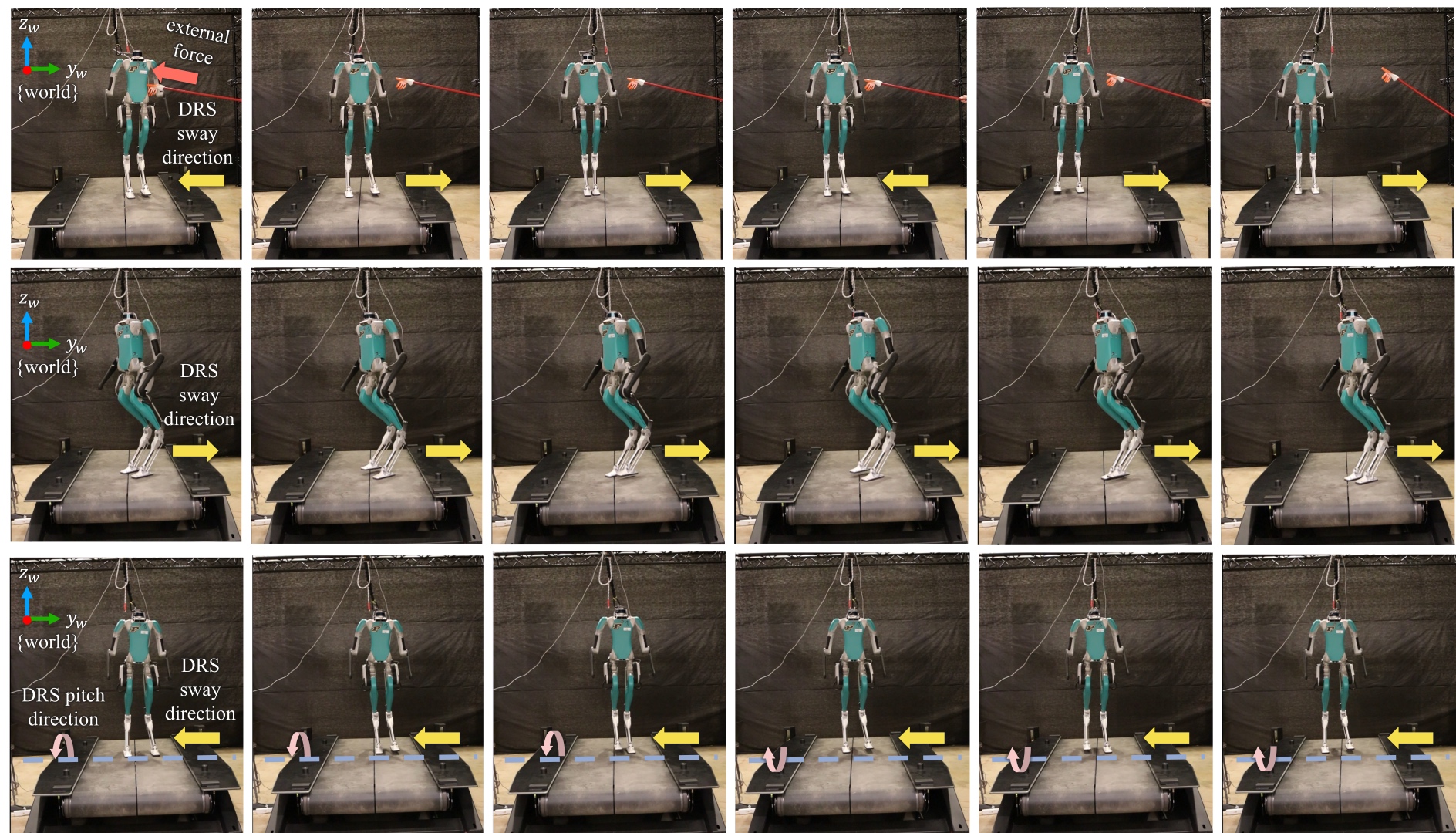}
    \vspace{-0.25 in}
    \caption{Illustrations of the Digit and DRS motions under different uncertainty tests.
    Top: Digit steps in place on swaying DRS while experiencing unknown sudden pushes. 
    Middle: Due to Digit's heading direction at $45^{\degree}$ to the DRS sway direction, it faces uncertain DRS motions in both its sagittal and frontal planes.
   Bottom: Digit steps in place experiencing both sway and unknown pitch motions of the DRS. Note that although the treadmill sways left and right in all experiments, the snapshots in the middle and bottom rows capture moments when the treadmill's movement points in the same direction.}
    \label{fig: aggressive_exp}
    \vspace{-0.05 in}
\end{figure*}

The phase portraits of the lateral CoM position $y_{SC}$ and contact angular momentum $L_{x,S}$ under these three types of uncertainties are given in Fig.~\ref{fig:CaseAggresive}.
Figure~\ref{fig: aggressive_exp} shows the corresponding snapshots of Digit walking, highlighting that the proposed control framework maintains stable walking despite the uncertainties.
The phase portraits in Fig.~\ref{fig:CaseAggresive} indicate that in the presence of the uncertainties, the robot's actual trajectories exhibit a larger deviation from the ideal phase portraits of the ALIP-DRS, compared to the scenarios without these uncertainties shown in Fig.~\ref{fig:CaseABC.png}. 
This is primarily due to the higher model inaccuracies induced by the unknown surface motions and external force disturbances.

In particular, under the unknown treadmill rocking motion, the robot's actual phase portraits notably deviate from the ALIP-DRS's, as shown in Fig.~\ref{fig:CaseAggresive}.
This deviation may be due to the large modeling errors caused by the unmodeled vertical ground motions.
The relative CoM height above the support point is assumed to be constant in the ALIP-DRS model, but a rocking motion can introduce a noticeable vertical CoM velocity.
Moreover, the treadmill's rocking motion can introduce non-negligible centroidal momentum $\mathbf{L}_{CoM}$, which is assumed to be zero in the ALIP-DRS model.
Still, despite the degraded tracking performance compared to the nominal cases, the robot successfully maintains stable walking and steps in place near the treadmill's center under all tested uncertainties.
Potential solutions that address complex surface motions based on the current work are discussed in the next section.

\section{DISCUSSION}
\label{section: discussion}
This sections discusses the capabilities and limitations of the proposed control approach, along with potential directions for future research.

To control underactuated bipedal walking on swaying DRS, this study introduces a new reduced-order robot model, ALIP-DRS, to approximate the associated hybrid, nonlinear, time-varying, nonhomogeneous unactuated robot dynamics.
The key novelty of this model lies in its explicit consideration of the DRS sway motion.
Due to this sway, the ALIP-DRS is nonhomogeneous, fundamentally distinguishing it from classical LIP models and their variations, such as the time-invariant H-LIP\cite{xiong20223} and ALIP~\cite{gong2022zero} and the time-varying HT-LIP~\cite{iqbal2023asymptotic} and variable-height LIP~\cite{caron2019capturability}, which are all homogeneous.
The close alignment between Digit's unactuated movement variables and the ALIP-DRS's trajectories is confirmed through both simulations (Figs.~\ref{fig: matlab_traj}, \ref{fig: caseA_DRS_int_regulated_sudden_push}, and \ref{fig: caseC_DRS_int_regulated_10kg}) and hardware experiments (Figs.~\ref{fig:exp_case_B_D.png}, \ref{fig:CaseABC.png}, \ref{fig:box_plot}, and \ref{fig:CaseAggresive}),
highlighting the reasonable accuracy of the ALIP-DRS.
This model's generality allows for extensions to other legged robot platforms beyond humanoids.

In addition, the stability for the nonhomogeneous closed-loop ALIP-DRS model under a discrete footstep control law is provably verified based on the stability of its homogeneous portion.
The previously developed stability conditions for existing LIP models, which are homogeneous, are not directly applicable to the nonhomogeneous ALIP-DRS. 
The proposed stability analysis, based on the general stability theory of linear impulsive nonhomogeneous systems, can be generalized to continuous-phase control laws for pendulum models as well as alternative state variable choices beyond those employed in this study.

The study also introduces a hierarchical control framework that incorporates the ALIP-DRS's foot-placement controller as its higher-layer footstep planner. By stabilizing the ALIP-DRS, the discrete control law indirectly stabilizes the actual unactuated dynamics of the robot during walking. The stability conditions for the complete full-order system under the control framework are also provided. Due to their general applicability, this stability analysis can be extended to prior hierarchical control methods based on LIP models~\cite{xiong20223, gong2022zero}. Experimental results confirm that the proposed framework effectively achieves stable underactuated walking under various DRS sway motions and uncertainties.

Building on these theoretical findings, future work will extend the stability conditions of the ALIP-DRS model to accommodate DRS motions in both vertical and horizontal directions. In the presence of vertical and horizontal surface motions, the hybrid ALIP-DRS model~\eqref{eq: open loop ALIP-DRS} will be updated with a time-varying matrix $\mathbf{A}_x$, whose explicit time dependence is induced by vertical surface motion~\cite{iqbal2023analytical}. The stability conditions for this generalized ALIP-DRS model will be developed by integrating the stability analysis from this study with that for locomotion control on vertical surface motions. This future research will address the observed deviation between the actual robot state and the ideal ALIP-DRS model during unknown DRS pitching motions (Fig.~\ref{fig:CaseAggresive}). Moreover, future efforts will explore real-time surface motion estimation using onboard robot sensors, so as to enable the proposed hierarchical control framework to adapt more effectively to unknown surface motions and further enhance locomotion robustness.


\section{CONCLUSION}
\label{section: conclusions}

This paper has introduced a real-time, provably stabilizing control approach that employs a new reduced-order robot model to enable underactuated bipedal walking on a swaying dynamic rigid surface (DRS). The classical angular-momentum-based linear inverted pendulum (ALIP) model was analytically extended to explicitly consider the time-varying DRS motion, resulting in a time-varying, nonhomogeneous system, which was referred to as the ALIP-DRS model. A discrete-time footstep control strategy was constructed, which was proven to be exponentially stabilizing for the ALIP-DRS model.
With the ALIP-DRS footstep controller serving as a motion planner that generates the desired swing-foot landing positions in real-time, a hierarchical control framework was developed for the full-order robot model of an underactuated walking robot.
The closed-loop stability conditions for the hybrid full-order control system were constructed by analyzing both the unactuated and fully actuated subsystems under the proposed control framework.
Both numerical simulations and hardware experiments on a Digit humanoid robot confirmed that the proposed ALIP-DRS model is reasonably accurate in describing the unactuated dynamics of the actual robot.
Further, the proposed framework achieved stable walking under various DRS sway motions, and demonstrated high robustness under different uncertainties including sudden pushes and uncertain DRS motions in different directions.

\vspace{-0.1 in}
\section*{Acknowledgment}
The authors would like to thank I-Chia Chang for supporting the hardware experiments and providing constructive comments on this paper.

\vspace{-0.05 in}
\bibliography{Reference1}

\begin{thebibliography}{10}

\bibitem{kim2007walking}
J.-Y. Kim, I.-W. Park, and J.-H. Oh, ``Walking control algorithm of biped humanoid robot on uneven and inclined floor,'' {\em J. Intel. Rob. Syst.}, vol.~48, pp.~457--484, 2007.

\bibitem{caron2019stair}
S.~Caron, A.~Kheddar, and O.~Tempier, ``Stair climbing stabilization of the hrp-4 humanoid robot using whole-body admittance control,'' in {\em Proc. IEEE Int. Conf. Rob. Autom.}, pp.~277--283, 2019.

\bibitem{pajon2017walking}
A.~Pajon, S.~Caron, G.~De~Magistri, S.~Miossec, and A.~Kheddar, ``Walking on gravel with soft soles using linear inverted pendulum tracking and reaction force distribution,'' in {\em Proc. IEEE-RAS Int. Conf. Humanoid Rob.}, pp.~432--437, 2017.

\bibitem{bauby2000active}
C.~E. Bauby and A.~D. Kuo, ``Active control of lateral balance in human walking,'' {\em J. Biomech.}, vol.~33, no.~11, pp.~1433--1440, 2000.

\bibitem{kim2018computationally}
D.~Kim, J.~Lee, J.~Ahn, O.~Campbell, H.~Hwang, and L.~Sentis, ``Computationally-robust and efficient prioritized whole-body controller with contact constraints,'' in {\em Proc. IEEE Int. Conf. Intel. Rob. Syst.}, pp.~1--8, 2018.

\bibitem{vukobratovic2006zmp}
M.~Vukobratovi{\'c}, B.~Borovac, and V.~Potkonjak, ``Zmp: A review of some basic misunderstandings,'' {\em Int. J. Humanoid Rob.}, vol.~3, no.~02, pp.~153--175, 2006.

\bibitem{firmani2013theoretical}
F.~Firmani and E.~J. Park, ``Theoretical analysis of the state of balance in bipedal walking,'' {\em J. Biomechan. Engn.}, vol.~135, no.~4, p.~041003, 2013.

\bibitem{kajita20013d}
S.~Kajita, F.~Kanehiro, K.~Kaneko, K.~Yokoi, and H.~Hirukawa, ``The 3{D} linear inverted pendulum mode: A simple modeling for a biped walking pattern generation,'' in {\em Proc. IEEE Int. Conf. Intel. Rob. Syst.}, vol.~1, pp.~239--246, 2001.

\bibitem{tedrake2015closed}
R.~Tedrake, S.~Kuindersma, R.~Deits, and K.~Miura, ``A closed-form solution for real-time zmp gait generation and feedback stabilization,'' in {\em Proc. IEEE-RAS Int. Conf. Humanoid Rob.}, pp.~936--940, 2015.

\bibitem{sugihara20213d}
T.~Sugihara, K.~Imanishi, T.~Yamamoto, and S.~Caron, ``3d biped locomotion control including seamless transition between walking and running via 3d zmp manipulation,'' in {\em Proc. IEEE Int. Conf. Rob. Autom.}, pp.~6258--6263, 2021.

\bibitem{pratt2006capture}
J.~Pratt, J.~Carff, S.~Drakunov, and A.~Goswami, ``Capture point: A step toward humanoid push recovery,'' in {\em Proc. IEEE-RAS Int. Conf. Humanoid Rob.}, pp.~200--207, 2006.

\bibitem{hof2005condition}
A.~L. Hof, M.~Gazendam, and W.~Sinke, ``The condition for dynamic stability,'' {\em J. Biomech.}, vol.~38, no.~1, pp.~1--8, 2005.

\bibitem{englsberger2011bipedal}
J.~Englsberger, C.~Ott, M.~A. Roa, A.~Albu-Sch{\"a}ffer, and G.~Hirzinger, ``Bipedal walking control based on capture point dynamics,'' in {\em Proc. IEEE Int. Conf. Intel. Rob. Syst.}, pp.~4420--4427, 2011.

\bibitem{caron2019capturability}
S.~Caron, A.~Escande, L.~Lanari, and B.~Mallein, ``Capturability-based pattern generation for walking with variable height,'' {\em IEEE Trans. Rob.}, vol.~36, no.~2, pp.~517--536, 2019.

\bibitem{takenaka2009real}
T.~Takenaka, T.~Matsumoto, and T.~Yoshiike, ``Real time motion generation and control for biped robot-1 st report: Walking gait pattern generation,'' in {\em Proc. IEEE Int. Conf. Intel. Rob. Syst.}, pp.~1084--1091, 2009.

\bibitem{englsberger2013three}
J.~Englsberger, C.~Ott, and A.~Albu-Sch{\"a}ffer, ``Three-dimensional bipedal walking control using divergent component of motion,'' in {\em Proc. IEEE Int. Conf. Intel. Rob. Syst.}, pp.~2600--2607, 2013.

\bibitem{englsberger2015three}
J.~Englsberger, C.~Ott, and A.~Albu-Sch{\"a}ffer, ``Three-dimensional bipedal walking control based on divergent component of motion,'' {\em IEEE Trans. Rob.}, vol.~31, no.~2, pp.~355--368, 2015.

\bibitem{gong2022zero}
Y.~Gong and J.~W. Grizzle, ``Zero dynamics, pendulum models, and angular momentum in feedback control of bipedal locomotion,'' {\em ASME J. Dyn. Sys., Meas., Contr.}, vol.~144, no.~12, p.~121006, 2022.

\bibitem{collins2005bipedal}
S.~H. Collins and A.~Ruina, ``A bipedal walking robot with efficient and human-like gait,'' in {\em Proc. IEEE Int. Conf. Rob. Autom.}, pp.~1983--1988, 2005.

\bibitem{reher2021dynamic}
J.~Reher and A.~D. Ames, ``Dynamic walking: Toward agile and efficient bipedal robots,'' {\em Annual Review of Control, Robotics, and Autonomous Systems}, vol.~4, pp.~535--572, 2021.

\bibitem{westervelt2007feedback}
E.~R. Westervelt, C.~Chevallereau, J.~H. Choi, B.~Morris, and J.~W. Grizzle, {\em Feedback control of dynamic bipedal robot locomotion}.
\newblock CRC press, 2007.

\bibitem{poulakakis2009spring}
I.~Poulakakis and J.~W. Grizzle, ``The spring loaded inverted pendulum as the hybrid zero dynamics of an asymmetric hopper,'' {\em IEEE Trans. Autom. Contr.}, vol.~54, no.~8, pp.~1779--1793, 2009.

\bibitem{sreenath2011compliant}
K.~Sreenath, H.-W. Park, I.~Poulakakis, and J.~W. Grizzle, ``A compliant hybrid zero dynamics controller for stable, efficient and fast bipedal walking on mabel,'' {\em Int. J. Rob. Res.}, vol.~30, no.~9, pp.~1170--1193, 2011.

\bibitem{sreenath2013embedding}
K.~Sreenath, H.-W. Park, I.~Poulakakis, and J.~W. Grizzle, ``Embedding active force control within the compliant hybrid zero dynamics to achieve stable, fast running on mabel,'' {\em Int. J. Rob. Res.}, vol.~32, no.~3, pp.~324--345, 2013.

\bibitem{ramezani2014performance}
A.~Ramezani, J.~W. Hurst, K.~Akbari~Hamed, and J.~W. Grizzle, ``Performance analysis and feedback control of atrias, a three-dimensional bipedal robot,'' {\em ASME J. Dyn. Syst. Meas. Contr.}, vol.~136, no.~2, p.~021012, 2014.

\bibitem{Hereid2017FROST}
A.~Hereid and A.~D. Ames, ``F{R}{O}{S}{T}: Fast robot optimization and simulation toolkit,'' in {\em Proc. Int. Conf. Intel. Rob. Syst.}, 2017.

\bibitem{gao2019global}
Y.~Gao and Y.~Gu, ``Global-position tracking control of a fully actuated {NAO} bipedal walking robot,'' in {\em Proc. Amer. Contr. Conf.}, pp.~4596--4601, 2019.

\bibitem{paredes2020dynamic}
V.~Paredes and A.~Hereid, ``Dynamic locomotion of a lower-limb exoskeleton through virtual constraints based zmp regulation,'' in {\em Proc. ASME Dyn. Syst. Contr. Conf.}, vol.~84270, p.~V001T14A001, American Society of Mechanical Engineers, 2020.

\bibitem{wensing2023optimization}
P.~M. Wensing, M.~Posa, Y.~Hu, A.~Escande, N.~Mansard, and A.~Del~Prete, ``Optimization-based control for dynamic legged robots,'' {\em IEEE Trans. Rob.}, 2023.

\bibitem{gao2019dscc}
Y.~Gao and Y.~Gu, ``Global-position tracking control of multi-domain planar bipedal robotic walking,'' in {\em Proc. ASME Dyn. Syst. Contr. Conf.}, vol.~59148, p.~V001T03A009, 2019.

\bibitem{motahar2016composing}
M.~S. Motahar, S.~Veer, and I.~Poulakakis, ``Composing limit cycles for motion planning of 3d bipedal walkers,'' in {\em Proc. IEEE Conf. Dec. Contr.}, pp.~6368--6374, 2016.

\bibitem{hartley2017stabilization}
R.~Hartley, X.~Da, and J.~W. Grizzle, ``Stabilization of 3d underactuated biped robots: Using posture adjustment and gait libraries to reject velocity disturbances,'' in {\em Proc. IEEE Conf. Control Tech. Appl.}, pp.~1262--1269, 2017.

\bibitem{nguyen2017dynamic}
Q.~Nguyen, A.~Agrawal, X.~Da, W.~C. Martin, H.~Geyer, J.~W. Grizzle, and K.~Sreenath, ``Dynamic walking on randomly-varying discrete terrain with one-step preview,'' in {\em Rob Sc. Syst.}, vol.~2, pp.~384--399, 2017.

\bibitem{xiong2019orbit}
X.~Xiong and A.~D. Ames, ``Orbit characterization, stabilization and composition on 3d underactuated bipedal walking via hybrid passive linear inverted pendulum model,'' in {\em Proc. IEEE Int. Conf. Intel. Rob. Syst.}, pp.~4644--4651, 2019.

\bibitem{xiong20223}
X.~Xiong and A.~Ames, ``3-d underactuated bipedal walking via h-lip based gait synthesis and stepping stabilization,'' {\em IEEE Trans. Rob.}, vol.~38, no.~4, pp.~2405--2425, 2022.

\bibitem{xiong2021slip}
X.~Xiong and A.~Ames, ``Slip walking over rough terrain via h-lip stepping and backstepping-barrier function inspired quadratic program,'' {\em IEEE Rob. Autom. L.}, vol.~6, no.~2, pp.~2122--2129, 2021.

\bibitem{dai2022bipedal}
M.~Dai, X.~Xiong, and A.~Ames, ``Bipedal walking on constrained footholds: Momentum regulation via vertical com control,'' in {\em Proc. IEEE Int. Conf. Rob. Autom}, pp.~10435--10441, 2022.

\bibitem{scianca2020mpc}
N.~Scianca, D.~De~Simone, L.~Lanari, and G.~Oriolo, ``Mpc for humanoid gait generation: Stability and feasibility,'' {\em IEEE/ASME Trans. Mechatron.}, vol.~36, no.~4, pp.~1171--1188, 2020.

\bibitem{gibson2022terrain}
G.~Gibson, O.~Dosunmu-Ogunbi, Y.~Gong, and J.~Grizzle, ``Terrain-adaptive, alip-based bipedal locomotion controller via model predictive control and virtual constraints,'' in {\em Proc. IEEE Int. Conf. Intel. Rob. Syst.}, pp.~6724--6731, 2022.

\bibitem{acosta2023bipedal}
B.~Acosta and M.~Posa, ``Bipedal walking on constrained footholds with mpc footstep control,'' in {\em Proc. IEEE-RAS Int. Conf. Humanoid Rob.}, pp.~1--8, 2023.

\bibitem{gu2024robust}
Z.~Gu, Y.~Zhao, Y.~Chen, R.~Guo, J.~K. Leestma, G.~S. Sawicki, and Y.~Zhao, ``Robust-locomotion-by-logic: Perturbation-resilient bipedal locomotion via signal temporal logic guided model predictive control,'' {\em arXiv preprint arXiv:2403.15993}, 2024.

\bibitem{xie2018feedback}
Z.~Xie, G.~Berseth, P.~Clary, J.~Hurst, and M.~Van~de Panne, ``Feedback control for cassie with deep reinforcement learning,'' in {\em Proc. IEEE Int. Conf. Intel. Rob. Syst.}, pp.~1241--1246, 2018.

\bibitem{castillo2021robust}
G.~A. Castillo, B.~Weng, W.~Zhang, and A.~Hereid, ``Robust feedback motion policy design using reinforcement learning on a 3d digit bipedal robot,'' in {\em Proc. IEEE Int. Conf. Intel. Rob. Syst.}, pp.~5136--5143, 2021.

\bibitem{li2021reinforcement}
Z.~Li, X.~Cheng, X.~B. Peng, P.~Abbeel, S.~Levine, G.~Berseth, and K.~Sreenath, ``Reinforcement learning for robust parameterized locomotion control of bipedal robots,'' in {\em Proc. IEEE Int. Conf. Rob. Autom.}, pp.~2811--2817, 2021.

\bibitem{siekmann2021blind}
J.~Siekmann, K.~Green, J.~Warila, A.~Fern, and J.~Hurst, ``Blind bipedal stair traversal via sim-to-real reinforcement learning,'' {\em arXiv preprint arXiv:2105.08328}, 2021.

\bibitem{castillo2022reinforcement}
G.~A. Castillo, B.~Weng, W.~Zhang, and A.~Hereid, ``Reinforcement learning-based cascade motion policy design for robust 3d bipedal locomotion,'' {\em IEEE Acc.}, vol.~10, pp.~20135--20148, 2022.

\bibitem{li2023robust}
Z.~Li, X.~B. Peng, P.~Abbeel, S.~Levine, G.~Berseth, and K.~Sreenath, ``Robust and versatile bipedal jumping control through reinforcement learning,'' {\em arXiv preprint arXiv:2302.09450}, 2023.

\bibitem{9108552}
A.~Iqbal, Y.~Gao, and Y.~Gu, ``Provably stabilizing controllers for quadrupedal robot locomotion on dynamic rigid platforms,'' {\em IEEE/ASME Trans. Mechatron.}, vol.~25, no.~4, pp.~2035--2044, 2020.

\bibitem{iqbal2023asymptotic}
A.~Iqbal, S.~Veer, and Y.~Gu, ``Asymptotic stabilization of aperiodic trajectories of a hybrid-linear inverted pendulum walking on a vertically moving surface,'' in {\em Proc. Amer. Contr. Conf.}, pp.~3030--3035, 2023.

\bibitem{iqbal2023analytical}
A.~Iqbal, S.~Veer, and Y.~Gu, ``Analytical solution to a time-varying lip model for quadrupedal walking on a vertically oscillating surface,'' {\em Mechatron.}, vol.~96, p.~103073, 2023.

\bibitem{iqbal2022drs}
A.~Iqbal, S.~Veer, and Y.~Gu, ``D{R}{S}-{L}{I}{P}: Linear inverted pendulum model for legged locomotion on dynamic rigid surfaces,'' {\em arXiv preprint arXiv:2202.00151}, 2022.

\bibitem{gao2023time}
Y.~Gao, Y.~Gong, V.~Paredes, A.~Hereid, and Y.~Gu, ``Time-varying alip model and robust foot-placement control for underactuated bipedal robotic walking on a swaying rigid surface,'' in {\em Proc. Amer. Contr. Conf.}, pp.~3282--3287, 2023.

\bibitem{bainov1993impulsive}
D.~Bainov and P.~Simeonov, {\em Impulsive differential equations: periodic solutions and applications}, vol.~66.
\newblock CRC Press, 1993.

\bibitem{gong2021one}
Y.~Gong and J.~Grizzle, ``One-step ahead prediction of angular momentum about the contact point for control of bipedal locomotion: Validation in a lip-inspired controller,'' in {\em Proc. IEEE Int. Conf. Rob. Autom.}, pp.~2832--2838, 2021.

\bibitem{nakanishi2008operational}
J.~Nakanishi, R.~Cory, M.~Mistry, J.~Peters, and S.~Schaal, ``Operational space control: A theoretical and empirical comparison,'' {\em Int. J. Rob. Res.}, vol.~27, no.~6, pp.~737--757, 2008.

\bibitem{khalil1996noninear}
H.~K. Khalil, {\em Noninear systems}.
\newblock No.~5, Prentice Hall, 1996.

\bibitem{branicky1998multiple}
M.~S. Branicky, ``Multiple lyapunov functions and other analysis tools for switched and hybrid systems,'' {\em IEEE Trans. Autom. Contr.}, vol.~43, no.~4, pp.~475--482, 1998.

\bibitem{gao2023provably}
Y.~Gao, K.~Barhydt, C.~Niezrecki, and Y.~Gu, ``{Global-Position Tracking Control for Multi-Domain Bipedal Walking with Underactuation},'' {\em ASME J. Dyn. Syst. Meas. Contr.}, pp.~1--27, 04 2024.

\end{thebibliography}
\bibliographystyle{ieeetr}

\begin{IEEEbiography}[{\includegraphics[width=1.0in,height=1.2in,clip]{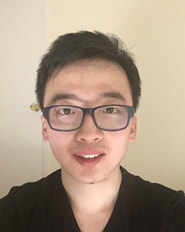}}]{Yuan Gao}
	received his B.S. degree in Mechanical Engineering from China Jiliang University, Hangzhou, China in 2014, the M.S. degree in Mechanical Engineering from Arizona State University in 2016, and the Ph.D. degree in Mechanical Engineering from University of Massachusetts Lowell in 2023.
	He is currently a control engineer at Brooks Automation.
 \vspace{-0.3 in}
\end{IEEEbiography}

\begin{IEEEbiography}[{\includegraphics[width=1.0in,height=1.2in,clip]{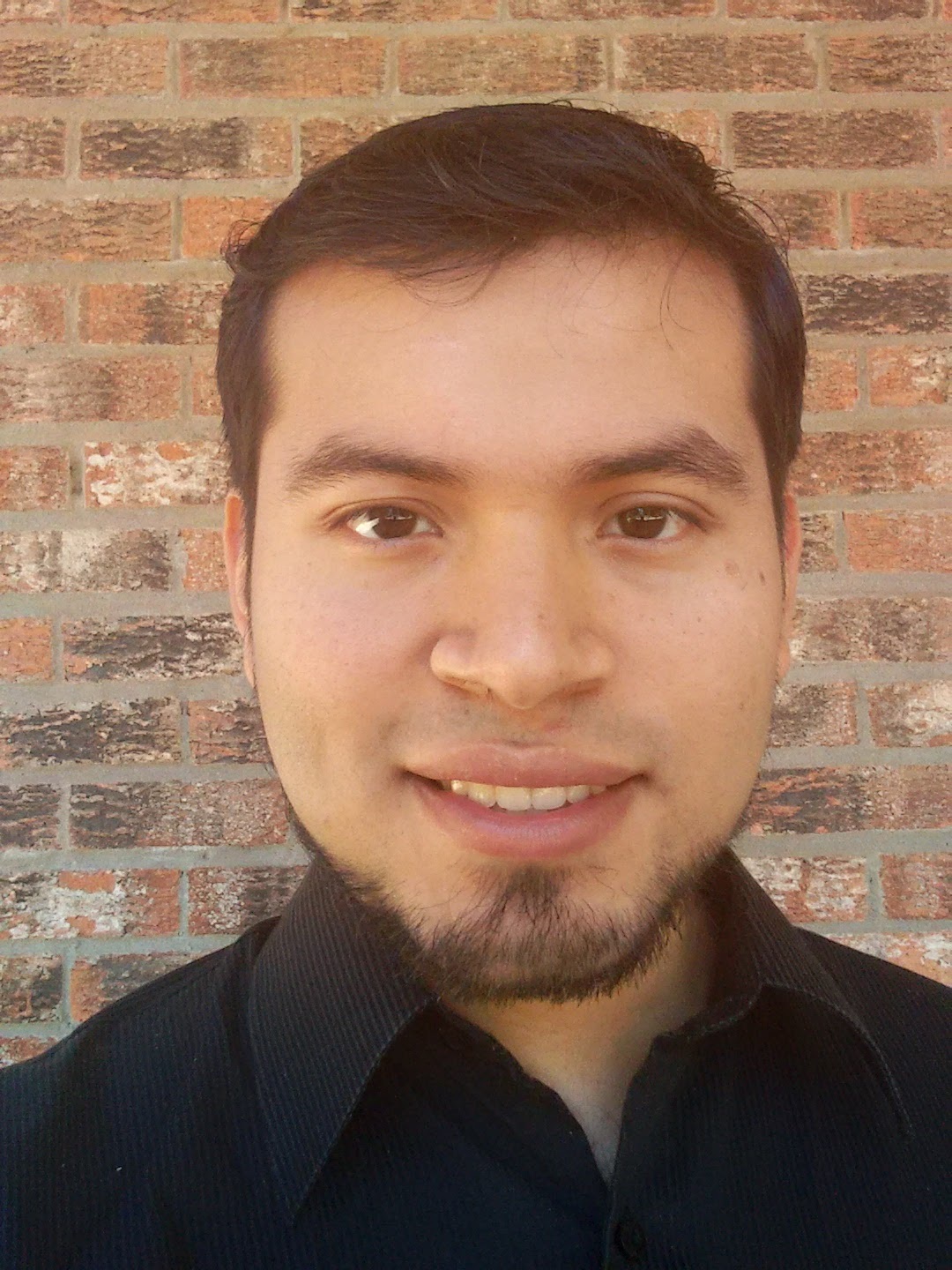}}]{Victor Paredes}
	received his BSc in Mechatronics from the National University of Engineering in Peru in 2012, and his MSc in Mechanical Engineering from Texas A\&M University in 2016. He is currently a PhD student at The Ohio State University, specializing in Humanoid Robotics. His research focuses on optimization, planning, and control to develop stable walking gaits, aiming to advance the field of humanoid robotics by enhancing the efficiency and stability of robotic locomotion.
  \vspace{-0.3 in}
\end{IEEEbiography}

\begin{IEEEbiography}[{\includegraphics[width=1.0in,height=1.3in,clip]{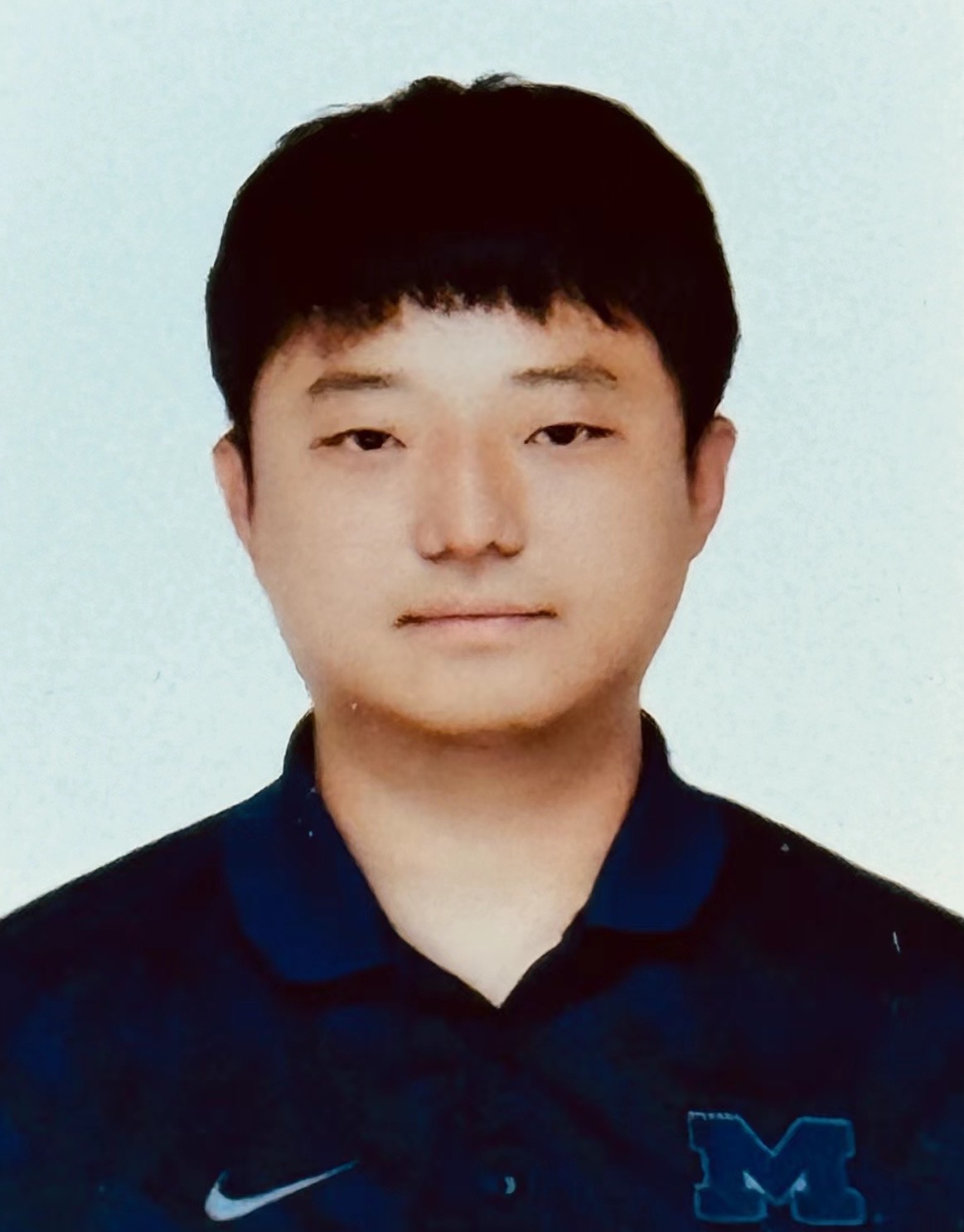}}]{Yukai Gong}
	received his B.E. degree in Naval Architecture and Ocean Engineering from Huazhong University of Science and Technology, Wuhan, China in 2015, his M.S.E degree in Mechanical Engineering and PhD. degree in Robotics from University of Michigan in 2022.
  \vspace{-0.3 in}
\end{IEEEbiography}

\begin{IEEEbiography}[{\includegraphics[width=1.0in,height=1.2in,clip]{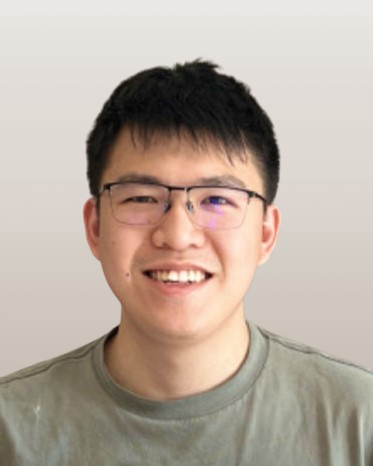}}]{Zijian He}
	received his B.Sc. and M.Sc. in Mechanical Engineering from Purdue University. He is currently a Ph.D. student at Purdue University. His research focuses on state estimation and control of bipedal humanoid robots.
   \vspace{-0.3 in}
\end{IEEEbiography}

\begin{IEEEbiography}[{\includegraphics[width=1in,height=1.2in,clip,keepaspectratio]{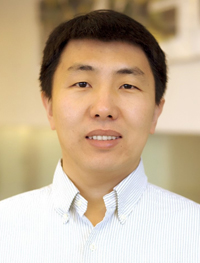}}]{Ayonga Hereid} is an Assistant Professor in the Department of Mechanical and Aerospace Engineering at the Ohio State University. He received his B.S. and M.S. degrees in Mechanical Engineering from Zhejiang University in 2007 and 2010, respectively, and received his Ph.D. degree in Mechanical Engineering from the Georgia Institute of Technology in 2016. Prior to joining the Ohio State University, he was a postdoctoral research fellow in the Department of Electrical Engineering and Computer Science at the University of Michigan in Ann Arbor. Dr. Hereid was the recipient of the NSF CAREER Award in 2022. His work was recognized for the Best Student Paper Award in 2014 from the ACM International Conference on Hybrid System: Computation and Control and was nominated as the Best Conference Paper Award Finalist in 2016 at the IEEE International Conference on Robotics and Automation. 
  \vspace{-0.3 in}
\end{IEEEbiography}

\begin{IEEEbiography}[{\includegraphics[width=1.0in,height=1.1in,clip]{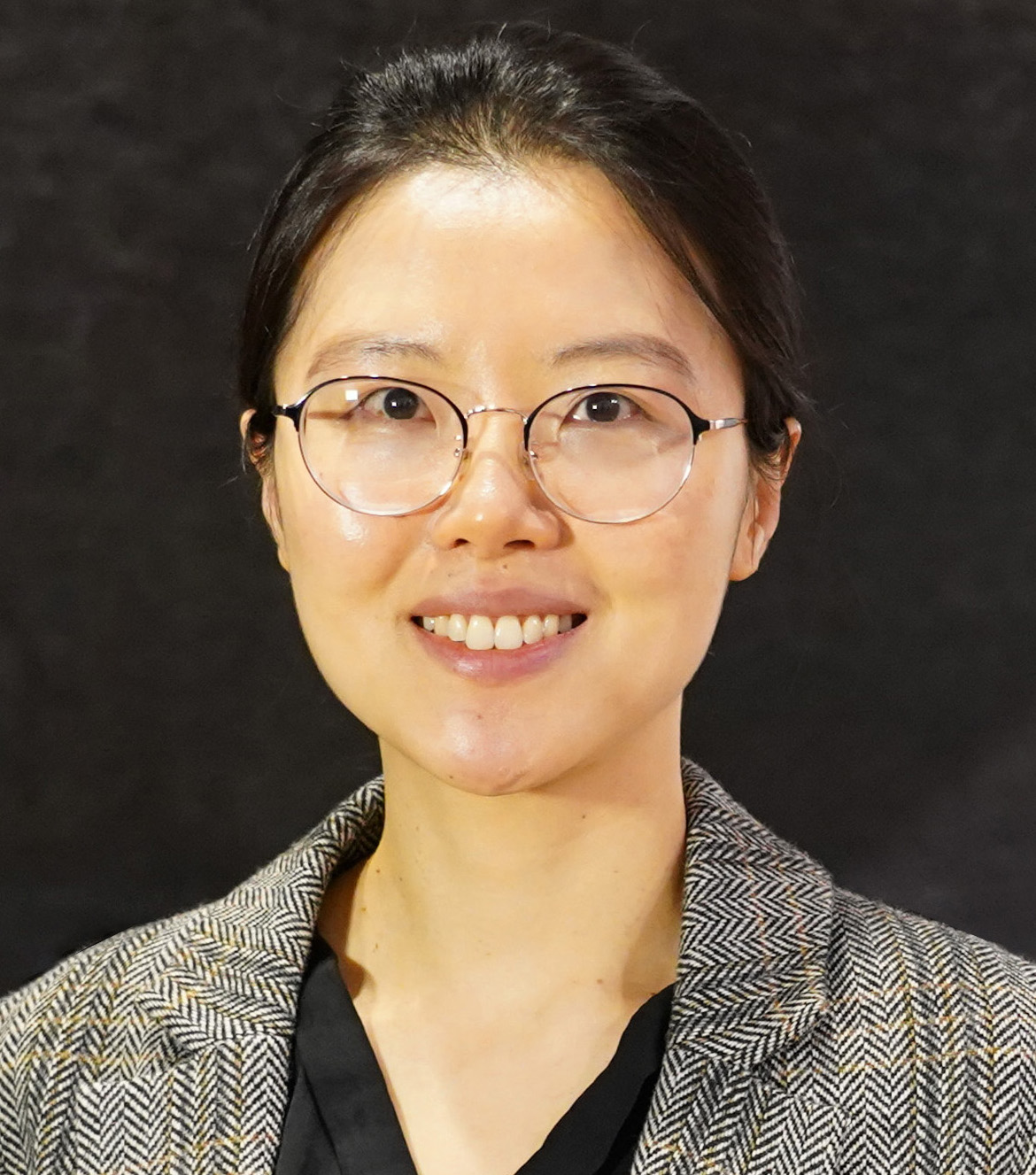}}]{Yan Gu}
received the B.S. degree in Mechanical Engineering from Zhejiang University, China, in June 2011 and the Ph.D. degree in Mechanical Engineering from Purdue University, West Lafayette, IN, USA, in August 2017.
She joined the faculty of the School of Mechanical Engineering at Purdue University in July 2022.
Prior to joining Purdue, she was an Assistant Professor with the Department of Mechanical Engineering at the University of Massachusetts Lowell.
Her research interests include nonlinear control, hybrid systems, and legged robotics.
She was the recipient of the NSF CAREER Award in 2021 and the ONR YIP Award in 2024.
  \vspace{-0.3 in}
\end{IEEEbiography}

\end{document}